\documentclass{article}

\usepackage[preprint]{corl_2026} %
\usepackage{graphicx}
\usepackage{multirow}
\usepackage{amssymb}
\usepackage{amsmath}
\usepackage{bm}
\usepackage{algorithm}
\usepackage{algpseudocode}
\usepackage{wrapfig}
\usepackage{placeins}
\usepackage{subcaption}
\usepackage[most]{tcolorbox}
\usepackage{booktabs}

\newcommand{\thiswork}{GIF}

\newtcolorbox{promptbox}[1][]{%
  enhanced,
  breakable,
  colback=gray!4,
  colframe=black!70,
  colbacktitle=black!70,
  coltitle=white,
  fonttitle=\bfseries\small\sffamily,
  boxrule=0.5pt,
  arc=2pt,
  left=6pt, right=6pt, top=4pt, bottom=4pt,
  fontupper=\scriptsize\ttfamily,
  before upper=\setlength{\parskip}{2pt}\raggedright,
  #1,
}
\newcommand{\promptsection}[1]{\par\smallskip\textbf{\textcolor{blue!60!black}{[#1]}}\par}
\newcommand{\authorsep}{\hspace{0.45em}}
\newcommand{\authorfont}{\fontsize{9.8}{11.5}\selectfont\bfseries}
\newcommand{\authorinfofont}{\fontsize{8.2}{9.8}\selectfont\normalfont}

\title{GIF: Agentic Generation of Interactive and Functional Object Compositions for Robot Learning}

\author{
  {\authorfont Long Xu\textsuperscript{*,2,4}\authorsep
  Zhiqi Zhang\textsuperscript{*,2,3}\authorsep
  Mi Yan\textsuperscript{*,1,2}\authorsep
  Shengliang Deng\textsuperscript{2,5}\authorsep
  Chong Xia\textsuperscript{2,6}}\\
  {\authorfont Mingyu Dong\textsuperscript{2,6}\authorsep
  Jiayi Chen\textsuperscript{1,2}\authorsep
  Jiangran Lyu\textsuperscript{1,2}\authorsep
  Fei Gao\textsuperscript{4}\authorsep
  Zhizheng Zhang\textsuperscript{\textdagger,2}\authorsep
  He Wang\textsuperscript{\textdagger,1,2}}\\[0.65em]
  {\authorinfofont
  \textsuperscript{1}CFCS, School of CS, Peking University \quad
  \textsuperscript{2}Galbot}\\[-0.05em]
  {\authorinfofont
  \textsuperscript{3}Peking University \quad
  \textsuperscript{4}Zhejiang University \quad
  \textsuperscript{5}The University of Hong Kong}\\[-0.05em]
  {\authorinfofont
  \textsuperscript{6}Tsinghua University}
}

\hypersetup{
  pdftitle={GIF: Agentic Generation of Interactive and Functional Object Compositions for Robot Learning},
  pdfauthor={Long Xu, Zhiqi Zhang, Mi Yan, Shengliang Deng, Chong Xia, Mingyu Dong, Jiayi Chen, Jiangran Lyu, Fei Gao, Zhizheng Zhang, He Wang},
  pdfkeywords={}
}

\makeatletter
\renewcommand{\@noticestring}{%
  \parbox{\textwidth}{%
    \raggedright
    \rule{12pc}{0.4pt}\\[0.08em]
    \textsuperscript{*}Equal contribution. \textsuperscript{\textdagger}Corresponding authors.\\[0.10em]
    Correspondence: \href{mailto:zhangzz@galbot.com}{\texttt{zhangzz@galbot.com}},
    \href{mailto:hewang@pku.edu.cn}{\texttt{hewang@pku.edu.cn}}.
  }%
}
\makeatother

\begin{document}
\maketitle

\begin{figure}[H]
  \centering
  \includegraphics[width=\linewidth]{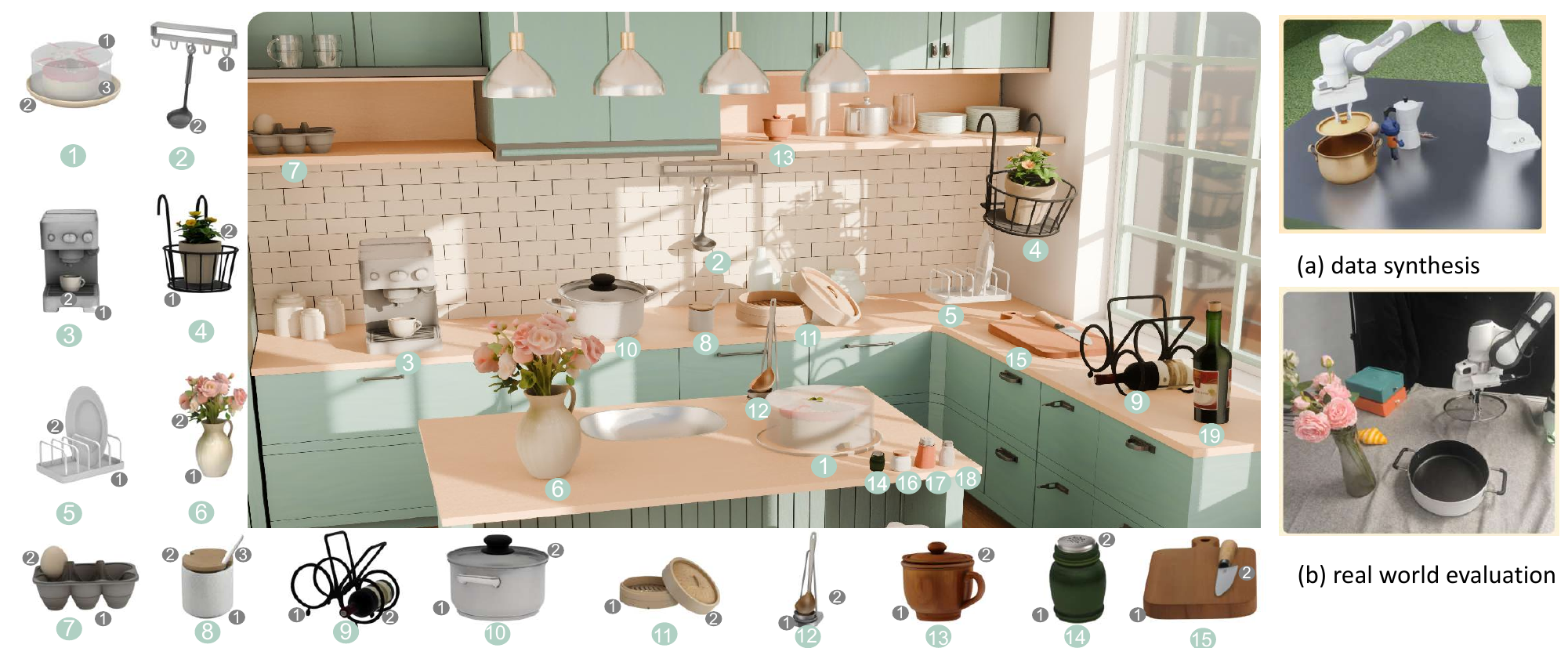}
  \caption{Overview of \thiswork{}. 
  Left: diverse, interactive and functional task compositions, shown individually and arranged in a kitchen scene. Cyan labels match each composition to its placement in the scene; gray labels identify the independently interactive components within each composition (e.g., a pot and its lid). Right: the generated assets support robot training data synthesis in simulation (a) and deployment of the trained policy on a real robot (b).}
  \label{fig:teaser}
\end{figure}

\begin{abstract}
Robot manipulation foundation models require scalable evaluation and data generation across diverse scenarios, with simulation providing an environment for both.
Automated scene generation offers a promising path, yet prior work has largely emphasized coarse-grained scene layouts rather than fine-grained functional object compositions.
Motivated by this gap, we present \thiswork{}, an agentic \textbf{G}eneration framework for \textbf{I}nteractive and \textbf{F}unctional object compositions. In this framework, we recast this problem as disentangled reconstruction followed by relative pose recovery.
CoGen produces instance-disentangled meshes with coarse initial poses leveraging complementary strengths of 2D and 3D generative models.
GPRM refines the relative pose under joint geometric and physical guidance, and a VLM verifier selects the candidate that best matches the structured specification.
We further construct a benchmark spanning eight representative contact-geometry classes and compare with state-of-the-art generators; \thiswork{} improves both asset quality and relation matching, while reducing collision rate to below $1\%$.
Finally, we synthesize data for policy learning, revealing diversity scaling in both simulation and real-world deployment.
\end{abstract}

\section{Introduction}
\label{sec:intro}

Recently, embodied foundation models~\citep{pi05} have shown increasingly general manipulation capabilities, necessitating scalable pipelines for robot evaluation and data generation. Simulation offers a low-cost, massively parallelizable alternative, but constructing simulation scenes has traditionally required substantial manual design by domain experts, motivating recent work on automated scene generation~\citep{deitke2024holodeck,pfaff2026scenesmith,yang2025sceneweaver,ning2025mesatask,mu2025robotwin,geniesim2025,tabletopgen2025} that has made promising progress on diverse simulated environments.

Existing systems have advanced diverse and semantically plausible scene generation, particularly for navigation and scene-level reasoning, but they mainly focus on scene-scale layout, specifying approximate placements on tables, furniture, or within rooms. Manipulation instead requires object-level functional relations governed by local geometric compatibility, collision avoidance, and physical stability. Tasks such as hanging, insertion, capping, and containment demand precise relative 6-DoF poses aligned with human intention and physical constraints. Motivated by this gap, we study interactive and functional object composition generation: given a language or image prompt, the goal is to generate independently interactive objects together with relative poses that realize precise functional contact, avoid interpenetration, and remain stable in simulation.

While existing systems already yield diverse and high-fidelity 3D assets~\citep{hunyuan3d25,sam3d}, their strategies still face substantial challenges when recovering precise relative poses. Language-prompted methods~\citep{geniesim2025,yang2025sceneweaver,pfaff2026scenesmith} ask VLMs to emit positions or rotation angles yet lack precise local 3D geometry understanding, while visual-prompted methods~\citep{wei2025rola,tabletopgen2025} rely on neural pose estimators that suffer out-of-distribution issues~\citep{wen2024foundationpose} or numerical optimization hindered by ill-posed objectives and unreliable initializations. To bridge this pose gap, motivated by the fine-grained controllability of modern generative models over appearance, geometry, and pose, we propose \thiswork{}, an agentic framework that casts scene generation as reconstruction and pose recovery. We focus on two-object interactions as the minimal setting where the pose problem already manifests, while the formulation extends naturally to multi-object scenes (Sec.~\ref{sec:exp:extension}). Since recovering correct relative poses presupposes geometrically complete and instance-disentangled assets, \thiswork{} first obtains such assets through CoGen, which couples 2D amodal completion with image-conditioned 3D reconstruction so that mutually occluded objects emerge as clean, separately interactive meshes. GPRM then refines the relative pose under joint geometric and physical guidance, using a VLM verifier to inspect simulated rollouts for selecting the candidate that best matches the structured specification. Each verified composition can serves as a compact task specification for downstream policy learning.

In summary, our contributions are threefold. 1) We formulate open-ended generation of functional two-object compositions as a contact-precise, simulation-ready scene-generation problem. 2) We introduce \thiswork{}, which combines CoGen for disentangled asset reconstruction with GPRM for geometry- and physics-guided pose refinement. 3) We build a benchmark spanning eight contact-geometry classes and show across 120 generated scenes that \thiswork{} outperforms strong baselines on relation matching, object quality, collision rate, and human preference.

\vspace{-0.2cm}
\section{Related Work}
\label{sec:related}
\vspace{-0.2cm}

\textbf{Simulation-Ready Scene Generation for Robot Learning.}
Robot learning has long relied on simulators and benchmarks that provide physics, assets, tasks, and data-generation interfaces, including AI2-THOR~\citep{ai2thor2017}, Habitat~\citep{savva2019habitat,szot2021habitat2}, iGibson~\citep{shen2021igibson,li2021igibson2}, TDW~\citep{gan2020threedworld}, BEHAVIOR-1K~\citep{li2023behavior1k}, SAPIEN~\citep{xiang2020sapien}, robosuite~\citep{zhu2020robosuite}, ManiSkill2~\citep{gu2023maniskill2}, RLBench~\citep{james2020rlbench}, and Isaac Gym~\citep{makoviychuk2021isaacgym}.
Indoor and embodied scene generation extends this infrastructure from priors, procedures, language, or agents, spanning layout datasets and models~\citep{fu20213dfront,ritchie2019fastflexible,paschalidou2021atiss,wang2021sceneformer,yang2023commonscenes,hollein2023text2room} and recent automated generators~\citep{deitke2022procthor,raistrick2024infinigen,tang2024diffuscene,deitke2024holodeck,deitke2025holodeck2,sun2024layoutvlm,deng2025sage,lu2025scenethesis,yang2025sceneweaver,lin2025pat3d,che2026mansion}.
Robotics-oriented systems further add task resources, asset acquisition, domain randomization, and simulator export~\citep{nasiriany2024robocasa,robogen2024,mu2025robotwin,geniesim2025,gensim2024,katara2024gen2sim,grs2025,nguyen2025regen,lee2025dynscene,pfaff2026scenesmith,ning2025mesatask,tabletopgen2025}.
These methods mainly reason over rooms, tables, support surfaces, or coarse spatial relations; \thiswork{} instead focuses on the contact-precise two-object composition needed for functional relations such as capping, hooking, slotting, spanning, and pegging.

\textbf{Image-Conditioned Asset Construction and Interactive Object Compositions.}
Contact-precise composition requires assets that are visually plausible, separately manipulable, and equipped with usable collision geometry.
Open-vocabulary perception and editing models support object isolation and reference-image control~\citep{liu2023groundingdino,li2022glip,minderer2022owlvit,suvorov2022lama,lugmayr2022repaint,zhang2023controlnet}, while text- and image-conditioned 3D generation has advanced through score-distillation and reconstruction models~\citep{poole2022dreamfusion,lin2023magic3d,chen2023fantasia3d,wang2023prolificdreamer,gao2022get3d,nichol2022pointe,jun2023shapee,liu2023zero123,liu2023syncdreamer,long2023wonder3d,liu2023one2345,hong2023lrm,tochilkin2024triposr,xu2024instantmesh,tang2024lgm,wang2026scenetransporter}.
Image-conditioned reconstruction and asset-acquisition pipelines recover geometry, texture, and pose from real or generated images~\citep{wei2025rola,dai2024digitalcousins,yao2024metascenes,chen2026anyrecon,vig2026,hunyuan3d2,hunyuan3d25,chen2024meshanything,sam3d,wu2025amodal3r}. Multi-instance compositional methods such as REPARO~\citep{han2025reparo} jointly generate multiple assets and align their 3D layout. Systems such as MIDI~\citep{huang2025midi}, TabletopGen~\citep{tabletopgen2025}, Interact3D~\citep{interact3d2026}, and WorldAct~\citep{hu2026worldact} assemble or activate multiple interactive objects.
For functional object pairs, however, mutual occlusion occurs exactly at the contact interface, causing direct reconstruction to fuse instances, leak partner geometry, or hallucinate hidden surfaces. CoGen addresses this asset-level bottleneck by producing instance-disentangled meshes and coarse initial poses for downstream refinement.

\textbf{Object-to-Object Affordance and Composition Pose Recovery.}
Once assets are available, existing placement and affordance methods still leave contact-precise relative pose recovery for functional object compositions largely unresolved.
Language- or agent-prompted scene methods delegate placement to coarse LLM/VLM specifications or layout optimization~\citep{geniesim2025,tabletopgen2025,yang2025sceneweaver,pfaff2026scenesmith,sun2024layoutvlm}, while visual-prompted systems rely on pose estimators that can fail out of distribution~\citep{wei2025rola,wen2024foundationpose,wang2019densefusion,hodan2018bop} or on classical registration with fragile initialization~\citep{icp}.
Object-to-object affordance, placement, rearrangement, and motion-planning work highlights the importance of relational geometry~\citep{simeonov2022ndf,pan2023taxpose,tian2025o3afford,eisner2024anyplace,liu2022structformer,liu2022structdiffusion,zeng2021transporters,shridhar2022cliport,shridhar2023peract,mo2021where2act,yang2024cluttergen,yang2024physcene,architect2024,sceneassistant2026,spatialvlm2025,dalal2024neuralmp}. OOR~\citep{baik2025learning} learns category-level spatial relation priors from a text-conditioned diffusion model, but assumes canonical category meshes and does not construct simulation-ready assets. Existing methods typically assume existing assets, fixed placement families, support-surface plausibility, or collision avoidance rather than tight semantic contact. GPRM instead refines CoGen's initial poses through registration, renderbased alignment, collision-aware optimization, simulation-aware search, and VLM selection, yielding compositions whose success depends on contact rather than mere non-overlap.

\vspace{-0.2cm}
\section{Method}
\vspace{-0.2cm}
\label{sec:method}
Given an input $\mathcal{I}$ specifying a functional object pair $(o_1, o_2)$ together with their attributes and spatial relation, either as a language instruction $\mathcal{I}=T$ or a reference image $\mathcal{I}=I_{\rm input}$, the goal is to generate a simulation-ready scene $\mathcal{S}=\{(m_i,p_i)\}_{i\in\{1,2\}}$, where $m_i$ is a mesh for the rigid object $o_i$ and $p_i=(r_i,t_i)\in\mathrm{SO}(3)\times\mathbb{R}^3$ is its camera-frame pose.

\begin{figure}[t]
    \centering
    \includegraphics[width=1.0\textwidth]{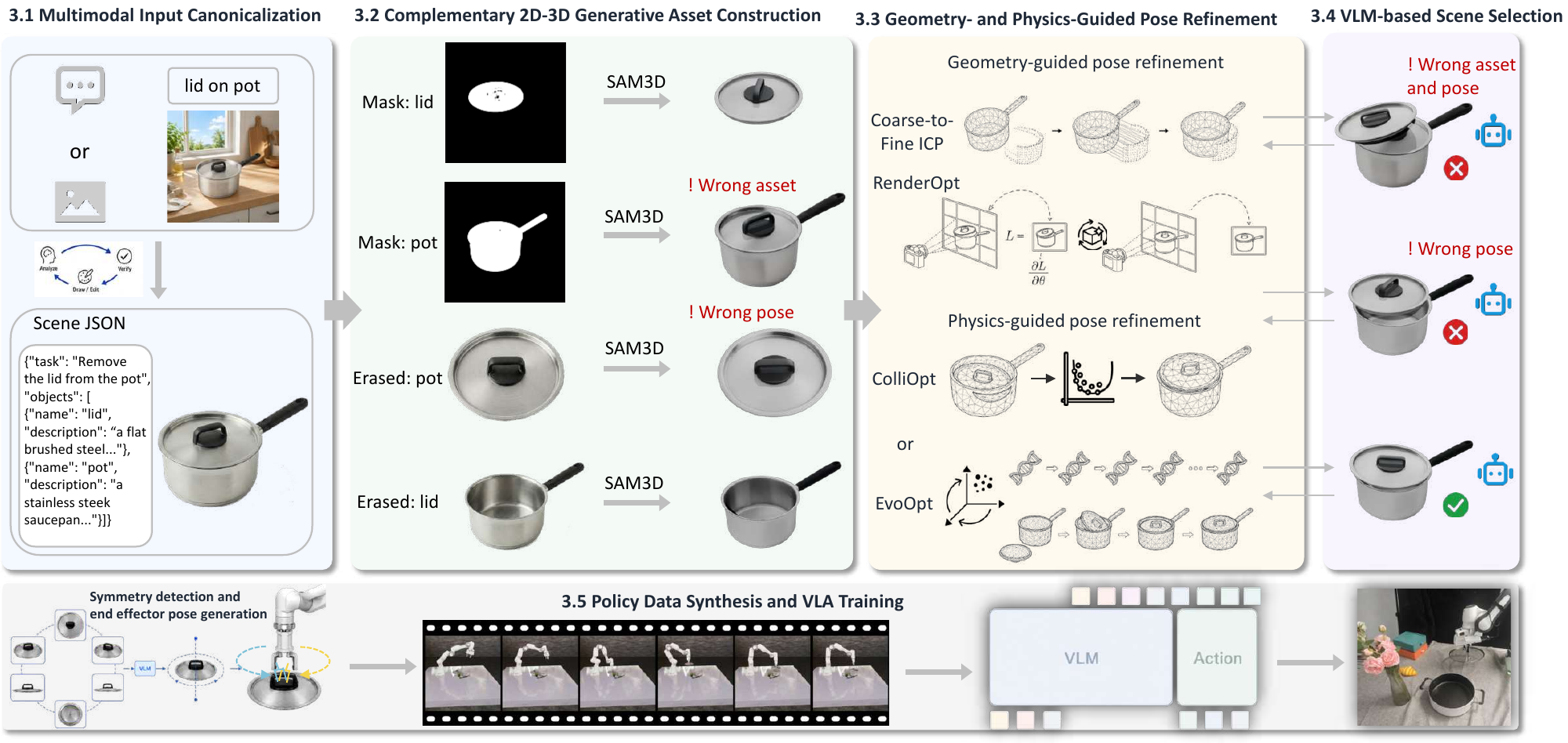}
    \caption{Overview of \thiswork.
    The multimodal input agent canonicalizes a language or image prompt into a unified specification $(I_{\rm ref}, J_{\rm scene})$, comprising a foreground-only reference image and a structured description of the objects, their attributes, and the target relation.
    Complementary 2D-3D Generative Asset Construction (CoGen) uses this specification to generate candidate meshes and initial poses for each object through direct image-to-3D reconstruction and an erase-and-regenerate branch that mitigates occlusion and instance entanglement.
    Geometry- and Physics-Guided Pose Refinement Modules (GPRM) iteratively refine the relative object poses, while a VLM verifies each candidate scene and selects a valid scene or requests further refinement.
    The verified scene defines the target configuration for trajectory synthesis, and the synthesized trajectories are used to train a VLA model for real-world deployment.}
    \label{fig:pipeline}
  \end{figure}

As shown in Fig.~\ref{fig:pipeline}, a multimodal agent (Sec.~\ref{sec:method:input}) canonicalizes $\mathcal{I}$ into a unified specification, CoGen (Sec.~\ref{sec:method:asset}) supplies instance-disentangled meshes together with coarse initial poses, GPRM (Sec.~\ref{sec:method:pose}) refines the relative pose under joint geometric and physical guidance, and a VLM selector (Sec.~\ref{sec:method:voter}) verifies each candidate scene and triggers further refinement when needed. The verified scene is then used as the terminal configuration for trajectory synthesis and VLA policy learning (Sec.~\ref{sec:method:data}).

\subsection{Multimodal Input Canonicalization}
\label{sec:method:input}
\vspace{-0.2cm}
A multimodal agent canonicalizes any input form of $\mathcal{I}$ into a unified specification $(I_{\mathrm{ref}}, J_{\mathrm{scene}})$, where $I_{\mathrm{ref}}$ is a foreground-only reference image of the desired object pair and $J_{\mathrm{scene}}$ is a structured JSON record of object identities, asset attributes, and the target relation. The agent runs an analyze--act--verify loop that first identifies $o_1$, $o_2$, and the relation, then invokes either an image-generation tool (when only $T$ is given) or an image-editing tool (when $I_{\mathrm{input}}$ is given) to produce a candidate $I_{\mathrm{ref}}$, after which a VLM verifier accepts or calls editing or redrawing until acceptance or a iteration budget is exhausted. How we prepare simulation ready assets with convex decomposition are provided in the supplementary.

\vspace{-0.2cm}
\subsection{CoGen: Complementary 2D-3D Generative Asset Construction}
\label{sec:method:asset}
\vspace{-0.2cm}
GPRM requires instance-disentangled meshes together with reasonably aligned initial poses. A natural approach lifts each object directly with an off-the-shelf image-to-3D model~\citep{sam3d}, conditioned on $I_{\rm ref}$ and a binary instance mask $b_i^{\rm ref}$ derived from $J_{\rm scene}$, yielding $(m_i^{\rm ref},p_i^{\rm ref})$. But this single-pass reconstruction is insufficient for contact-rich pairs. Existing segmentation and 3D models lack amodal reasoning, so heavy mutual occlusion (e.g., a flower stem inside a vase) leaves meshes that explain only the visible region, and limited instance disentanglement lets partner geometry leak through a single-object mask, e.g., reconstructing a pot together with its lid.

We therefore add a complementary erase-and-regenerate branch that exploits the larger training corpora and stronger object priors of 2D generative models. The branch erases $o_j$ from $I_{\rm ref}$ and completes $o_i$, obtaining an occlusion-free image $I_i^{\rm erase}$ that SAM3D~\citep{sam3d} lifts to $(m_i^{\rm erase},p_i^{\rm erase})$. Since 2D models still provide imperfect pose control and may slightly perturb the kept object, both candidates $(m_i^{\rm ref},p_i^{\rm ref})$ and $(m_i^{\rm erase},p_i^{\rm erase})$ are retained as initial mesh-pose pairs for GPRM. More details can be found in the supplementary materials.

\vspace{-0.2cm}
\subsection{GPRM: Geometry- and Physics-Guided Pose Refinement Modules}
\label{sec:method:pose}
\vspace{-0.2cm}

To obtain accurate relative poses, we refine CoGen's coarse mesh--pose candidates with geometry- and physics-guided modules and verify each refined scene with the VLM selector in Sec.~\ref{sec:method:voter}. For each $o_i$, let $a_i\in\mathcal{A}=\{\mathrm{ref},\mathrm{erase}\}$ index the two candidates $(m_i^{a_i},p_i^{a_i})$. We enumerate the four pairings $\mathcal{C}=\{(a_1,a_2)\}$, initialize $\mathcal{S}^{0}_{c}=\{(m_i^{a_i},p_i^{a_i})\}_{i=1}^{2}$ for each $c$, and run a GPRM--VLM loop $\mathcal{S}^{0}_{c}\!\rightarrow\!\cdots\!\rightarrow\!\mathcal{S}^{K_c}_{c}$ until acceptance or budget exhaustion, yielding $\widehat{\mathcal{S}}_{c}=\mathcal{S}^{K_c}_{c}$. 
Fig.~\ref{fig:geo_opt} illustrates two such cases. Further GPRM details are in the supplementary.

\noindent\textbf{Geometry-Guided Refinement I. Coarse-to-Fine ICP.}
ICP~\citep{icp} aligns a 3D model to observed geometry by iterating nearest-neighbour correspondences and minimizing point-to-point distance, requiring no task-specific priors. We sample the source cloud from $m_i^{a_i}$ at the candidate pose and back-project the reference depth from a monocular geometry estimator~\citep{mogev2} through the eroded instance mask to form the per-object target cloud, and progressively tighter ICP rounds.

\noindent\textbf{Geometry-Guided Refinement II. Render-Based Optimization (RenderOpt).}
ICP is sensitive to incomplete point clouds and to errors from the monocular estimator, both of which bias correspondence selection. We therefore add a 2D refinement stage that rasterises each candidate mesh with nvdiffrast~\citep{nvdiffrast} and updates the pose and scale so that the rendered silhouette and depth match the reference image, with per-object recall and precision terms preventing leakage between instances and a depth term resolving occlusion order. Regularization terms anchor the solution to the ICP initialization. A refined pose $p_i^{a_i,\rm rend}$ is accepted only if every per-object loss component improves.

\begin{figure}[t]
  \centering
  \includegraphics[width=\linewidth]{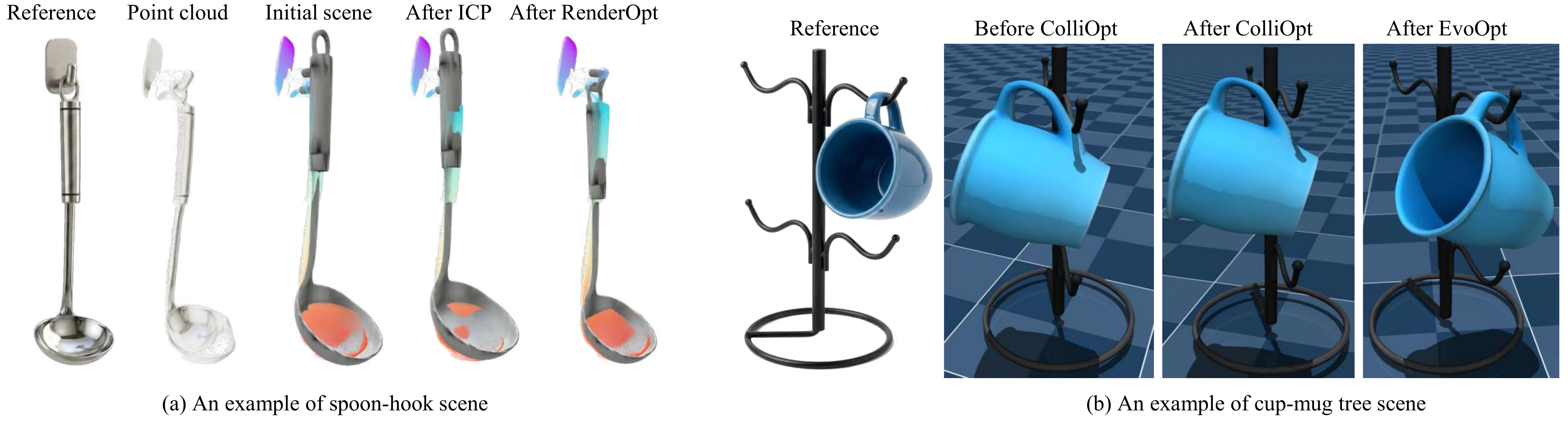}
  \caption{Two cases of geometry- and physics-guided refinement. (a) Spoon hooked on a wall hook: ICP drags the spoon inward, and RenderOpt tightens its scale so the ring catches the hook. (b) Cup hung on a mug tree: ColliOpt removes interpenetration but the cup still slips under gravity, while EvoOpt searches under simulation feedback to land on a pose that remains hooked.}
  \label{fig:geo_opt}
\end{figure}

\noindent\textbf{Physics-Guided Refinement I. Collision-Avoidance Optimization (ColliOpt).}
RenderOpt minimizes image-space discrepancies but does not guarantee physical validity, so the resulting meshes may still interpenetrate and yield undefined behaviour in MuJoCo. We add a constrained pass that finds the nearest collision-free pose to $p_i^{a_i,\rm rend}$ by enforcing separation between the convex-decomposition collision pieces of different objects, together with a ground-clearance constraint. The problem is built with Drake~\citep{drake} and solved with SNOPT~\citep{snopt}.

\noindent\textbf{Physics-Guided Refinement II. Simulation-Aware Evolutionary Optimization (EvoOpt).}
Resolving collisions in isolation does not guarantee that the relation survives gravity. As Fig.~\ref{fig:geo_opt}(b) shows, the cup remains collision-free yet still slips off the mug tree in simulation. We therefore close the loop by treating MuJoCo~\citep{todorov2012mujoco} as a black-box objective and search the pose neighbourhood with CMA-ES~\citep{cmase} from EvoTorch~\citep{evotorch}. Each candidate is encoded as a translation delta in $\mathbb{R}^3$ and a rotation delta in the 6D representation~\citep{continuity}, and is scored under $T_{\rm sim}$ steps by a fitness that penalizes drift, residual penetration, residual velocity, and deviation from the input pose. EvoOpt is launched whenever the VLM selector rejects all earlier refinements.

\subsection{VLM-based Scene Selection}
\label{sec:method:voter}

For each initial or refined scene, we load it into MuJoCo, simulate until convergence or a step budget is reached, and tile multi-view, multi-timestep renders into a single image whose sub-panels carry view labels and step indices. A VLM verifier inspects this image to check whether the target relation holds in the initial frame and is preserved in the final frame. Among scenes that pass verification, a second VLM call ranks the surviving candidates by horizontally concatenating their per-candidate panels and returning the index that best matches $J_{\rm scene}$. An example of the cross-candidate montage and full prompts are provided in the supplementary.

\subsection{Policy Data Synthesis and VLA Training}
\label{sec:method:data}

Given a VLM-verified target scene $\widehat{\mathcal{S}}$, we treat its arrangement as the task's terminal configuration and synthesize demonstrations that realize it. For each rollout we randomize the table-top poses to obtain $\mathcal{S}^{\rm rand}=\{(m_i,p_i^{\rm rand})\}_{i=1}^{2}$ and define a valid terminal scene $\mathcal{S}^{\star}$ via $(p_2^{\star})^{-1}p_1^{\star}=p_2^{-1}p_1$. With $m_1$ as manipuland and $m_2$ fixed, this yields $p_2^{\star}=p_2^{\rm rand}$ and $p_1^{\star}=p_2^{\rm rand} p_2^{-1} p_1$, after which a motion planner produces a collision-free trajectory from $p_1^{\rm rand}$ to $p_1^{\star}$. To avoid redundant supervision, a VLM annotates per-object symmetry classes and trajectory targets are defined over the resulting pose-equivalence classes. We pair high-fidelity ray-traced rendering with domain randomization over lighting, textures, camera parameters, and object poses to narrow the sim-to-real gap, following recent VLA pipelines~\citep{geniesim2025,mu2025robotwin,graspvla}. We train $\pi_{0.5}$~\cite{pi05} purely on the synthesized trajectories with wrist-view and third-person view images as input and action chunks as output for direct real-robot deployment. Implementation details are provided in the supplementary.

\section{Experimental Results}
\label{sec:exp}

\subsection{Experimental Setup}
\label{sec:exp:setup}

\textbf{Taxonomy and Benchmark Suite.}
We construct a taxonomy of pairwise contact-geometry relations grounded in real human manipulation by deploying an LLM-based agent to classify and aggregate activity descriptions from Ego4D~\citep{grauman2022ego4d}, yielding 8 representative types: C.1 Containment, C.2 Capping, C.3 Resting, C.4 Leaning, C.5 Hooking, C.6 Slotting, C.7 Spanning, and C.8 Pegging. The same agent then selects three manipulation tasks per category to form a $8\times 3=24$-task benchmark, with the full task list provided in the supplementary. Each task is instantiated with five semantic variants for a total of $24\times 5=120$ scenes that methods are evaluated on.

\textbf{Metrics.}
We evaluate asset, relation, and physical correctness with three metrics. Object Matching (O.M., $[0,10]$) compares the asset's appearance and geometry against the reference image, Relationship Matching (R.M., $[0,10]$) measures whether the final configuration realizes the specified functional contact (e.g., a lid capping a pot rather than merely placed nearby), and Collision Rate (C.R., $[0,1]$) reports the fraction of seeds whose MuJoCo configuration exhibits inter-object penetration. For language prompts, the VLM scorer receives only the task description and the settled scene; it does not receive an intermediate image from GIF and therefore judges O.M. by instruction-consistent object appearance rather than pixel agreement with a hidden reference. For image prompts, the scorer additionally receives the reference image. We use complementary VLM and human protocols, with detailed confidence intervals and prompts in the supplementary.

\textbf{Baselines and Comparison Protocol.}
We compare against four representative systems and field each one under the protocol that best matches its assumptions. RoLA~\citep{wei2025rola} receives manually prepared instance masks for its released interface, and TabletopGen~\citep{tabletopgen2025} receives table-conditioned image prompts because the system require a support surface. GenieSim~\citep{geniesim2025} is run fully automatically with human feedback disabled and a forced asset-acquisition instruction. SceneSmith~\citep{pfaff2026scenesmith} is evaluated only at its manipuland-placement stage, with a fixed desk and pairwise object scope. Human feedback has zero runtime budget for every method to evaluate what these systems achieve without human intervention, and image-to-3D modules are uniformly replaced with SAM3D~\citep{sam3d}; the complete adaptation protocol is in the supplementary.

\subsection{Main Results}
\label{sec:exp:main}

\begin{table*}[t]
\centering
\scriptsize
\setlength{\tabcolsep}{4pt}
\caption{Main results across the eight contact-geometry classes from C.1 Containment to C.8 Pegging. The upper block reports per-class Object Matching (O.M.\,$\uparrow$) and Relationship Matching (R.M.\,$\uparrow$), each averaged over five seeds per task and three tasks per class. The lower block summarizes Collision Rate (C.R.\,$\downarrow$) averaged across all eight classes. Bold marks the best per-column score within each input setting.}
\label{tab:main_results}

\begin{tabular*}{\textwidth}{@{\extracolsep{\fill}}lcccccccccc@{}}
\hline
\multirow{2}{*}{\rule{0pt}{2.6ex}\textbf{Method}}
& \multicolumn{8}{c|}{\rule{0pt}{2.6ex}\textbf{Object Matching\,$\uparrow$ \ \ /\ \ Relationship Matching\,$\uparrow$}}
& \multicolumn{1}{c}{\rule{0pt}{2.6ex}\textbf{Physical}} \\
\cline{2-10}
& \rule{0pt}{2.4ex}C.1 & C.2 & C.3 & C.4 & C.5 & C.6 & C.7 & C.8
& C.R.\,$\downarrow$ \\
\hline
\multicolumn{10}{l}{\textit{User Input: Image Prompt}} \\
RoLA~\citep{wei2025rola}
& 8.00 / 3.80 & 8.60 / 5.07 & 8.33 / 4.33 & 8.47 / 3.07
& 8.00 / 2.80 & 7.40 / 3.33 & 7.87 / 4.47 & 8.20 / 6.20
& 0.558 \\
TabletopGen~\citep{tabletopgen2025}
& 6.60 / 1.33 & 9.13 / 8.67 & 9.13 / 3.73 & 5.33 / 2.07
& 7.07 / 1.73 & 7.40 / 3.60 & 8.60 / 5.00 & 7.60 / 5.27
& \textbf{0.000} \\
Ours-Img
& \textbf{9.07} / \textbf{7.80} & \textbf{9.60} / \textbf{9.20} & \textbf{9.53} / \textbf{9.27} & \textbf{9.67} / \textbf{9.47}
& \textbf{9.00} / \textbf{4.47} & \textbf{9.40} / \textbf{7.33} & \textbf{9.60} / \textbf{7.53} & \textbf{8.73} / \textbf{7.53}
& \textbf{0.000} \\
\hline
\multicolumn{10}{l}{\textit{User Input: Language Prompt}} \\
GenieSim~\citep{geniesim2025}
& 4.87 / 3.47 & 4.40 / 1.07 & 4.07 / 2.13 & 2.73 / 0.80
& 2.53 / 0.87 & 6.07 / 2.13 & 3.07 / 1.00 & 2.87 / 0.33
& 0.217 \\
SceneSmith~\citep{pfaff2026scenesmith}
& \textbf{8.93} / 6.60 & \textbf{9.40} / 7.60 & 8.67 / 6.93 & 8.27 / 6.33
& 7.20 / 2.27 & 8.93 / 6.80 & 9.47 / 6.67 & 7.93 / 7.47
& 0.083 \\
Ours
& \textbf{8.93} / \textbf{7.60} & 9.27 / \textbf{9.00} & \textbf{9.33} / \textbf{8.20} & \textbf{9.13} / \textbf{8.27}
& \textbf{8.80} / \textbf{5.20} & \textbf{9.00} / \textbf{7.47} & \textbf{9.53} / \textbf{7.80} & \textbf{8.80} / \textbf{9.00}
& \textbf{0.008} \\
\hline
\end{tabular*}
\end{table*}

\begin{figure}[t]
    \centering
    \includegraphics[width=\linewidth]{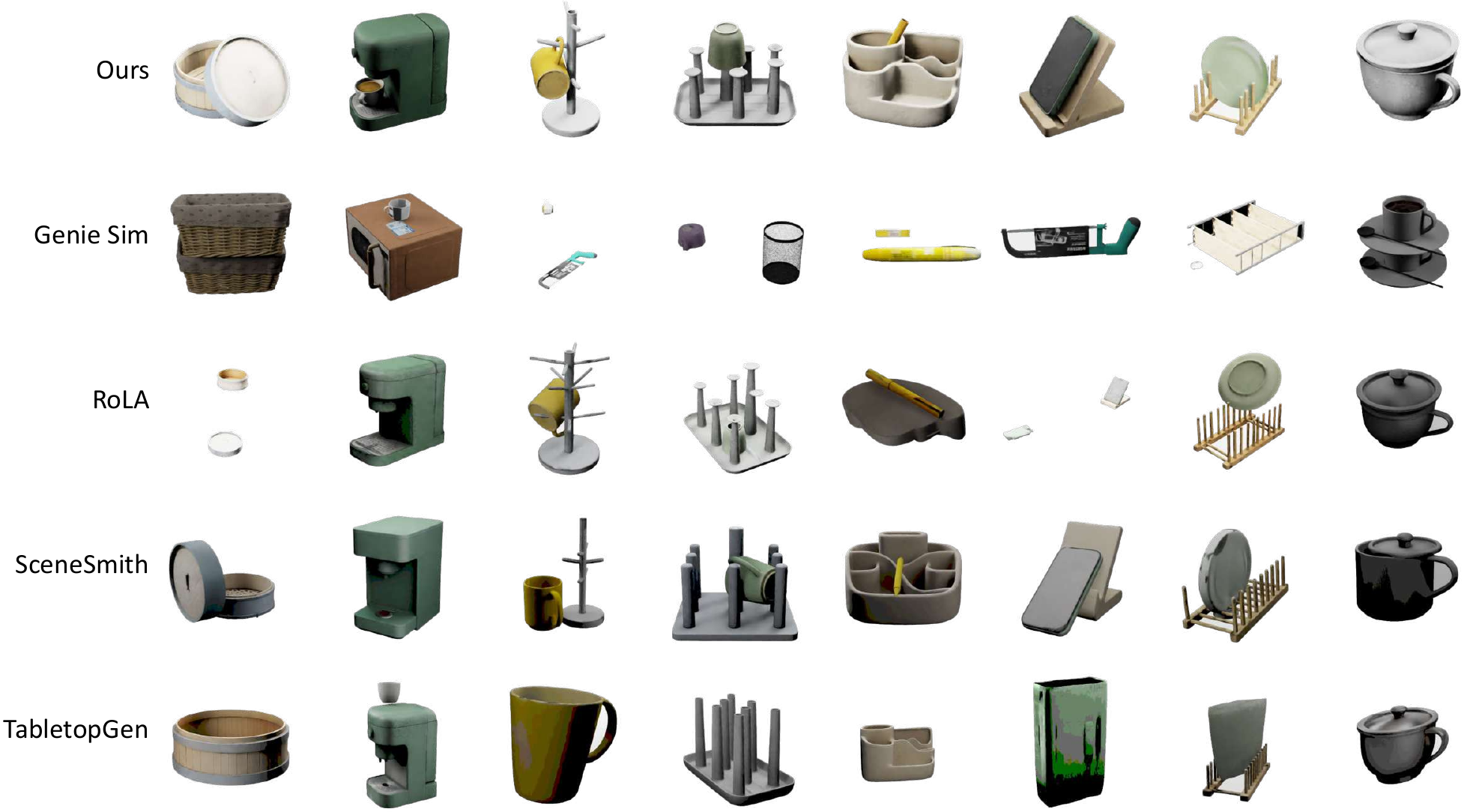}
    \caption{Comparisons between our method and baselines after MuJoCo settling. GenieSim~\citep{geniesim2025} is library-bound and frequently lacks a suitable asset. RoLA~\citep{wei2025rola} skips collision optimization, so its placements interpenetrate or fall apart under gravity. SceneSmith~\citep{pfaff2026scenesmith} emits poses via direct VLM output or hand-crafted heuristics and therefore misses local contact geometry. TabletopGen's~\citep{tabletopgen2025} per-instance segmentation either drops objects or fuses composites into a single instance. Our method produces contact-precise compositions that remain stable across every prompt.}
    \label{fig:imgcompare}
\end{figure}

\begin{wraptable}{r}{0.5\textwidth}
\centering
\scriptsize
\caption{User study pairwise win-rates. O.M. counts ties as 0.5 wins for each side; R.M. excludes ties. Wilson 95\% confidence intervals and Cohen's $h$ are reported.}
\label{tab:user_study}
\setlength{\tabcolsep}{2pt}
\renewcommand{\arraystretch}{1.4}
\resizebox{\linewidth}{!}{%
\begin{tabular}{lcccccc}
\hline
\multirow{2}{*}{\rule{0pt}{2.6ex}\textbf{Ours vs.}}
& \multicolumn{3}{c}{\rule{0pt}{2.6ex}\textbf{O.M.\ ($\uparrow$)}}
& \multicolumn{3}{c}{\rule{0pt}{2.6ex}\textbf{R.M.\ ($\uparrow$)}} \\
\cline{2-7}
& \rule{0pt}{2.4ex}Win\% & 95\% CI & $h$
& Win\% & 95\% CI & $h$ \\
\hline
SceneSmith~\citep{pfaff2026scenesmith}  & 80.7 & [75.0, 85.3] & +0.66 & 95.9 & [91.7, 98.0] & +1.16 \\
RoLA~\citep{wei2025rola}        & 78.5 & [73.0, 83.1] & +0.61 & 97.2 & [94.0, 98.7] & +1.23 \\
TabletopGen~\citep{tabletopgen2025} & 80.8 & [75.1, 85.4] & +0.66 & 95.9 & [91.7, 98.0] & +1.16 \\
GenieSim~\citep{geniesim2025}    & 98.4 & [95.8, 99.4] & +1.32 & 100.0 & [98.2, 100.0] & +1.57 \\
\hline
\end{tabular}%
}
\end{wraptable}

As shown in Tab.~\ref{tab:main_results}, under both language- and image-prompt settings, our method attains the highest R.M. in every one of the eight relation classes, averaging 7.82 against 6.33 for SceneSmith (the strongest baseline) and 1.48 for GenieSim~\citep{geniesim2025}, with C.5 (Hooking) the hardest class yet still at least 1.6 points ahead of every baseline. On O.M., we exceed SceneSmith by 0.50 on average and never trail by more than 0.13 in any class, confirming that CoGen preserves asset fidelity while improving pose accuracy. Owing to physics-guided refinement, our C.R. is near-zero. The only residual collisions appear in the spoon-hook task, where convex decomposition produces a small ring radius that EvoOpt must trade against the collision term to stay hooked.

Human raters corroborate the VLM findings with even larger margins.
Against all four baselines, scenes from our method are preferred with large Cohen's $h$ effect sizes and tight Wilson confidence intervals (Tab.~\ref{tab:user_study}). The advantage is most pronounced on R.M., where ours wins 95.9\% and 100\% against SceneSmith~\citep{pfaff2026scenesmith} and GenieSim~\citep{geniesim2025}, respectively. Visual comparisons are shown in Fig.~\ref{fig:imgcompare}.

\subsection{Ablation Studies}
\label{sec:exp:ablation}

\begin{table*}[t]
\centering
\small
\setlength{\tabcolsep}{4pt}
\caption{Ablation on complementary reconstruction (c.r.) across 8 classes.}
\label{tab:ablation_edit}
\resizebox{\textwidth}{!}{%
\begin{tabular}{lcccccccccccccccc}
\hline
\multirow{2}{*}{\rule{0pt}{2.6ex}\textbf{Method}}
& \multicolumn{2}{c}{\rule{0pt}{2.6ex}\textbf{C.1}}
& \multicolumn{2}{c}{\rule{0pt}{2.6ex}\textbf{C.2}}
& \multicolumn{2}{c}{\rule{0pt}{2.6ex}\textbf{C.3}}
& \multicolumn{2}{c}{\rule{0pt}{2.6ex}\textbf{C.4}}
& \multicolumn{2}{c}{\rule{0pt}{2.6ex}\textbf{C.5}}
& \multicolumn{2}{c}{\rule{0pt}{2.6ex}\textbf{C.6}}
& \multicolumn{2}{c}{\rule{0pt}{2.6ex}\textbf{C.7}}
& \multicolumn{2}{c}{\rule{0pt}{2.6ex}\textbf{C.8}} \\
\cline{2-17}
& \rule{0pt}{2.4ex}O.M. & R.M.
& O.M. & R.M.
& O.M. & R.M.
& O.M. & R.M.
& O.M. & R.M.
& O.M. & R.M.
& O.M. & R.M.
& O.M. & R.M. \\
\hline
w/o c.r. & 7.00 & 5.80 & 9.20 & 2.80 & 9.07 & 7.27 & 8.53 & 6.53 & 8.13 & 2.60 & 8.93 & 6.73 & 8.87 & 7.20 & 7.60 & 4.00 \\
w c.r.        & \textbf{8.93} & \textbf{7.60} & \textbf{9.27} & \textbf{9.00} & \textbf{9.33} & \textbf{8.20} & \textbf{9.13} & \textbf{8.27} & \textbf{8.80} & \textbf{5.20} & \textbf{9.00} & \textbf{7.47} & \textbf{9.53} & \textbf{7.80} & \textbf{8.80} & \textbf{9.00} \\
\hline
\end{tabular}%
}
\end{table*}

\begin{figure}[t]
    \centering
    \includegraphics[width=0.92\linewidth]{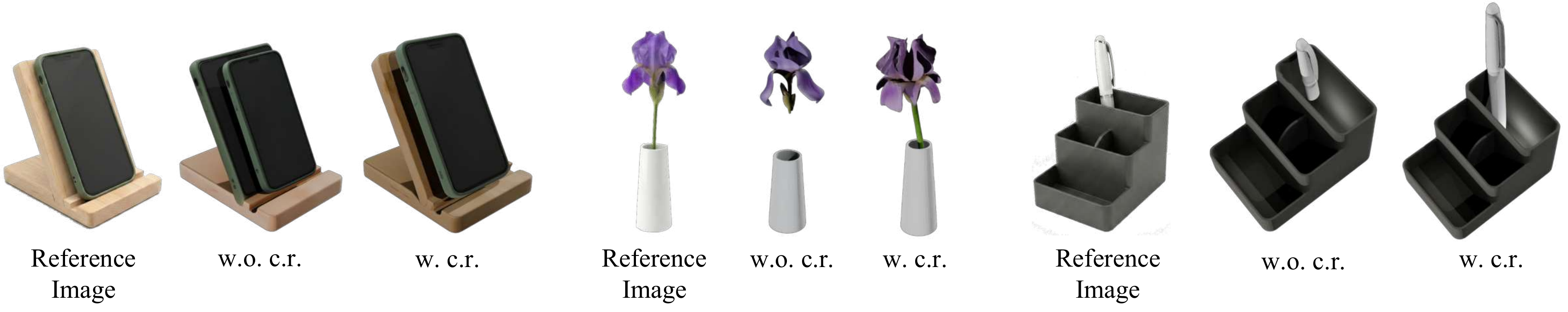}
    \caption{Three examples that show the effectiveness of CoGen. Beyond the over-segmentation case in Fig.~\ref{fig:pipeline} (pot lid mask absorbs the pot), three further failure modes drive disentangled reconstruction. Phone--phone-stand reconstruction lets the metallic phone texture leak onto the stand. Flower--vase reconstruction drops the slender stem under incomplete foreground masks, recovered by the erase-and-regenerate branch from a partner-removed view. Pen--pen-holder reconstruction truncates the pen along its hidden length under occlusion.}
    \label{fig:abcogen}
\end{figure}

\noindent\textbf{Effectiveness of CoGen.}
We ablate complementary reconstruction (c.r.) within CoGen in Tab.~\ref{tab:ablation_edit}: disabling c.r. causes a consistent drop in both O.M. and R.M. across all eight classes, as occlusion and interference from the partner object yield unreasonable assets that degrade relation matching. Visual comparisons are shown in Fig.~\ref{fig:abcogen}.

\begin{wraptable}{r}{0.5\textwidth}
\centering
\vspace{-0.8em}
\footnotesize
\setlength{\tabcolsep}{3pt}
\caption{Ablation on pose refinement modules across 8 classes, measured by R.M. . Methods are cumulative, with each row adding one stage to the configuration in the row above.}
\label{tab:ablation_pose}
\renewcommand{\arraystretch}{1.2}
\resizebox{\linewidth}{!}{%
\begin{tabular}{lcccccccc}
\hline
\textbf{Method} & \textbf{C.1} & \textbf{C.2} & \textbf{C.3} & \textbf{C.4} & \textbf{C.5} & \textbf{C.6} & \textbf{C.7} & \textbf{C.8} \\
\hline
SAM3D Only   & 5.47 & 7.07 & 7.47 & 7.60 & 4.60 & 6.13 & 6.40 & 7.47 \\
$+$ ICP      & 6.80 & 8.40 & 7.93 & 8.13 & 4.67 & 6.93 & 7.13 & 7.47 \\
$+$ RenderOpt & 7.33 & 8.93 & 8.00 & \textbf{8.27} & 4.87 & 7.33 & 7.67 & 8.93 \\
$+$ EvoOpt   & \textbf{7.60} & \textbf{9.00} & \textbf{8.20} & \textbf{8.27} & \textbf{5.20} & \textbf{7.47} & \textbf{7.80} & \textbf{9.00} \\
\hline
\end{tabular}%
}
\end{wraptable}

\noindent\textbf{Effectiveness of GPRM.}
Tab.~\ref{tab:ablation_pose} presents a cumulative ablation of GPRM stages measured by R.M. Each successive module yields consistent gains over the SAM3D initial pose, and EvoOpt provides the largest gain on C.5 (Hooking, $+$0.33) where simulation stability is most sensitive to small errors, confirming that physical simulation feedback is essential for generating such scenes.

\FloatBarrier
\subsection{\thiswork-Generated Tasks for Policy Learning}
\label{sec:exp:policy}

\begin{wrapfigure}{r}{0.46\textwidth}
\centering
\includegraphics[width=0.92\linewidth]{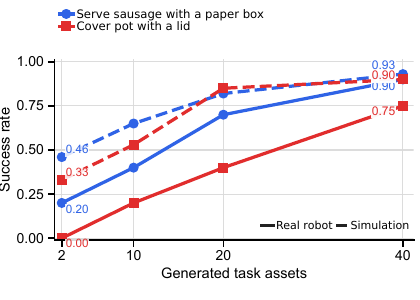}
\caption{Policy learning on two real-robot tasks. Solid lines report real-robot success and dashed lines report simulation success as the number of \thiswork{}-generated task-asset variants grows.}
\label{fig:policy_scaling}
\end{wrapfigure}

We evaluate \thiswork{}-generated compositions as policy-learning tasks. For each task, CoGen and GPRM instantiate task-asset variants, each consisting of simulation-ready assets and a VLM-verified terminal relative pose. We synthesize demonstrations by randomizing initial object poses, scene illumination, and camera poses, train $\pi_{0.5}$ only on the simulated trajectories, and deploy directly on the real robot for two contact-rich tasks: ``serve sausage with a paper box'' and ``cover pot with a lid''.

Fig.~\ref{fig:policy_scaling} shows that real-robot success improves with the asset budget: serving rises from $0.20$ with two assets to $0.90$ with forty, while covering rises from $0.00$ to $0.75$. Simulation success is higher in absolute value but follows the same monotonic trend, suggesting that \thiswork{} assets and relative poses define executable policy-learning goals and that asset diversity improves real-world transfer. Training hyperparameters, data-generation details, and more demonstrations are in the supplementary.

\subsection{Preliminary Extension to Multi-Object Scenes}
\label{sec:exp:extension}

\begin{wrapfigure}{r}{0.4\textwidth}
\centering
\vspace{-1.4em}
\includegraphics[width=\linewidth]{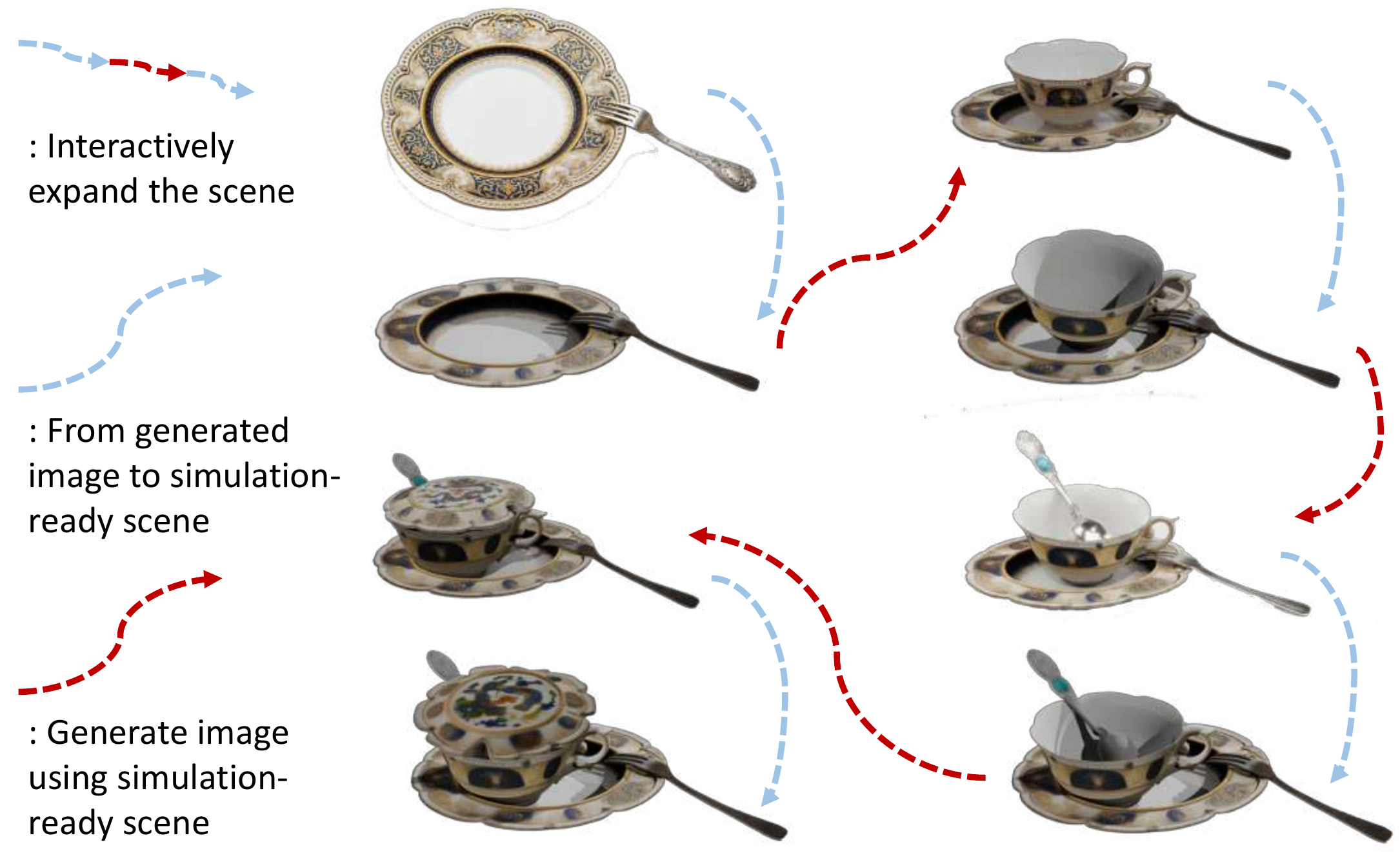}
\caption{Preliminary extension to a multi-object scene. Further results spanning all eight relation classes are in the supplementary.}
\label{fig:multiobj}
\vspace{-1em}
\end{wrapfigure}

Our pipeline extends to scenes with more objects through an iterative loop. At each step, a multimodal agent proposes the next object and its spatial relation, and the image-editor inserts it into the rendered view of the current scene while preserving viewpoint, lighting, and existing layout. CoGen and GPRM are then rerun on the newcomer, with the merged geometry of all previously placed objects held fixed during geometry- and physics-guided refinement. An example is shown in Fig.~\ref{fig:multiobj}.

\vspace{-0.8em}
\section{Limitations and Conclusion}
\label{sec:limitations}
\vspace{-0.5em}

\thiswork{} targets rigid object pairs, since convex decomposition~\citep{wei2022coacd,vhacd} and MuJoCo settling assume rigid geometry, leaving articulated and deformable assets out of scope. Room-scale layouts also remain future work; a natural direction is hierarchical decomposition, as in SceneSmith~\citep{pfaff2026scenesmith}, with \thiswork{} serving as the contact-precise leaf solver. In summary, \thiswork{} casts contact-precise scene generation as reconstruction plus relative-pose recovery, coupling CoGen for instance-disentangled meshes, GPRM for pose refinement, and a VLM verifier. Across eight contact-geometry classes, \thiswork{} outperforms strong baselines on relationship matching, object fidelity, collision rate, and human preference. Its simulation data generation, policy training, and real-world deployment further demonstrate that verified functional compositions can support downstream robot learning.

\relax

\bibliography{main}

\clearpage
\appendix
\setcounter{section}{0}
\setcounter{subsection}{0}
\setcounter{figure}{0}
\setcounter{table}{0}
\setcounter{equation}{0}
\renewcommand{\thesection}{\Alph{section}}
\renewcommand{\thesubsection}{\Alph{section}.\arabic{subsection}}
\renewcommand{\thefigure}{S\arabic{figure}}
\renewcommand{\thetable}{S\arabic{table}}
\renewcommand{\theequation}{S\arabic{equation}}

\begin{center}
{\LARGE\bf Supplementary Materials}\\[0.5em]
{\large GIF: Agentic Generation of Interactive and Functional Object Compositions for Robot Learning}
\end{center}
\vspace{1em}

\section{Methodology Details}
\label{sec:supp:method}

This section provides implementation details for the modules introduced in Sec.~\ref{sec:method}, while keeping the presentation aligned with the pipeline abstraction used in the main paper. We follow the same order as the main text: multimodal input canonicalization, complementary 2D--3D asset construction (CoGen), geometry- and physics-guided pose refinement (GPRM), VLM-based scene selection, and policy data synthesis for VLA training.

\subsection{Implementation Backends}
\label{sec:supp:method:backends}

Tab.~\ref{tab:supp:backends} summarizes the model and software backends used by the agentic pipeline. The prompt templates in Sec.~\ref{sec:supp:prompts} define the task-level behavior of the LLM/VLM calls; the table below specifies the underlying modules.

\begin{table}[H]
\centering
\caption{Implementation backends used in \thiswork{}.}
\label{tab:supp:backends}
\small
\begin{tabular}{p{0.34\linewidth}p{0.58\linewidth}}
\toprule
Component & Backend\\
\midrule
Pure-text LLM calls & gpt-4o-mini\\
VLM verifier, candidate voter, and VLM scene scorer & qwen3-vl-plus~\citep{qwen3vl}\\
Reference-image generation and image editing & gemini-3.1-flash-image-preview~\citep{gemini31flashimage} for input canonicalization, reference-image synthesis, and the CoGen erase branch\\
Instance-mask preparation & SAM3~\citep{sam3} binary object masks derived from the canonical scene specification and reference image, then used as SAM3D conditioning masks\\
Image-to-3D lifting & SAM3D~\citep{sam3d}\\
Monocular geometry for ICP targets & MoGeV2~\citep{mogev2}\\
Convex decomposition & CoACD~\citep{wei2022coacd}\\
Differentiable rendering & nvdiffrast~\citep{nvdiffrast}\\
Collision optimization and solver & Drake~\citep{drake} with SNOPT~\citep{snopt}\\
Physics and simulation-aware search & MuJoCo~\citep{todorov2012mujoco} with CMA-ES from EvoTorch~\citep{cmase,evotorch}\\
Policy-data trajectory engine & GraspSimulator with CuRobo planning and Isaac Sim ray-traced rendering\\
\bottomrule
\end{tabular}
\end{table}

\subsection{Multimodal Input Canonicalization}
\label{sec:supp:method:input}

The multimodal input agent in Sec.~\ref{sec:method:input} is implemented as a unified goal-conditioned loop, rather than as a set of modality-specific branches. It accepts a task instruction, a reference scene image, optional object-level asset images, or any combination of these inputs, and returns the canonical pair $(I_{\rm ref},J_{\rm scene})$. Here $I_{\rm ref}$ is the foreground-only reference image used by CoGen, and $J_{\rm scene}$ records the objects, their attributes, and the desired relation.

\paragraph{Tools and state.}
The agent uses two image operations: prompt-conditioned generation for text-only or asset-guided inputs, and instruction-conditioned editing when a usable scene image is already available. Both operations call the same image-generation and image-editing backends used later in CoGen, which keeps the visual distribution of $I_{\rm ref}$ consistent across the pipeline. The agent also maintains the current candidate image across iterations, enabling corrective edits to build on the previous result instead of reinitializing the scene.

\paragraph{Analyze, act, verify.}
The prompt in Sec.~\ref{sec:supp:prompt:input_scene_agent} specifies the admissible relation vocabulary, scene-quality constraints, and output schema. Under this prompt, the agent repeatedly performs three operations: it first analyzes the available modalities to identify the objects, and relation; it then generates or edits a candidate reference image, using any supplied images as appearance references; finally, it verifies the candidate against constraints on background simplicity, object count, visibility, contact plausibility, camera viewpoint, physical plausibility, and rigid-body separability. Failed candidates are repaired through targeted edits whenever possible, which helps preserve the user-specified objects while correcting localized violations.

\paragraph{Asset GLB pre-rendering.}
If the input includes a 3D asset rather than an RGB image, we first render it from a perspective viewpoint on a plain background and pass the rendered view to the same analyzer. This preprocessing step makes mesh and image inputs share the same visual interface, so the canonicalization prompt and verification rules do not depend on the original asset format.

\paragraph{Convex decomposition for simulation-ready assets.}
Once a reference image is accepted, the object entries in $J_{\rm scene}$ are used to prepare SAM3~\citep{sam3} binary instance masks $b_i^{\rm ref}$ for the target objects. Each cleaned object image is lifted by SAM3D~\citep{sam3d} into a textured visual mesh and an initial camera-frame transform. For simulation, we separately convert the visual mesh into a set of convex collision parts with CoACD~\citep{wei2022coacd}; the textured mesh is retained for rendering, while the convex decomposition is used for stable contact simulation in MuJoCo~\citep{todorov2012mujoco}. This separation between visual and collision geometry is used throughout the downstream pose pipeline.

\subsection{CoGen: Complementary 2D-3D Generative Asset Construction}
\label{sec:supp:method:cogen}

CoGen converts the canonical pair $(I_{\rm ref},J_{\rm scene})$ into two mesh--pose candidates per object: a direct reconstruction $(m_i^{\rm ref},p_i^{\rm ref})$ from the original pair image and a partner-removed reconstruction $(m_i^{\rm erase},p_i^{\rm erase})$. GPRM later refines all cross-branch pairings, and the VLM selector chooses among the physically settled candidates. We highlight two details that make this complementary construction robust on long-tail contact pairs.

\paragraph{Synonym-driven retry for mask preparation.}
Mask preparation is sensitive to the surface form of an object name: a phrase such as ``cup-rack tree'' may fail even when visually equivalent prompts such as ``mug rack'' or ``cup tree'' succeed. We therefore add a synonym-driven retry stage. When the first mask-preparation attempt is empty or low confidence, an LLM proposes five alternative noun phrases using the prompt in Sec.~\ref{sec:supp:prompt:synonyms}. The alternatives are tried in rank order until a mask passes basic area and shape checks. If all phrasings fail, the branch falls back to a coarse spatial prompt on $I_{\rm ref}$. This mechanism improves coverage of long-tail household objects without introducing a hand-written vocabulary.

\paragraph{Erase-and-regenerate as a closed-loop agent.}
The erase branch constructs $I_i^{\rm erase}$ by removing the partner object $o_j$ while preserving the geometry and pose of $o_i$. A single inpainting call is often insufficient for contact-rich pairs: it can shift the preserved object, leave traces of the erased object, or introduce background artefacts near the contact region. We instead use a closed-loop erase agent that alternates between image editing and visual verification. The verifier compares the edited image with the original and assigns a score under a pose-first rubric: pose drift of the preserved object receives the largest penalty, followed by incomplete removal, appearance changes, and background artefacts. The loop returns the first image above the acceptance threshold, or the best-scored image if the iteration budget is exhausted. The operative erase prompt is listed in Sec.~\ref{sec:supp:prompt:erase_agent}, and the scoring weights are summarized in Tab.~\ref{tab:supp:erase_rubric}.

\begin{table}[h]
\centering
\caption{Score deductions used by the erase agent's verification rubric. Pose drift dominates because subsequent ICP and render-based optimization can absorb mild appearance change but not a shifted partner object.}
\label{tab:supp:erase_rubric}
\renewcommand{\arraystretch}{1.05}
\small
\begin{tabular}{lr}
\toprule
Failure mode & Deduction\\
\midrule
Preserved object pose changed (position, scale, or orientation differs) & $-50$\\
Target object still partially visible after erase & $-25$\\
Preserved object appearance changed while pose intact & $-15$\\
Background artefacts or unnatural fill & $-10$\\
\bottomrule
\end{tabular}
\end{table}

\paragraph{Image-to-3D lifting and pose readout.}
Both branches use the same image-to-3D model to predict a textured mesh and a coarse camera-frame transform. The transform initializes $p_i^{a_i}$ for GPRM, while the mesh is converted into the visual and collision geometries described above. The reference branch conditions reconstruction on $I_{\rm ref}$ and the instance mask, whereas the erase branch reconstructs from the partner-removed image $I_i^{\rm erase}$. Because the two branches fail in different ways, the four cross-branch pairings $\mathcal{C}=\{(a_1,a_2)\}$ provide meaningful alternatives for downstream geometric and physical selection.

\subsection{GPRM Optimization Details}
\label{sec:supp:method:gprm}

This section gives the optimization objectives underlying the GPRM stages summarized in Sec.~\ref{sec:method:pose}.

\paragraph{ICP target cloud.}
For each object $o_i$ we lift the reference image $I_{\rm ref}$ into a metric depth map $d\in\mathbb{R}_+^{H\times W}$ with a monocular geometry estimator~\citep{mogev2}, and back-project it through the camera intrinsics $K$ to a scene cloud $\mathcal{P}=\{K^{-1}[u\,d_{uv},\,v\,d_{uv},\,d_{uv}]^\top\mid d_{uv}>0\}$. The per-object target cloud $\mathcal{P}_i^{\rm tgt}=\{p\in\mathcal{P}\mid (u,v)\in b_i^{ref,\rm erode}\}$ is obtained by sampling $\mathcal{P}$ inside an eroded instance mask. The source cloud is sampled from the candidate mesh surface $m_i^{a_i}$ at pose $p_i^{a_i}$. We then run progressively tighter ICP rounds with shrinking distance thresholds, yielding the refined transform $p_i^{a_i,\rm icp}$.

\paragraph{RenderOpt objective.}
Given $(m_i^{a_i},p_i^{a_i})$, we rasterise the mesh with nvdiffrast~\citep{nvdiffrast} to obtain the per-object soft mask $\hat{b}_i\in[0,1]^{H\times W}$ and depth $\hat{d}_i\in\mathbb{R}_+^{H\times W}$, together with their composited counterparts $\hat{b}$ and $\hat{d}$ across all objects. The objective is
\begin{align}
\mathcal{L}_{\rm rend} &=
  \underbrace{\mathcal{L}_{\rm BCE}(\hat{b},b^{ref}) + \mathcal{L}_{\rm L1}(\hat{d},d)}_{\text{global}}+\underbrace{\lambda_p\mathcal{L}_{\rm pose\text{-}reg} + \lambda_s\mathcal{L}_{\rm scale\text{-}reg}}_{\text{regularization}}
\nonumber\\
&\quad+\sum_{i}\Bigl[
  \mathcal{L}_{\rm recall}(\hat{b}_i,b_i^{ref}) + \mathcal{L}_{\rm prec}(\hat{b}_i,b^{ref})
\nonumber\\
&\qquad\qquad + \mathcal{L}_{\rm d\text{-}self}(\hat{d}_i,d,b_i^{ref}) + \mathcal{L}_{\rm d\text{-}occ}(\hat{d}_i,d,b_i^{ref})
\Bigr]_{\text{per-object}},
\label{eq:supp:rend_loss}
\end{align}
where $b^{ref}=\bigcup_i b_i^{ref}$ and $\mathcal{L}_{\rm BCE}$ is the focal variant~\citep{focal}. The per-object terms enforce that each rendered mask covers its target ($\mathcal{L}_{\rm recall}$), does not spill beyond valid image regions ($\mathcal{L}_{\rm prec}$), matches observed depth within the object region ($\mathcal{L}_{\rm d\text{-}self}$), and does not incorrectly occlude other objects ($\mathcal{L}_{\rm d\text{-}occ}$). The regularizers anchor the pose and scale to the ICP initialization. The optimized pose $p_i^{a_i,\rm rend}$ is accepted only if all four per-object loss components improve, which guards against degenerate optima.

\paragraph{ColliOpt formulation.}
Let $\mathrm{cvx}(m_i,p_i)$ denote the convex decomposition of $m_i$ at pose $p_i$. Starting from $p_i^{\rm rend}$, ColliOpt seeks the nearest collision-free pose by solving
\begin{align}
&\min_{\{p_i\}}\;\sum_{i}\!\left[\bigl(r_i\ominus r_i^{\rm rend}\bigr)^\top\!(r_i\ominus r_i^{\rm rend}) + \|t_i - t_i^{\rm rend}\|^2\right]\nonumber\\
&\quad\text{s.t.}\quad d_{\rm min}\!\left(\mathrm{cvx}(m_i,p_i),\,\mathrm{cvx}(m_j,p_j)\right) \geq -\varepsilon,\quad \forall i\neq j,
\label{eq:supp:drake_opt}
\end{align}
where $r_i\ominus r_i^{\rm rend}$ is the geodesic distance in $SO(3)$, and $d_{\rm min}$ is the minimum signed separation over every cross-object pair of convex-decomposition pieces; it is not the distance between whole-object convex hulls. The tolerance $\varepsilon$ permits small residual penetration. We further add a ground-clearance constraint $t_{i,z}\geq z_{\rm floor}+\varepsilon$ to prevent objects from sinking below the floor. The problem is built with Drake~\citep{drake} and solved with SNOPT~\citep{snopt}.

\paragraph{EvoOpt fitness.}
EvoOpt searches in the neighbourhood of $p_i^{a_i,\rm rend}$ rather than $p_i^{a_i,\rm coll}$, since the latter may already be at a bad local minimum. Each CMA-ES~\citep{cmase} candidate is a perturbation $x_i\in\mathbb{R}^9$ per body, encoding a translation delta in $\mathbb{R}^3$ and a rotation delta in the 6D representation~\citep{continuity}. Denote the perturbed pose $\tilde{p}_i^{a_i}(x_i)$. We score it by loading the perturbed MuJoCo~\citep{todorov2012mujoco} scene and running $T_{\rm sim}$ physics steps,
\begin{equation}
f(x_i) = \underbrace{\sum_{t=1}^{T_{\rm sim}}\Bigl[w_{q}\|q_t - q_{t-1}\|}_{\text{drift}} + \underbrace{w_{\rm col}\!\sum_{c_{t}}\max\!\left(0,\,-d_{c_{t}}\right)}_{\text{collision}} + \underbrace{w_v\|v_t\|}_{\text{velocity}}\Bigr] + \underbrace{w_e\|\tilde{p}_i^{a_i}(x_i) - p_i^{a_i,\rm rend}\|}_{\text{effort}},
\label{eq:supp:evo_fitness}
\end{equation}
where $q_t$ is the generalized position at step $t$, $d_{c_t}$ is the signed penetration depth of contact pair $c_t$, and $v_t$ is the generalized velocity. The four terms penalize dynamic drift, residual interpenetration, residual kinetic energy, and deviation from the input pose. EvoOpt is launched only when the VLM selector rejects all candidates produced by the earlier three refinements.

\subsection{Cross-Candidate Voting Example}
\label{sec:supp:method:voter}

\begin{figure}[H]
    \centering
    \includegraphics[width=\linewidth]{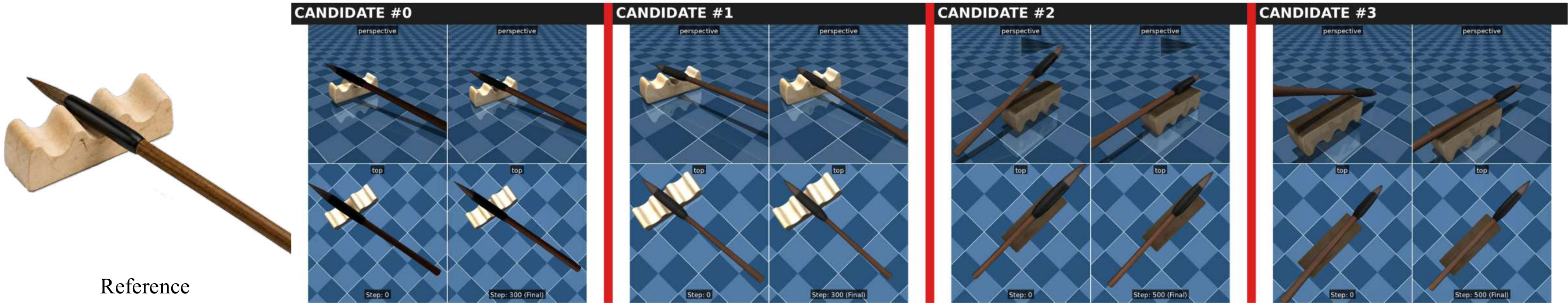}
    \caption{Cross-candidate voting. Per-candidate timeline images are tiled into a single montage with bold labels and red separators. The VLM selects the best candidate based on $J_{\rm scene}$. In each panel, left or right columns show the initial or the gravity-settled final states of the simulation. Top and bottom rows show perspective and top-down views, respectively.}
    \label{fig:supp:voter_montage}
\end{figure}

Sec.~\ref{sec:method:voter} describes a two-stage VLM selector that first verifies each refined scene in isolation and then ranks the surviving candidates jointly. Fig.~\ref{fig:supp:voter_montage} visualizes a representative input to the cross-candidate stage, where each panel is itself a single-scene verification image. Detailed prompts for the verifier and the voter are provided in Sec.~\ref{sec:supp:prompt:verify_multi} and Sec.~\ref{sec:supp:prompt:voter}.

\subsection{Policy Data Synthesis and VLA Training}
\label{sec:supp:method:policy}

\paragraph{Trajectory synthesis.}
For each VLM-verified terminal composition $\widehat{\mathcal{S}}=\{(m_i,p_i)\}_{i=1}^{2}$, we generate demonstrations by randomizing an initial tabletop scene and planning a trajectory that realizes the verified relative pose. Let $m_1$ be the manipulated object and $m_2$ be the reference object. The scene-generation pipeline defines the target relative transform $\Delta p=p_2^{-1}p_1$. For each rollout, we sample both the initial manipulated-object pose $p_1^{\rm rand}$ and the reference-object pose $p_2^{\rm rand}$. The reference object is placed at this sampled pose and is not manipulated by the planner; the desired terminal pose of $m_1$ is obtained by transporting $\Delta p$ into the randomized reference-object frame:
\begin{equation}
    p_2^{\star}=p_2^{\rm rand}, \qquad
    p_1^{\star}=p_2^{\rm rand}\Delta p
    =p_2^{\rm rand}p_2^{-1}p_1 .
\end{equation}
The planner then produces a collision-free pick-and-place trajectory from $p_1^{\rm rand}$ to $p_1^{\star}$ with approach, grasp, lift, transfer, place, and retreat phases. Failed plans, unstable terminal states, and executions that break the target relation after MuJoCo settling are discarded before policy training.

\paragraph{Symmetry-aware target sampling.}
Many functional relations admit equivalent terminal poses; for example, a cylindrical object inserted into a container can often rotate about its long axis without changing the task outcome. We therefore run the symmetry analyzer of Sec.~\ref{sec:supp:prompt:symmetry} on each task object and obtain a task-conditioned symmetry group $G_i$ over the admissible axes. During trajectory synthesis, the target pose is treated as an equivalence class $\{p_i^\star g\mid g\in G_i\}$ rather than a single rigid transform. For finite cyclic symmetries ($C_2,C_3,C_4$), we sample one of the admissible rotations; for continuous symmetry ($C_{\infty}$), we sample a rotation angle uniformly around the accepted axis. This avoids penalizing functionally identical terminal states and reduces redundant supervision in the generated trajectories.

\paragraph{Policy architecture.}
We use the $\pi_{0.5}$ VLA architecture~\citep{pi05} as the policy template, but replace its original VLM backbone with Qwen3-VL-2B-Instruct; references to $\pi_{0.5}$ in this section therefore denote the architecture template rather than the original released checkpoint. The policy uses a flow-matching action expert for continuous action prediction. Training starts from the pretrained VLM weights, while the action expert is initialized from scratch. The vision encoder and visual merger are frozen, while the language backbone and action expert are trainable. The model observes one third-person main camera and one wrist camera, both resized to $224\times224$, together with robot proprioception. Actions and proprioception are represented in 7 dimensions, the proprioceptive history length is 2, the action chunk length is 4, and the control interval is $\Delta t=0.3$s.

\begin{table}[H]
\centering
\caption{VLA policy training configuration used for the real-robot policy-learning experiments.}
\label{tab:supp:vla_training}
\small
\begin{tabular}{p{0.34\linewidth}p{0.58\linewidth}}
\toprule
Item & Setting\\
\midrule
Policy template & $\pi_{0.5}$ architecture with Qwen3-VL-2B-Instruct VLM backbone\\
Prediction head & Flow-matching action expert, 256 action tokens\\
Trainable modules & Language backbone and action expert\\
Frozen modules & Vision encoder and visual merger\\
Training data & $50{,}000$ simulated trajectories per policy-learning task\\
Training steps & $50{,}000$ optimizer steps from pretrained VLM weights and a randomly initialized action expert\\
Camera inputs & Third-person main view and wrist view\\
Image resolution & $224\times224$\\
Action / proprioception dimension & 7 / 7\\
Proprioception history / action chunk & 2 steps / 4 steps\\
Control interval & $0.3$s\\
Optimizer & AdamW, learning rate $1.6\times10^{-4}$, constant schedule\\
Regularization & Weight decay 0, gradient norm clipped to 1.0\\
Precision and parallelism & bfloat16, FSDP full-shard\\
Batch size & 24 per GPU, 384 global\\
Hardware & 2 nodes, 8 H100 GPUs per node\\
\bottomrule
\end{tabular}
\end{table}

\paragraph{Policy-scaling protocol.}
For the asset-scaling experiment in Fig.~\ref{fig:policy_scaling}, the number of training trajectories is held fixed across asset budgets: every policy is trained with $50{,}000$ simulated trajectories for the corresponding task. The asset budget controls how many \thiswork{}-generated task-asset variants are used to synthesize those trajectories. Simulation success is evaluated on newly generated held-out assets that are not used in policy training, with 80 rollouts per task and asset budget. The simulator uses the same programmatic success predicates as data filtering: the sausage task requires the sausage's projected body to lie inside the paper box after release, and the pot-lid task requires the lid to be released, approximately horizontal, centered over the pot opening, overlapping the container footprint, and supported near the pot rim without sinking into the pot.

\paragraph{Simulation randomization.}
During data generation we randomize the initial object poses, distractor objects, table geometry, scene illumination, camera poses, and visual materials so that the policy sees a broad distribution around each verified terminal relation. The final rendered observations use the same two-view convention as the real robot: one third-person front camera and one wrist-mounted camera. Tab.~\ref{tab:supp:data_generation} summarizes the settings that materially affect the generated demonstrations; low-level engineering parameters such as worker count, queue size, and batching are omitted because they affect throughput rather than the data distribution.

\begin{table}[H]
\centering
\caption{Trajectory-generation and rendering settings for policy-learning data.}
\label{tab:supp:data_generation}
\small
\begin{tabular}{p{0.30\linewidth}p{0.62\linewidth}}
\toprule
Item & Setting\\
\midrule
Trajectory engine & GraspSimulator with MuJoCo physics, CuRobo motion planning, and Isaac Sim ray-traced rendering\\
Robot embodiment & Franka arm with a parallel gripper\\
Task specification & Object identities, the manipulated object, the reference object, and the \thiswork{}-verified terminal relative transform used to compute $p_1^\star$ for each sampled reference-object pose\\
Scene content & Target object pair plus sampled distractors, with 2--8 objects total per scene\\
Execution phases & Approach, grasp, lift, transfer, place or lower, release, and retreat; unreachable grasps, failed plans, unstable rollouts, and relation failures are discarded\\
Camera views & One randomized third-person front view and one randomized wrist-mounted view. The third-person camera perturbs azimuth, elevation, distance, look-at point, and roll around the workspace, while the wrist camera perturbs its mounting offset and orientation around the end effector. Samples that fail to keep the task objects and gripper visible are rejected\\
Rendered resolution & $256\times256$ during data generation; images are resized to $224\times224$ for VLA training\\
Lighting & Randomized area-light color temperature, intensity, size, position, and orientation\\
Visual randomization & Randomized table, ground, background, robot, and object material parameters, with additional procedural texture variation for rendered backgrounds\\
Data-generation hardware & RTX 4090 GPU workers\\
\bottomrule
\end{tabular}
\end{table}

\paragraph{Real-robot evaluation protocol.}
The real-robot setup uses the same observation interface as the simulation data: one fixed third-person main camera, one wrist camera, and robot proprioception. The sim-trained policy is deployed directly on the robot without real-world fine-tuning. Tab.~\ref{tab:supp:real_robot_setup} summarizes the evaluation setup and the initial-state protocol.

\begin{table}[H]
\centering
\caption{Real-robot evaluation setup.}
\label{tab:supp:real_robot_setup}
\small
\begin{tabular}{p{0.34\linewidth}p{0.58\linewidth}}
\toprule
Item & Setting\\
\midrule
Robot platform & Franka arm with a parallel gripper\\
Observation inputs & Intel RealSense fixed third-person main camera, ZED wrist-mounted camera, and robot proprioception\\
Main-camera placement & RealSense camera placed opposite the robot, within a $0$--$40^{\circ}$ horizontal azimuth range around the robot-forward direction, matching the simulation camera-randomization envelope\\
Policy interface & Same as simulation training: two $224\times224$ RGB views, 7-D proprioception, 7-D actions, action chunk length 4, and control interval $\Delta t=0.3$s\\
Deployment & Simulation-trained policy evaluated directly on the real robot, without real-robot fine-tuning\\
Tasks & Serving a sausage with a paper box; covering a pot with a lid\\
Initial object layouts & Target objects are randomly placed on the tabletop within the Franka reachable workspace. For each task and asset-budget condition, success is computed from 10 fixed initial layouts with three repeats per layout\\
Distractors & 1--3 distractor objects per trial\\
Trials & 30 real-robot trials per task and asset-budget condition, corresponding to 10 layouts $\times$ 3 repeats\\
\bottomrule
\end{tabular}
\end{table}

For the sausage task, a trial is successful if the robot places the sausage into the paper box and the sausage remains inside without falling out; at most two repeated grasp attempts on the sausage are allowed within a trial. For the pot-lid task, a trial is successful if the robot grasps the lid within at most two attempts, places it accurately on the pot rim so that it covers the opening, and leaves it stably supported by the pot without dropping it, falling outside the pot, or sinking into the pot interior.

\section{Experimental Design Details}
\label{sec:supp:exp}

This section gives the experimental details behind Sec.~\ref{sec:exp}: how the benchmark is derived from egocentric manipulation data, how the automatic and human metrics are computed, how each baseline is adapted to the benchmark, and how the preliminary multi-object extension is evaluated.

\subsection{Benchmark Scene Setup}
\label{sec:supp:exp:benchmark}

\paragraph{Source corpus and extraction.}
We derive the contact-geometry taxonomy from Ego4D~\citep{grauman2022ego4d}, using both free-form camera-wearer narrations and the structured FHO action annotations. From the narration stream, we retain camera-wearer utterances marked as action descriptions and keep their video identifiers, timestamps, and pass indices. From the FHO stream, we retain valid narrated actions together with their normalized verb, free-form verb, state transition, and temporal extent. The two sources are deduplicated at the annotation level and stored in a unified corpus for downstream filtering. This preprocessing yields millions of narration-level manipulation descriptions and roughly $0.6$M valid FHO actions across thousands of videos.

\paragraph{Filtering for two-object contact events.}
The raw corpus includes many activities outside the scope of pairwise object interaction, such as navigation, conversation, and whole-body motion. We filter it in two stages. First, we keep only hand-object interaction verbs derived from the FHO verb taxonomy. Second, we remove descriptions that do not name a concrete manipulated object or that involve more than one explicit contact partner. The remaining examples describe one target object being manipulated with respect to one contact object, matching the pairwise scene specification used by \thiswork{}. We draw a random subset of $50{,}000$ filtered examples for taxonomy construction.

\paragraph{LLM-based clustering into eight contact-geometry classes.}
An LLM-based agent clusters the filtered examples by contact geometry. To reduce sensitivity to wording, the agent first summarizes each shard into candidate relation phrases, then merges equivalent phrases across shards, and finally clusters the merged phrases according to geometric contact pattern rather than linguistic surface form. The clustering prompt requires a mutually exclusive and exhaustive partition with six to ten classes. The stable solution contains eight classes: C.1 Containment, C.2 Capping, C.3 Resting, C.4 Leaning, C.5 Hooking, C.6 Slotting, C.7 Spanning, and C.8 Pegging. The prompts for shard summarization, phrase merging, and final clustering are listed in Sec.~\ref{sec:supp:prompt:taxonomy}.

\paragraph{Per-class task selection and variant generation.}
For each class, the agent selects three representative tasks subject to three constraints: the task must instantiate the class with two distinct everyday objects, the two objects must be separable rigid bodies, and the selected tasks should cover a mixture of household and workplace scenarios. This produces the $8\times3=24$ task suite. We then instantiate each task with five semantic variants that change material, color, or substyle while preserving the same relation, yielding $24\times5=120$ benchmark scenes. Sec.~\ref{sec:supp:tasks} lists the full task descriptions and shows the corresponding reference-image galleries.

\subsection{Metrics}
\label{sec:supp:exp:metrics}

\paragraph{Prompt-conditioned scoring protocol.}
For language-conditioned methods, the scorer receives the task description and the gravity-settled MuJoCo render only; GIF's intermediate canonical image is never provided. O.M. therefore measures whether the generated objects are semantically and visually consistent with the instruction, without penalizing valid appearance differences from an unobserved reference. For image-conditioned methods, the scorer additionally receives the input reference image and evaluates O.M. against it. R.M. is judged from the named relation and settled configuration in both settings. The two operative scorer prompts are provided in Sec.~\ref{sec:supp:prompt:score}.

\subsubsection{VLM-based scoring}
\label{sec:supp:exp:vlm_scoring}

For each generated scene, we render the gravity-settled MuJoCo state from a fixed perspective view and provide the render and task description to the VLM scorer in Sec.~\ref{sec:supp:prompt:score}; only the image-conditioned protocol additionally receives the input reference image. The scorer returns integer Object Matching (O.M.) and Relationship Matching (R.M.) scores in $[0,10]$. O.M. measures instruction-consistent object appearance for language inputs and reference-image fidelity for image inputs. R.M. evaluates whether the target relation is realized after settling, using a checklist covering contact geometry, common-sense placement, interpenetration, and physical plausibility. Scores are averaged over the five variants of each task and then aggregated by relation class.

\subsection{Confidence Intervals for Automatic Metrics}
\label{sec:supp:exp:ci}
The following tables transcribe the 95\% confidence intervals stored in the supplementary data workbooks. The main-results table reports O.M. and R.M. intervals for both prompt modalities; the subsequent tables report intervals for the cumulative GPRM ablation and the CoGen complementary-reconstruction ablation.
\begin{table}[H]
\centering
\caption{Per-class main results with confidence intervals. Each entry reports the mean followed by the 95\% confidence interval in brackets.}
\label{tab:supp:main_results_ci}
\scriptsize
\setlength{\tabcolsep}{2.5pt}
\resizebox{\linewidth}{!}{%
\begin{tabular}{lcccccccc}
\toprule
Method & C.1 & C.2 & C.3 & C.4 & C.5 & C.6 & C.7 & C.8 \\
\midrule
\multicolumn{9}{l}{\textit{Image prompt: O.M. $\uparrow$}} \\
RoLA & 8 [5.48, 9.3] & 8.6 [6.14, 9.6] & 8.33 [5.84, 9.47] & 8.47 [5.99, 9.53] & 8 [5.48, 9.3] & 7.4 [4.87, 8.95] & 7.87 [5.34, 9.22] & 8.2 [5.69, 9.4] \\
TabletopGen & 6.6 [4.11, 8.44] & 9.13 [6.77, 9.82] & 9.13 [6.77, 9.82] & 5.33 [3.01, 7.52] & 7.07 [4.55, 8.74] & 7.4 [4.87, 8.95] & 8.6 [6.14, 9.6] & 7.6 [5.07, 9.07] \\
Ours-Img & 9.07 [6.68, 9.79] & 9.6 [7.37, 9.95] & 9.53 [7.28, 9.94] & 9.67 [7.47, 9.97] & 9 [6.6, 9.77] & 9.4 [7.1, 9.9] & 9.6 [7.37, 9.95] & 8.73 [6.29, 9.66] \\
\midrule
\multicolumn{9}{l}{\textit{Image prompt: R.M. $\uparrow$}} \\
RoLA & 3.8 [1.84, 6.25] & 5.07 [2.8, 7.31] & 4.33 [2.23, 6.71] & 3.07 [1.34, 5.58] & 2.8 [1.17, 5.32] & 3.33 [1.52, 5.83] & 4.47 [2.33, 6.82] & 6.2 [3.75, 8.16] \\
TabletopGen & 1.33 [0.374, 3.79] & 8.67 [6.21, 9.63] & 3.73 [1.79, 6.19] & 2.07 [0.741, 4.59] & 1.73 [0.565, 4.23] & 3.6 [1.7, 6.07] & 5 [2.74, 7.26] & 5.27 [2.96, 7.47] \\
Ours-Img & 7.8 [5.27, 9.18] & 9.2 [6.85, 9.84] & 9.27 [6.93, 9.86] & 9.47 [7.19, 9.92] & 4.47 [2.33, 6.82] & 7.33 [4.8, 8.91] & 7.53 [5.0, 9.03] & 7.53 [5.0, 9.03] \\
\midrule
\multicolumn{9}{l}{\textit{Language prompt: O.M. $\uparrow$}} \\
GenieSim & 4.87 [2.64, 7.15] & 4.4 [2.28, 6.77] & 4.07 [2.03, 6.48] & 2.73 [1.13, 5.26] & 2.53 [1.01, 5.06] & 6.07 [3.63, 8.07] & 3.07 [1.34, 5.58] & 2.87 [1.21, 5.39] \\
SceneSmith & 8.93 [6.52, 9.74] & 9.4 [7.1, 9.9] & 8.67 [6.21, 9.63] & 8.27 [5.77, 9.44] & 7.2 [4.68, 8.83] & 8.93 [6.52, 9.74] & 9.47 [7.19, 9.92] & 7.93 [5.41, 9.26] \\
Ours & 8.93 [6.52, 9.74] & 9.27 [6.93, 9.86] & 9.33 [7.02, 9.88] & 9.13 [6.77, 9.82] & 8.8 [6.37, 9.68] & 9 [6.6, 9.77] & 9.53 [7.28, 9.94] & 8.8 [6.37, 9.68] \\
\midrule
\multicolumn{9}{l}{\textit{Language prompt: R.M. $\uparrow$}} \\
GenieSim & 3.47 [1.61, 5.95] & 1.07 [0.261, 3.48] & 2.13 [0.778, 4.66] & 0.8 [0.162, 3.15] & 0.867 [0.185, 3.23] & 2.13 [0.778, 4.66] & 1 [0.234, 3.4] & 0.333 [0.0349, 2.53] \\
SceneSmith & 6.6 [4.11, 8.44] & 7.6 [5.07, 9.07] & 6.93 [4.42, 8.66] & 6.33 [3.87, 8.25] & 2.27 [0.853, 4.79] & 6.8 [4.29, 8.57] & 6.67 [4.17, 8.48] & 7.47 [4.94, 8.99] \\
Ours & 7.6 [5.07, 9.07] & 9 [6.6, 9.77] & 8.2 [5.69, 9.4] & 8.27 [5.77, 9.44] & 5.2 [2.9, 7.42] & 7.47 [4.94, 8.99] & 7.8 [5.27, 9.18] & 9 [6.6, 9.77] \\
\bottomrule
\end{tabular}%
}
\end{table}

\begin{table}[H]
\centering
\caption{GPRM cumulative ablation with confidence intervals. Each R.M. entry reports the mean followed by the 95\% confidence interval in brackets.}
\label{tab:supp:ablation_pose_ci}
\scriptsize
\setlength{\tabcolsep}{2.5pt}
\resizebox{\linewidth}{!}{%
\begin{tabular}{lcccccccc}
\toprule
Method & C.1 & C.2 & C.3 & C.4 & C.5 & C.6 & C.7 & C.8 \\
\midrule
SAM3D only & 5.47 [3.12, 7.62] & 7.07 [4.55, 8.74] & 7.47 [4.94, 8.99] & 7.6 [5.07, 9.07] & 4.6 [2.43, 6.93] & 6.13 [3.69, 8.11] & 6.4 [3.93, 8.3] & 7.47 [4.94, 8.99] \\
+ ICP & 6.8 [4.29, 8.57] & 8.4 [5.91, 9.5] & 7.93 [5.41, 9.26] & 8.13 [5.62, 9.37] & 4.67 [2.48, 6.99] & 6.93 [4.42, 8.66] & 7.13 [4.61, 8.79] & 7.47 [4.94, 8.99] \\
+ RenderOpt & 7.33 [4.8, 8.91] & 8.93 [6.52, 9.74] & 8 [5.48, 9.3] & 8.27 [5.77, 9.44] & 4.87 [2.64, 7.15] & 7.33 [4.8, 8.91] & 7.67 [5.14, 9.11] & 8.93 [6.52, 9.74] \\
+ EvoOpt & 7.6 [5.07, 9.07] & 9 [6.6, 9.77] & 8.2 [5.69, 9.4] & 8.27 [5.77, 9.44] & 5.2 [2.9, 7.42] & 7.47 [4.94, 8.99] & 7.8 [5.27, 9.18] & 9 [6.6, 9.77] \\
\bottomrule
\end{tabular}%
}
\end{table}

\begin{table}[H]
\centering
\caption{CoGen complementary-reconstruction ablation with confidence intervals. Each entry reports the mean followed by the 95\% confidence interval in brackets.}
\label{tab:supp:ablation_edit_ci}
\scriptsize
\setlength{\tabcolsep}{2.5pt}
\resizebox{\linewidth}{!}{%
\begin{tabular}{lcccccccc}
\toprule
Method & C.1 & C.2 & C.3 & C.4 & C.5 & C.6 & C.7 & C.8 \\
\midrule
\multicolumn{9}{l}{\textit{O.M. $\uparrow$}} \\
w/o c.r. & 7 [4.48, 8.7] & 9.2 [6.85, 9.84] & 9.07 [6.68, 9.79] & 8.53 [6.06, 9.57] & 8.13 [5.62, 9.37] & 8.93 [6.52, 9.74] & 8.87 [6.44, 9.71] & 7.6 [5.07, 9.07] \\
Full model & 8.93 [6.52, 9.74] & 9.27 [6.93, 9.86] & 9.33 [7.02, 9.88] & 9.13 [6.77, 9.82] & 8.8 [6.37, 9.68] & 9 [6.6, 9.77] & 9.53 [7.28, 9.94] & 8.8 [6.37, 9.68] \\
\midrule
\multicolumn{9}{l}{\textit{R.M. $\uparrow$}} \\
w/o c.r. & 5.8 [3.4, 7.87] & 2.8 [1.17, 5.32] & 7.27 [4.74, 8.87] & 6.53 [4.05, 8.39] & 2.6 [1.05, 5.13] & 6.73 [4.23, 8.53] & 7.2 [4.68, 8.83] & 4 [1.98, 6.43] \\
Full model & 7.6 [5.07, 9.07] & 9 [6.6, 9.77] & 8.2 [5.69, 9.4] & 8.27 [5.77, 9.44] & 5.2 [2.9, 7.42] & 7.47 [4.94, 8.99] & 7.8 [5.27, 9.18] & 9 [6.6, 9.77] \\
\bottomrule
\end{tabular}%
}
\end{table}

\subsubsection{User study}
\label{sec:supp:exp:user_study}

This subsection describes the human evaluation behind Tab.~\ref{tab:user_study}. We follow the pairwise comparison format used by SceneSmith~\citep{pfaff2026scenesmith}, adapted to our contact-geometry tasks.

\begin{figure}[H]
    \centering
    \includegraphics[width=0.7\linewidth]{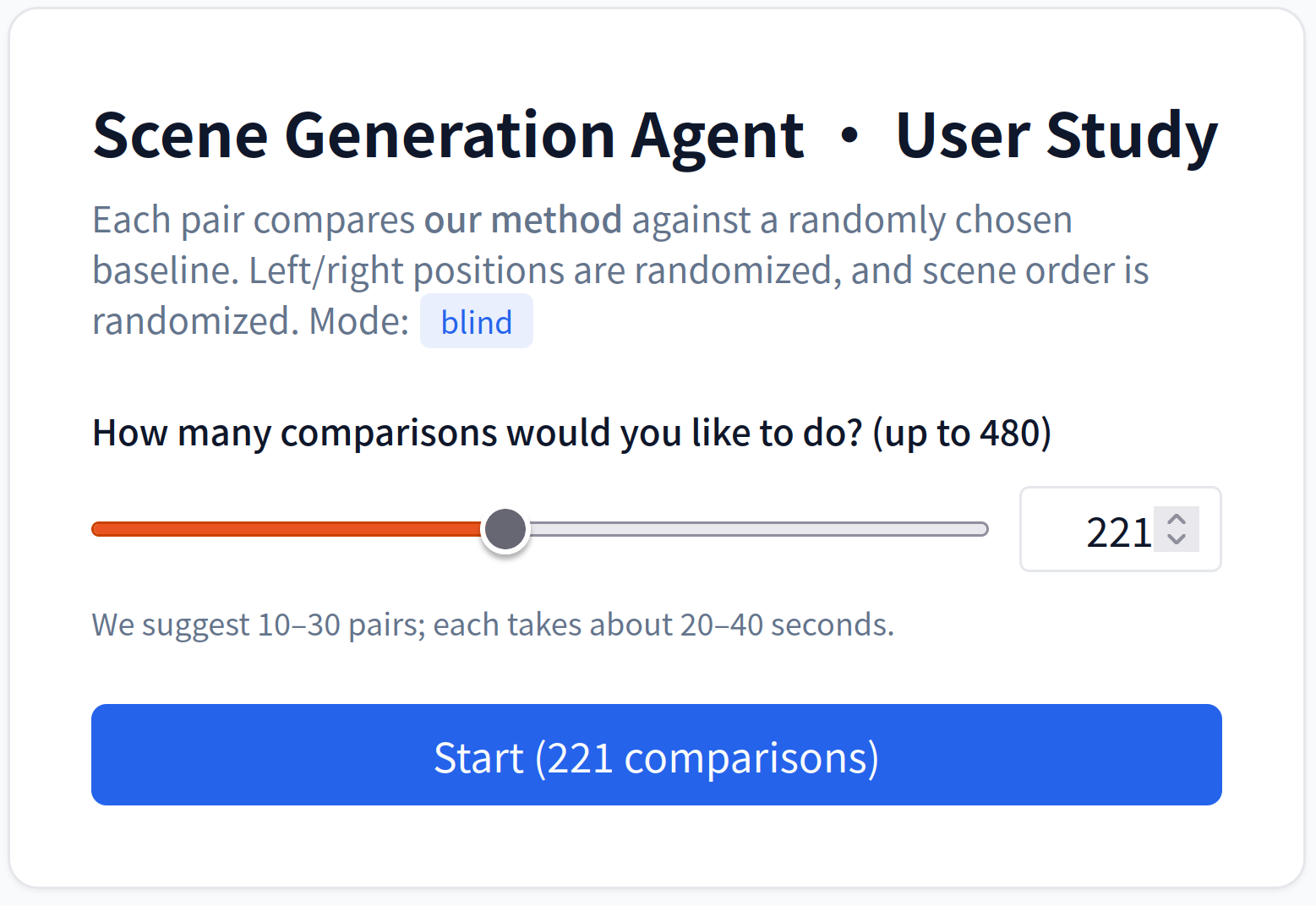}
    \caption{Tutorial landing page that introduces the rating task before any judgments are collected.}
    \label{fig:supp:user_study:landing}
\end{figure}

\paragraph{Participants.}
We recruited 107 raters from a university population and who had normal vision and basic familiarity with everyday object interactions. They were blind to the methods under comparison and to the study hypothesis. Before rating, each participant read a tutorial page explaining the task, the two metrics, and the \emph{Equal} option (Fig.~\ref{fig:supp:user_study:landing}). Across all sessions, we collected more than $1{,}000$ pairwise judgments distributed across the four baseline pairings and both prompt settings.

\begin{figure}[H]
    \centering
    \includegraphics[width=0.85\linewidth]{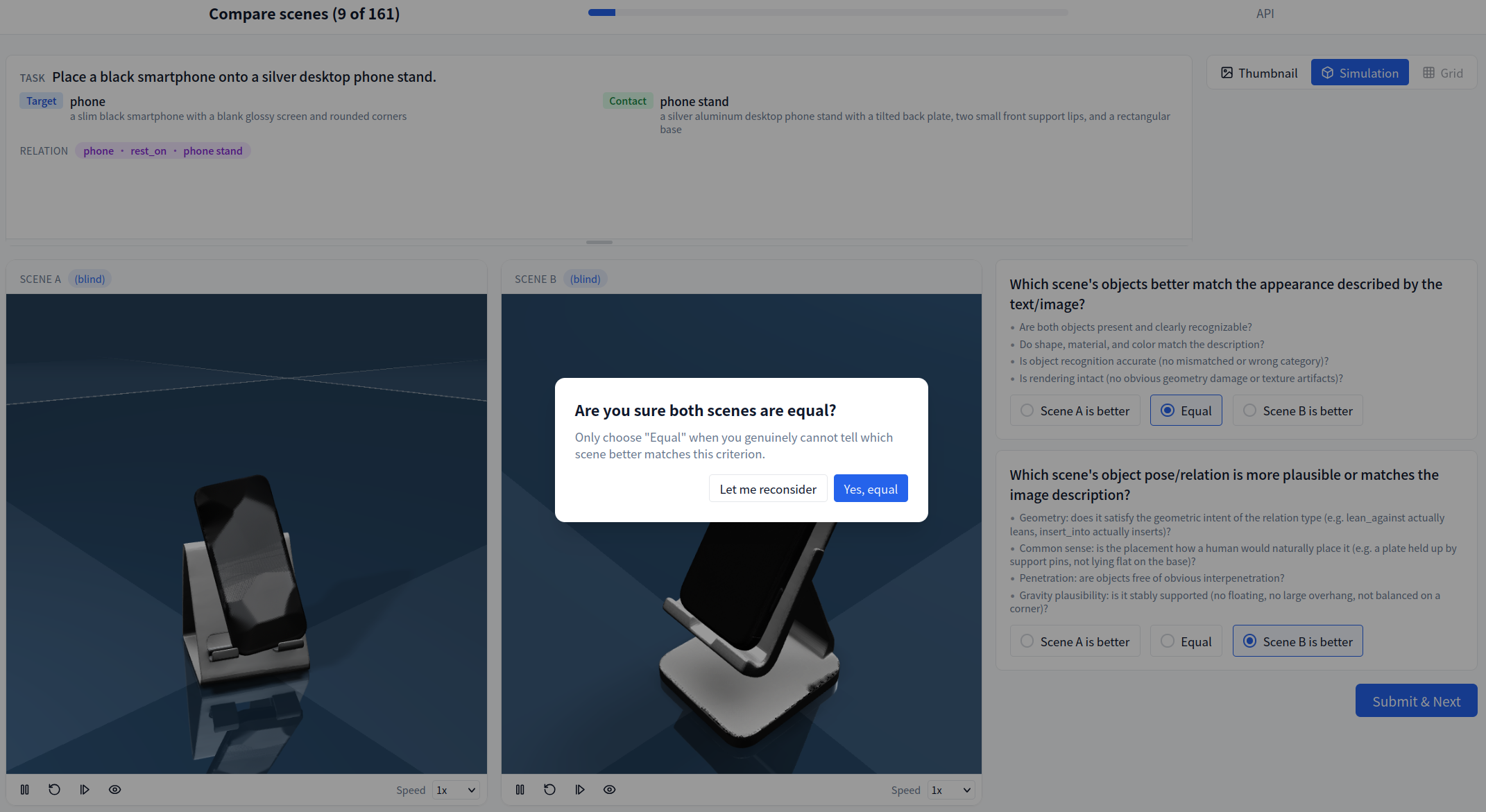}
    \caption{Confirmation dialog shown when a rater selects \emph{Equal}.}
    \label{fig:supp:user_study:equal}
\end{figure}

\begin{figure}[H]
    \centering
    \begin{minipage}[t]{0.49\linewidth}
        \centering
        \includegraphics[width=\linewidth]{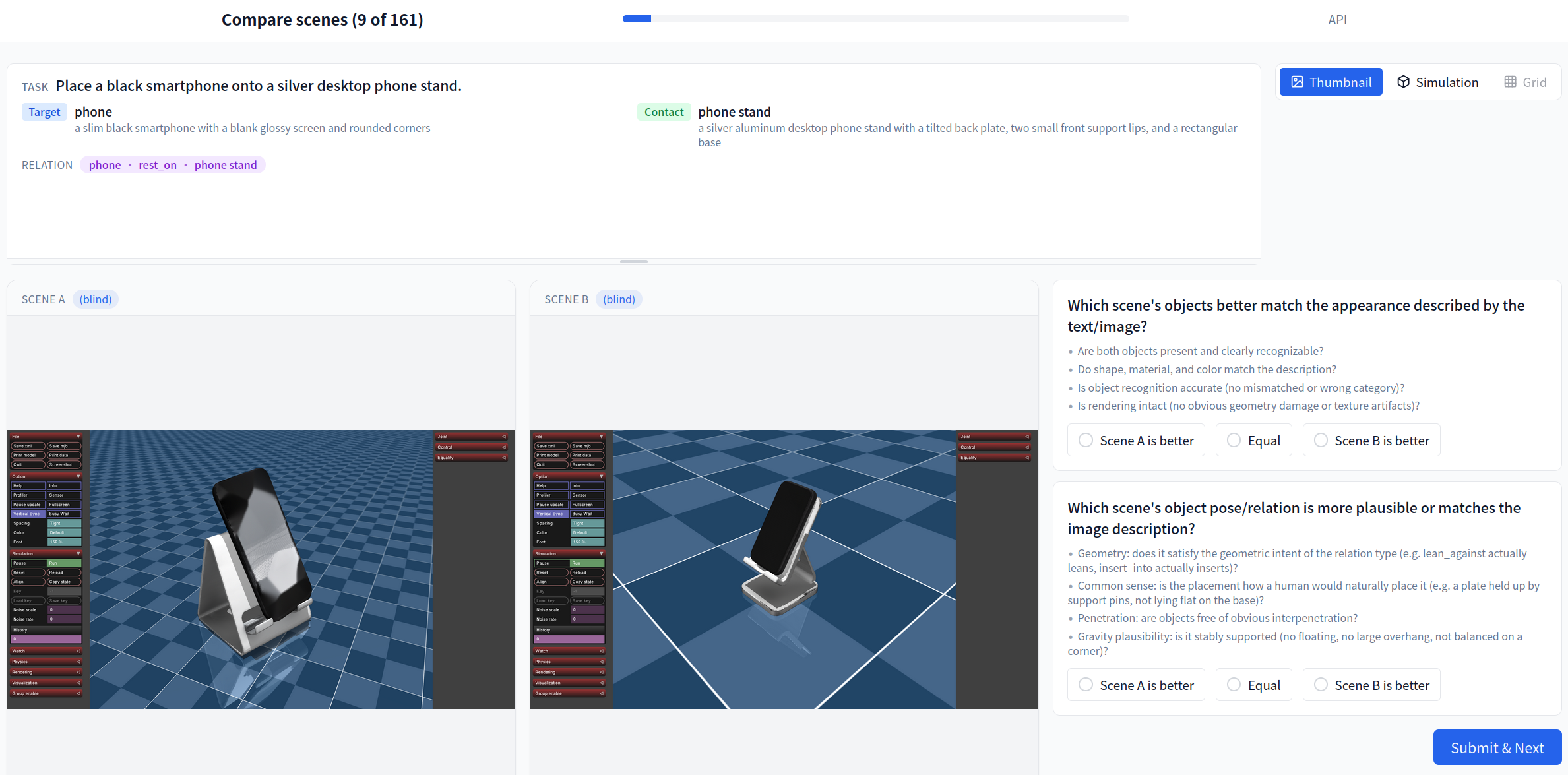}
        \\\small (a) Language prompt, settled composite in MuJoCo.
    \end{minipage}
    \hfill
    \begin{minipage}[t]{0.49\linewidth}
        \centering
        \includegraphics[width=\linewidth]{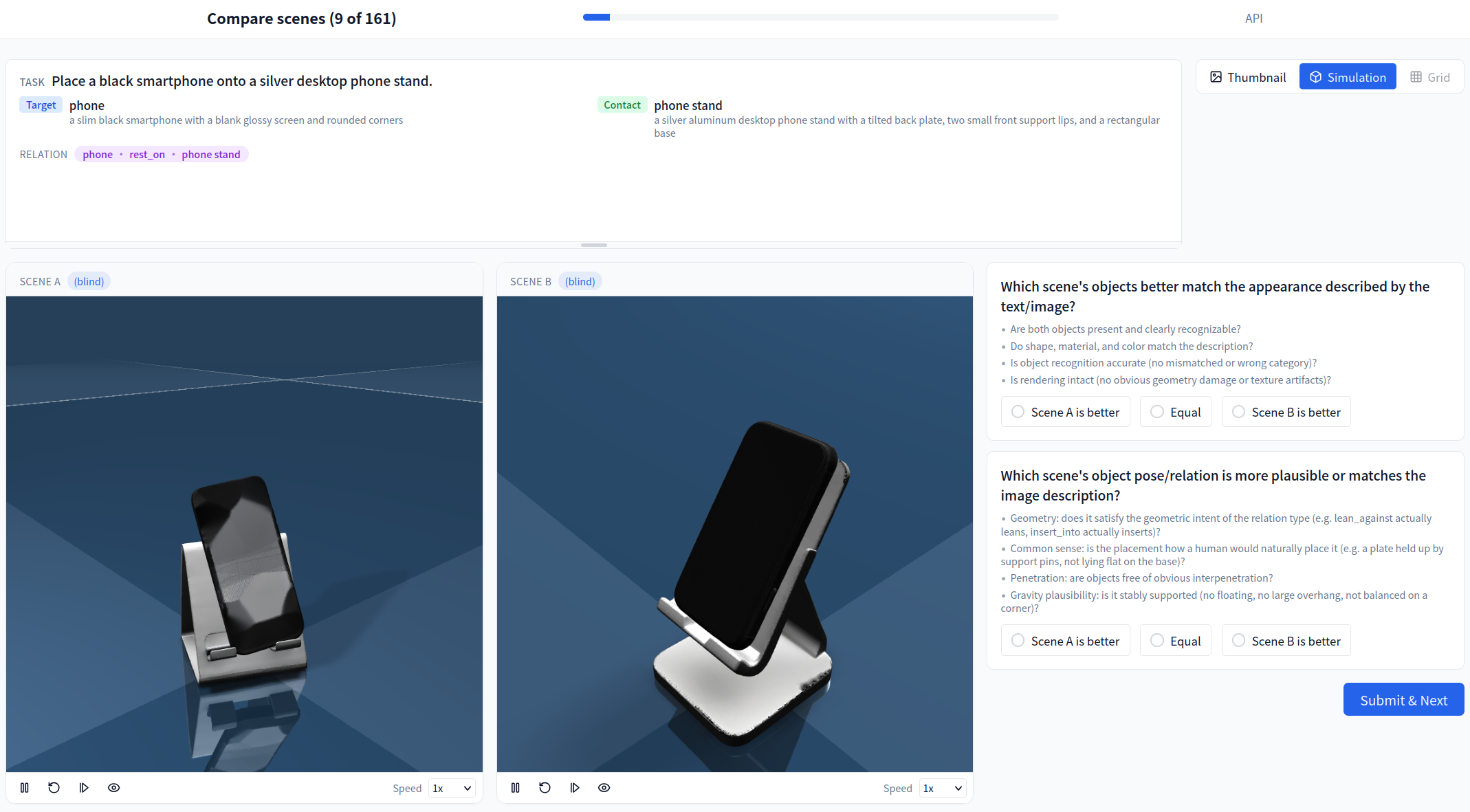}
        \\\small (b) Language prompt, MuJoCo render.
    \end{minipage}
    \\[0.6em]
    \begin{minipage}[t]{0.49\linewidth}
        \centering
        \includegraphics[width=\linewidth]{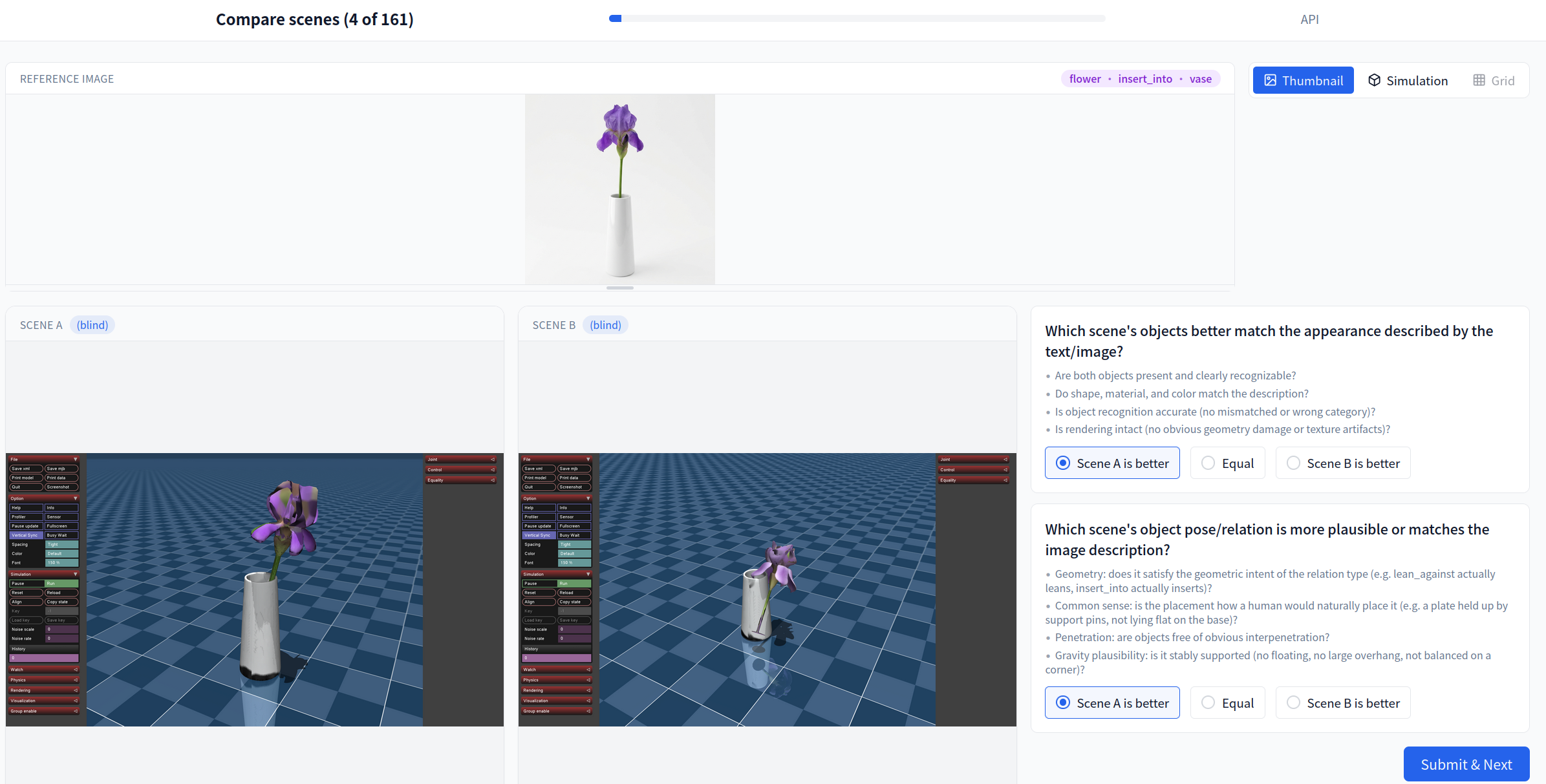}
        \\\small (c) Image prompt, settled composite in MuJoCo.
    \end{minipage}
    \hfill
    \begin{minipage}[t]{0.49\linewidth}
        \centering
        \includegraphics[width=\linewidth]{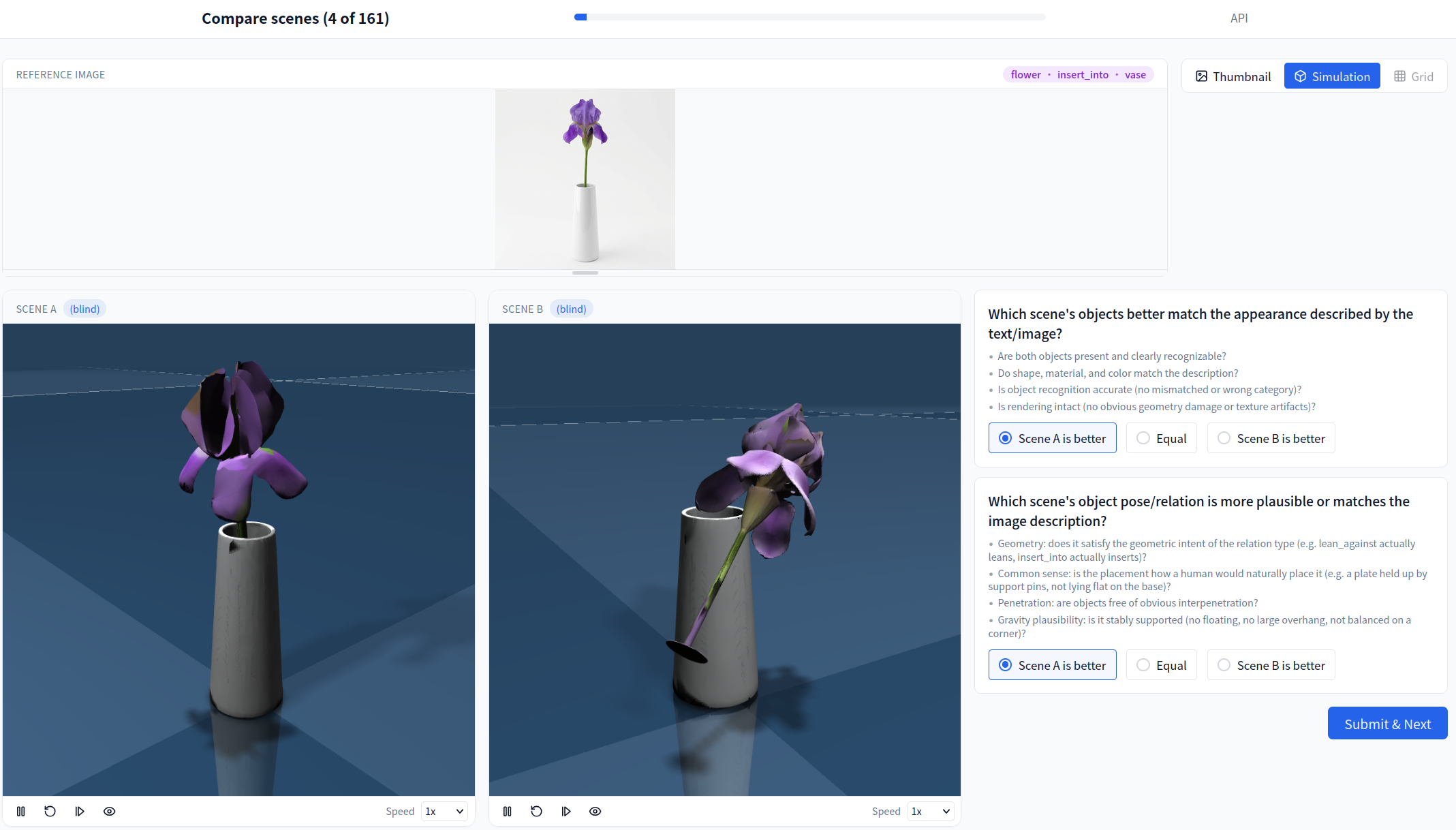}
        \\\small (d) Image prompt, MuJoCo render.
    \end{minipage}
    \caption{Four rating panels obtained by crossing the prompt modality, language and image, with the view modality, settled composite in MuJoCo and MuJoCo render. The MuJoCo render supports interactive inspection, including running, pausing, and resetting the simulation.}
    \label{fig:supp:user_study:panels}
\end{figure}

\paragraph{Rating interface and protocol.}
Each trial shows the input prompt and two anonymized outputs, one from our method and one from a baseline. Left--right placement is randomized and method names are hidden. The rater chooses which output is better, or selects \emph{Equal} when the two are indistinguishable; an additional confirmation dialog is shown for ties (Fig.~\ref{fig:supp:user_study:equal}). O.M. is judged from the generated assets and their rendered appearance, while R.M. is judged from the gravity-settled MuJoCo state and its render. The four interface variants, crossing prompt modality with two view types, are shown in Fig.~\ref{fig:supp:user_study:panels}.

\paragraph{Aggregation and statistical analysis.}
For O.M., tied votes contribute half a win to each method. For R.M., ties are excluded before computing the win rate, so the reported value reflects decisive relation-quality preferences. For every comparison we report the win rate of our method, the $95\%$ Wilson confidence interval, and Cohen's $h$ effect size against the $50\%$ null.

\paragraph{Quality control.}
Each session begins with a short calibration set containing clearly successful and clearly unsuccessful scene pairs; raters below the calibration threshold are excluded from the final analysis. Trial order, side assignment, and method-baseline pairing are independently randomized, and method names remain hidden throughout the study.

\subsection{Baselines and Comparison Protocol}
\label{sec:supp:exp:baselines}

This subsection specifies how each baseline is adapted to the $24\times5=120$ benchmark scenes. We set the runtime human-feedback budget to zero for every system so that outcomes reflect the automated pipelines rather than operator skill. Whenever a baseline uses an image-to-3D backend, we replace that backend with SAM3D~\citep{sam3d} in our evaluation, so that comparisons focus on scene composition, relation reasoning, and physical placement rather than on differences in asset-generation quality.

\begin{table}[H]
\centering
\caption{Summary of baseline adaptations for the pairwise contact-geometry benchmark. All exported scenes are settled in MuJoCo and evaluated with the same renderer and scoring protocols described in Sec.~\ref{sec:supp:exp:metrics}.}
\label{tab:supp:baseline_adaptation}
\scriptsize
\setlength{\tabcolsep}{2pt}
\renewcommand{\arraystretch}{1.12}
\begin{tabular}{@{}p{0.15\linewidth}p{0.20\linewidth}p{0.20\linewidth}p{0.25\linewidth}p{0.14\linewidth}@{}}
\toprule
Baseline & Original assumption & Benchmark mismatch & Adaptation & Fairness control / residual caveat \\
\midrule
RoLA~\citep{wei2025rola}
& Reconstructs simulation-ready assets from a single RGB image and instance masks.
& Requires image inputs and masks, whereas the benchmark includes canonical language and image specifications.
& Provide masks through the SAM2 annotation interface distributed with RoLA; pass the resulting crops through RoLA's reconstruction and pose-composition stages with SAM3D as the shared 3D backend.
& Uses RoLA's expected input format; mask preparation is applied consistently across RoLA runs. \\
\midrule
TabletopGen~\citep{tabletopgen2025}
& Generates tabletop assets independently and places them on a fixed support surface.
& Assumes a tabletop support surface, while some benchmark relations are not naturally table-only.
& Use table-conditioned benchmark images and replace the image-to-3D backend with SAM3D, leaving the remaining segmentation, asset-generation, and layout stages unchanged.
& Preserves the released pipeline except for the shared 3D backend; the protocol satisfies TabletopGen's table assumption. \\
\midrule
RoLA / TabletopGen image protocol
& Both systems expect an image of a tabletop scene.
& Foreground-only references omit the support surface required by these baselines.
& Edit each foreground reference into a canonical white-table scene while preserving object viewpoint and adding no unrelated items; for wall-hanging cases, add a minimal wall support.
& The same composites are used for RoLA, TabletopGen, and Ours-Img. \\
\midrule
GenieSim~\citep{geniesim2025}
& Interactive generation with optional human-in-the-loop editing and asset acquisition.
& Human intervention cannot be used in a fully automated 120-scene comparison.
& Automate the released web interface with Playwright, fix the model choice, disable human editing, and append an instruction requiring asset acquisition rather than early termination.
& Gives every run an automatic export path; does not evaluate manual interactive correction. \\
\midrule
SceneSmith~\citep{pfaff2026scenesmith}
& Hierarchical room-level scene synthesis with room context and routing strategies.
& The benchmark evaluates pairwise rigid-object contact rather than room-scale layout or articulated/thin-covering routing.
& Evaluate the manipuland-placement stage in a fixed empty room with a single work desk; supply object names as required manipulands and express the task as a tabletop placement constraint.
& Matches the pairwise object scope; room-level context and routing strategies outside the benchmark scope are not evaluated. \\
\bottomrule
\end{tabular}
\end{table}

\paragraph{RoLA.}
RoLA~\citep{wei2025rola} reconstructs simulation-ready assets from a single RGB image and composes them into a physical scene. The released implementation supports SAM3D~\citep{sam3d} as an alternative to its original Hunyuan3D~\citep{hunyuan3d2} backend, but it expects instance masks as input. We therefore provide masks using the SAM2~\citep{ravi2024sam2} annotation interface distributed with their source code, applying the same mask preparation procedure across all RoLA inputs. The resulting object crops are passed to RoLA's mesh reconstruction and pose-composition stages.

\paragraph{TabletopGen.}
TabletopGen~\citep{tabletopgen2025} generates each tabletop asset independently and places the assets on a fixed support surface. We replace its original Hunyuan3D~\citep{hunyuan3d2} backend with SAM3D~\citep{sam3d} and otherwise leave its segmentation, asset generation, and layout stages unchanged.

\paragraph{Image prompt protocol for table-required baselines.}
RoLA and TabletopGen expect an image of a tabletop scene. To match their input assumptions, we synthesize a table-conditioned image prompt for each benchmark instance. The image-editing backend in Tab.~\ref{tab:supp:backends} receives the foreground reference and a canonical white-table image, and is instructed to place the same objects on the table while preserving their viewpoint and adding no unrelated items. For wall-hanging relations, the editor is additionally instructed to add a minimal wall support on the table and hang the object from it. These composites serve as the image inputs to RoLA, TabletopGen and ``Ours-Img''.

\paragraph{GenieSim.}
GenieSim~\citep{geniesim2025} is evaluated without its human-in-the-loop editing stage. Since the released system is primarily exposed through an interactive web interface, we automate the same interaction with Playwright~\citep{playwright}: the task description is submitted to the generator panel, the model choice is fixed across all runs, and the exported MuJoCo scene is collected for evaluation. We append a short instruction that requires the system to acquire or synthesize the necessary assets rather than terminating when a library asset is unavailable. Each generated scene is then settled under gravity and rendered with the same evaluation renderer used for all methods.

\paragraph{SceneSmith.}
SceneSmith~\citep{pfaff2026scenesmith} targets hierarchical room-level synthesis. To align its scope with our pairwise benchmark, we evaluate only the manipuland-placement stage and provide a fixed empty room containing a single rectangular work desk. Objects names are supplied as required manipulands, and the task description is expressed as a tabletop placement constraint on that desk. SceneSmith then runs its standard asset routing, image-editing, and pose-composition stages before export to MuJoCo. We disable routing strategies for articulated objects and thin coverings, since our benchmark focuses on separable rigid-body pairs.

\subsection{Efficiency, API Usage, and Statistical Uncertainty}
\label{sec:supp:exp:efficiency}
\begin{figure}[H]
    \centering
    \includegraphics[width=\linewidth]{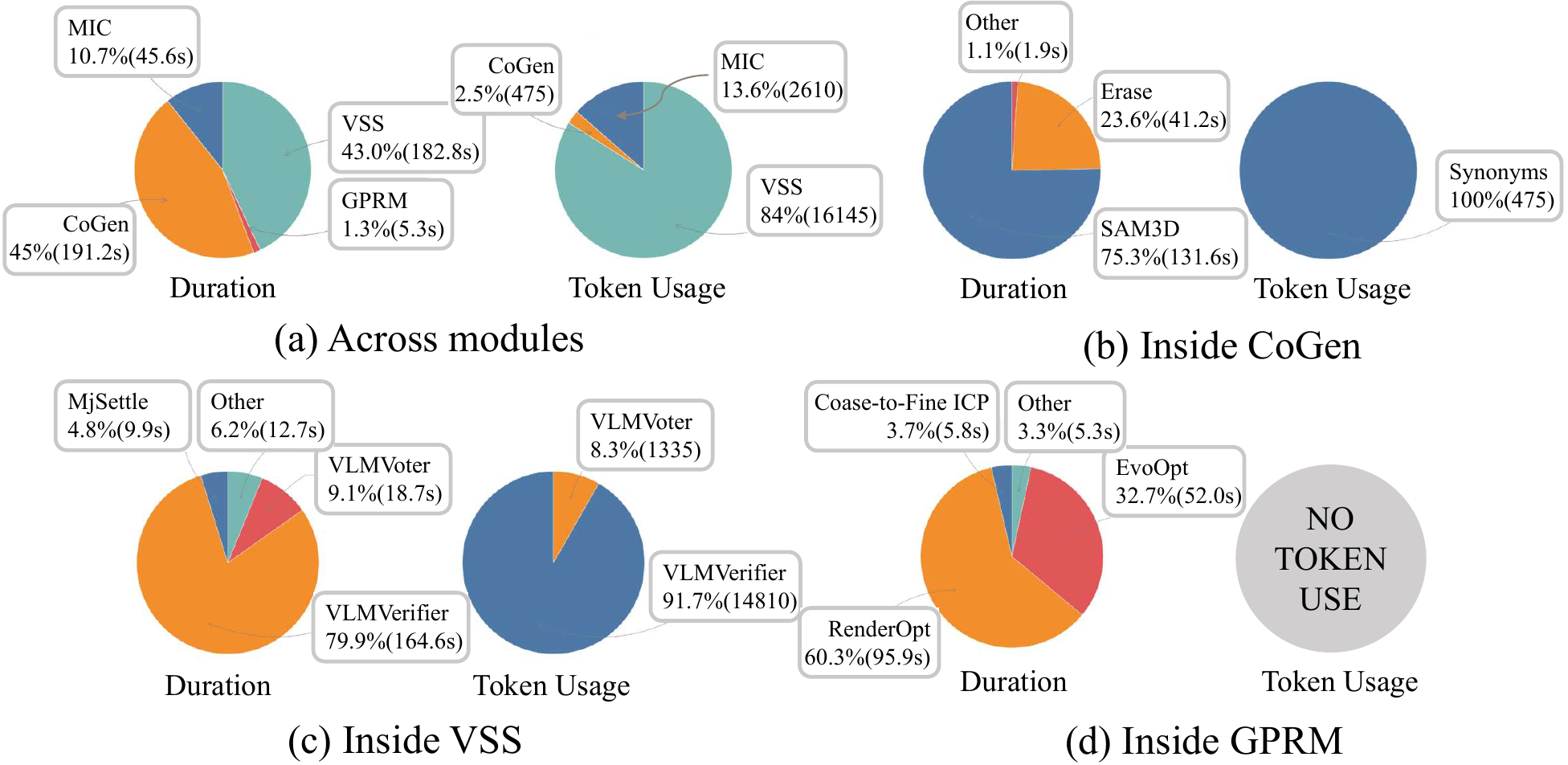}
    \caption{Runtime and token usage across GIF modules and their principal tools, averaged over the 120 image-prompt benchmark scenes. Erase denotes object removal by the image-editing model, and MjSettle denotes gravity settling and rendering in MuJoCo.}
    \label{fig:supp:cost}
\end{figure}

Tab.~\ref{tab:supp:runtime} reports per-module cost averaged over 120 image-prompt scenes, and Fig.~\ref{fig:supp:cost} gives the module shares with the toollevel breakdown inside the three modules. GIF requires an average of 425 s per scene (range: 243--935 s), uses 19.2K API tokens (8.4K--31.5K), and costs approximately \$0.25 per scene. The VLM verifier accepts 92\% of attempts. GPRM consumes 5.3 s and no tokens at all. Its wall clock appears short because we run the pose candidates with multiprocessing, whereas the timing recorded inside a single candidate is far larger. We have not logged token counts for image generation, but the call count averages 1.1 per scene in MIC and 2.3 in the Erase tool inside CoGen.

\begin{table}[H]
\centering
\caption{Average wall-clock time (A.W.C., seconds) and API tokens (A.T., thousands) per module over 120 image-prompt scenes. Parentheses give the observed minimum and maximum. MIC denotes Multimodal Input Canonicalization and VSS denotes VLM-based Scene Selection.}
\label{tab:supp:runtime}
\small
\begin{tabular}{lccccc}
\toprule
 & MIC & CoGen & GPRM & VSS & \textbf{All} \\
\midrule
A.W.C. & 45.6 (27--154) & 191.2 (161--328) & 5.3 (2--11) & 182.8 (55--673) & \textbf{425.0 (243--935)} \\
A.T. (K) & 2.6 (2.5--2.8) & 0.5 (0.0--2.6) & 0.0 & 16.1 (5.8--26.6) & \textbf{19.2 (8.4--31.5)} \\
\bottomrule
\end{tabular}
\end{table}

\subsection{Preliminary Extension to Multi-Object Scenes}
\label{sec:supp:exp:multiobj}

\begin{figure}[H]
    \centering
    \includegraphics[width=\linewidth]{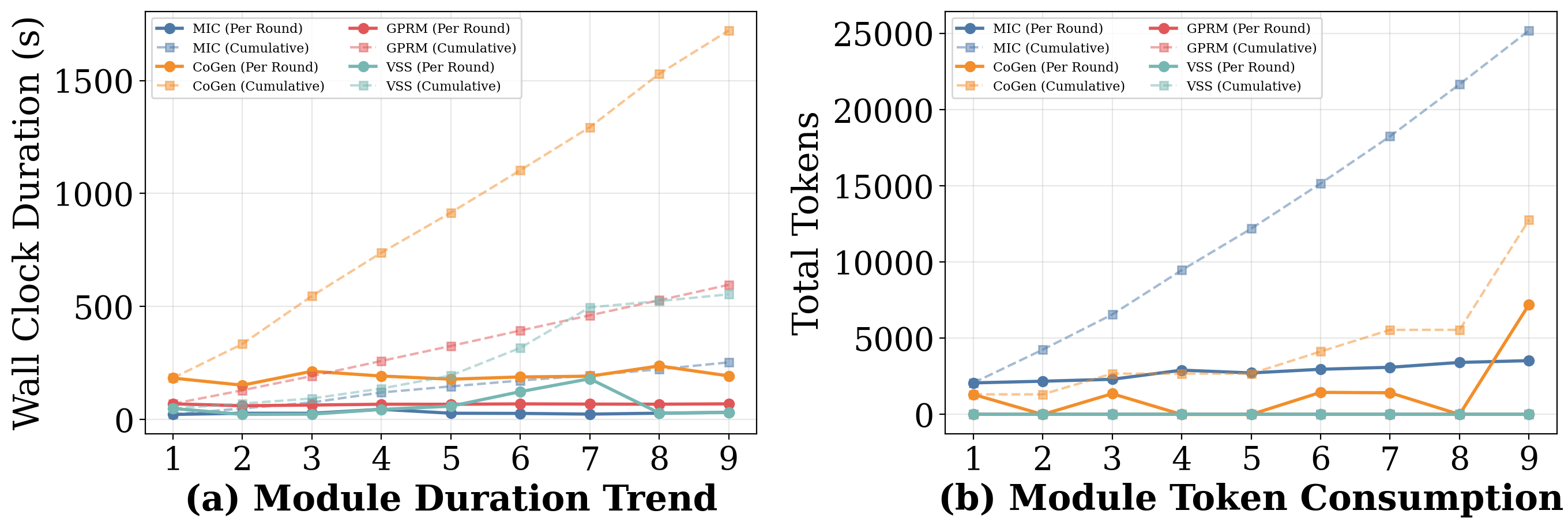}
    \caption{Runtime and token-cost curves for sequential multi-object expansion. Each point adds one object while retaining all previously verified geometry.}
    \label{fig:supp:multiobj_cost}
\end{figure}

\begin{figure}[H]
    \centering
    \includegraphics[width=\linewidth]{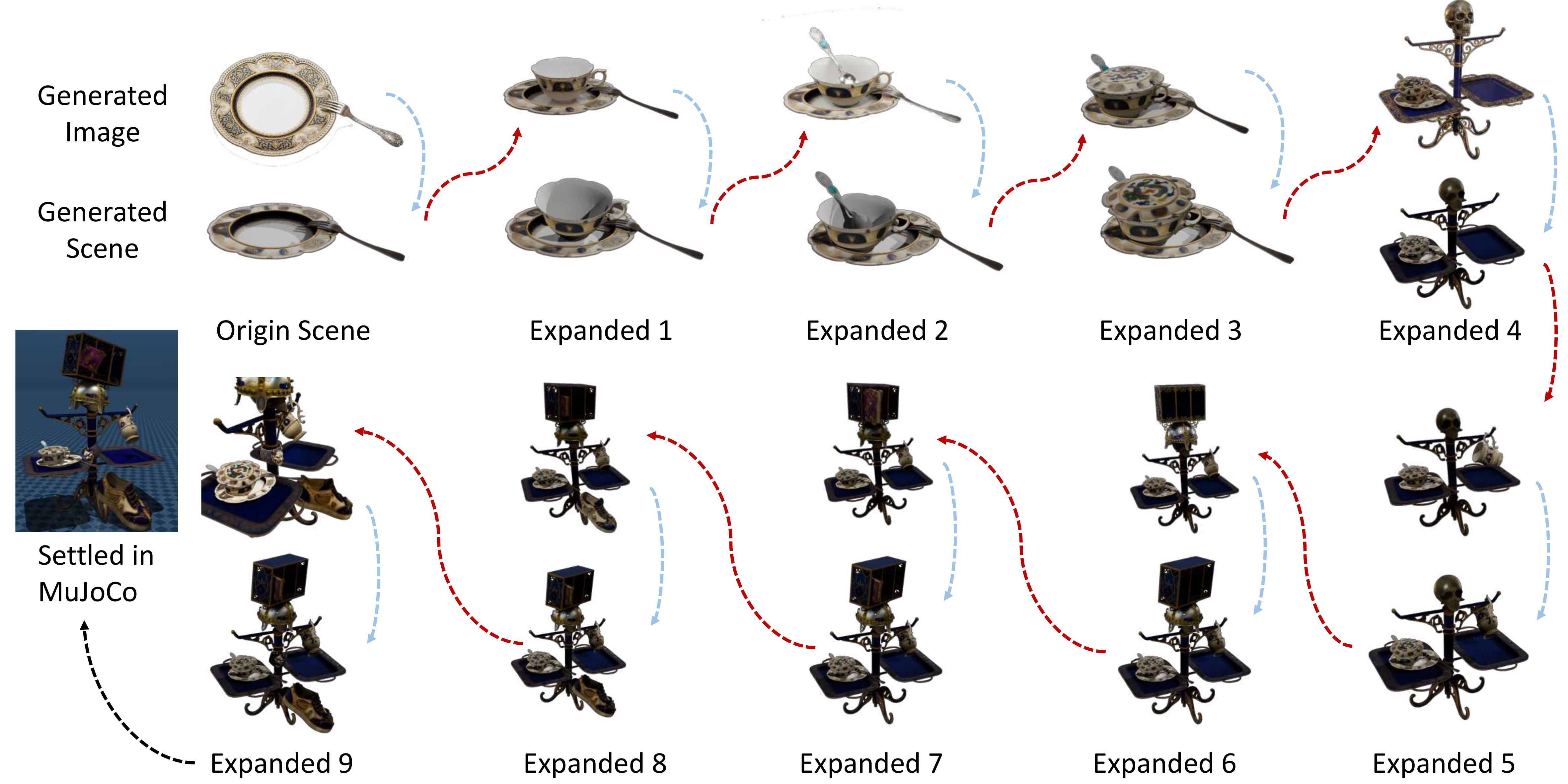}
    \caption{Preliminary extension to multi-object scenes. Starting from a verified fork-on-plate pair, the multimodal agent inserts one new object at a time, and CoGen together with GPRM is rerun on the newcomer while the merged geometry of all previously placed objects is held fixed. The nine insertions are: an empty cup placed on the plate, a spoon set into the cup, a notched lid capping the cup, an ornate skull-shaped rack standing beside the plate, a mug hooked onto a side branch, a helmet-shaped bookshelf resting on the skull, a magic book slotted into the shelf, a skateboard shoe leaning against the central pole, and a ring pegged around it. Together, these additions span all eight taxonomy classes from \textbf{C.1 Containment} to \textbf{C.8 Pegging}.}
    \label{fig:supp:multiobj}
\end{figure}

Fig.~\ref{fig:supp:multiobj} shows the full nine-step expansion summarized in Sec.~\ref{sec:exp:extension}. We should state plainly that the multi-object extension is preliminary. The example shown in our paper was produced with a human in the loop, who adjusted parameters, switched submodules or retried tools, and acted as the verifier. Difficult steps such as placing a book into the helmet-shaped shelf at the sixth extension needed several attempts and manual intervention. What does stay controlled is the cost of a single attempt, since placed objects are frozen as merged static geometry and only the newcomer passes through GIF, so per-attempt wall clock and token cost grow close to linearly with the object count, as shown in Fig.~\ref{fig:supp:multiobj_cost}. We agree with the that the system-level cost may still grow sharply once the number of retries is counted, and we regard this as a genuine open risk. Our future work is to design a more reasonable framework for multi object scenes, and to conduct more detailed evaluations and handle the risk of significant cost increases that may arise as a result.

\section{Task Suite Descriptions and Scene Galleries}
\label{sec:supp:tasks}

This section lists the $8\times3=24$ benchmark tasks and the five semantic variants used for each task, giving the natural-language prompt for every one of the $120$ reference scenes. The galleries are organized by contact-geometry class: each $3\times5$ grid uses one row per task and five columns for its variants.

\subsection{C.1 Containment}
\label{sec:supp:tasks:c1}

\noindent\textbf{Task 1: Place a flower into a vase.}
\begin{itemize}\setlength{\itemsep}{1pt}
    \item The leafless flower stem is inserted vertically through the mouth of the vase, with the blossom fully visible above the rim. The flower is a single real red rose with one long green stem, a full red blossom, no leaves anywhere on the stem, and natural green sepals allowed beneath the blossom. The vase is a narrow transparent glass bud vase with a round base, a slim neck, and an open circular mouth.
    \item The leafless flower stem is inside the vase opening and the white tulip bloom stands above the vase. The flower is a single real white tulip with a smooth pale green stem, a closed cup-shaped white bloom, no leaves attached to the stem, and natural sepals allowed near the bloom. The vase is a small matte blue ceramic vase with a rounded body, narrow neck, and open top.
    \item The leafless flower stem is inserted into the center of the vase opening, with the yellow flower head clear of the rim. The flower is a single real yellow daisy with a dark center, a thin green stem, no leaves attached to the stem, and small natural sepals allowed under the flower head. The vase is a short brushed stainless steel vase with vertical ribbing and a flared open lip.
    \item The leafless flower stem goes down into the vase mouth and the purple iris blossom is upright above the vase. The flower is a single real purple iris with a slender green stem, layered purple petals, no leaves on the stem, and natural sepals allowed at the base of the flower. The vase is a tall glossy white porcelain vase with a tapered cylindrical body and a simple open mouth.
    \item The bare leafless flower stem is inserted into the square opening of the empty vase while the pink gerbera flower head remains outside and above it. The flower is a single real pink gerbera daisy with a round flower head, a long bare green stem, no leaves, no side shoots, no bracts along the stem, and only a small natural calyx directly under the flower head. The vase is a clear square glass vase with thick walls, a flat base, and an open square top.
\end{itemize}

\noindent\textbf{Task 2: Place a pen into a multi-compartment desktop organizer.}
\begin{itemize}\setlength{\itemsep}{1pt}
    \item The pen is standing inside one narrow rectangular compartment of the white modern desktop organizer. The pen is a single slim black ballpoint pen with a smooth cylindrical barrel and a small clip. The desktop organizer is a multi-compartment desktop organizer with an open top structure, white matte plastic finish, clean rectangular sections of different sizes, crisp edges, and a compact minimalist geometric form.
    \item The pen stands inside the tallest rear compartment of the dark gray complex organizer box. The pen is a glossy white gel pen with a transparent grip section, a silver clip, and a fine conical tip. The complex organizer box is a dark gray stepped plastic organizer box with three different-height rectangular compartments, rounded internal dividers, and a low front storage tray, all compartments empty.
    \item The pen rests inside one narrow vertical compartment of the wooden desktop organizer. The pen is a single red marker style pen with a rounded cap and simple cylindrical body. The desktop organizer is a multi-compartment desktop organizer made of light wood, with open top rectangular sections of different sizes, clean sharp edges, a compact block-like structure, and a modern desk accessory appearance.
    \item The pen is inserted diagonally into the tall rectangular compartment of the acrylic complex organizer box. The pen is a navy blue rollerball pen with a straight cylindrical barrel, a rounded cap end, and a subtle metal clip. The complex organizer box is a clear smoky acrylic organizer box with visible thick internal dividers, one tall rectangular compartment, two small square wells, and a low front tray, all compartments empty.
    \item The pen is inserted into the deep oval opening of the ceramic complex organizer box. The pen is a yellow fine liner pen with a slim plastic barrel, a small cap, and a simple pointed tip. The complex organizer box is a matte cream ceramic organizer box with curved wave-like internal partitions, one deep oval opening, two smaller rounded pockets, and a shallow empty tray area.
\end{itemize}

\noindent\textbf{Task 3: Place a flowering plant pot into a hanging flower pot stand.}
\begin{itemize}\setlength{\itemsep}{1pt}
    \item The flowering plant pot is sitting inside the circular basket ring of the black hanging flower pot stand. The flowering plant pot is a beige ceramic flower pot containing a compact green leafy plant with several small yellow flowers. The hanging flower pot stand is a black metal hanging flower pot stand with a circular basket ring, vertical wire side bars, a flat round support base, and two curved hanging hooks, empty except for the pot.
    \item The flowering plant pot sits in the open cage of the black wire hanging flower pot stand. The flowering plant pot is a glossy cream ceramic flower pot with a leafy green plant and small white flowers rising above the rim. The hanging flower pot stand is a simple black wire hanging flower pot stand with a circular top ring, open cage-like sides, a round base support, and two slim upward hanging hooks.
    \item The flowering plant pot is seated inside the brass circular basket of the hanging flower pot stand. The flowering plant pot is a small matte charcoal flower pot containing broad green leaves and several red flowers, the pot body fully inside the frame. The hanging flower pot stand is a decorative brass-colored hanging flower pot stand with a circular basket rim, curved vertical ribs, a raised round bottom ring, and two symmetrical hanging hooks.
    \item The flowering plant pot rests on the shallow circular tray inside the black hanging flower pot stand. The flowering plant pot is a white cylindrical flower pot with compact green foliage and tiny orange flowers. The hanging flower pot stand is a compact black metal hanging flower pot stand with a shallow circular tray, short vertical guard bars, a top retaining ring, and rear hooks shaped for a balcony rail.
    \item The flowering plant pot is nested inside the wide circular basket of the green hanging flower pot stand. The flowering plant pot is a tan clay flower pot containing lush green leaves and a few small blue flowers. The hanging flower pot stand is a wide dark green metal hanging flower pot stand with an open circular basket, a low bottom ring, spaced vertical rods, and two rounded hook arms.
\end{itemize}

\begin{figure}[h]
\centering
\setlength{\tabcolsep}{1pt}
\renewcommand{\arraystretch}{0}
\begin{tabular}{ccccc}
\includegraphics[width=0.18\linewidth]{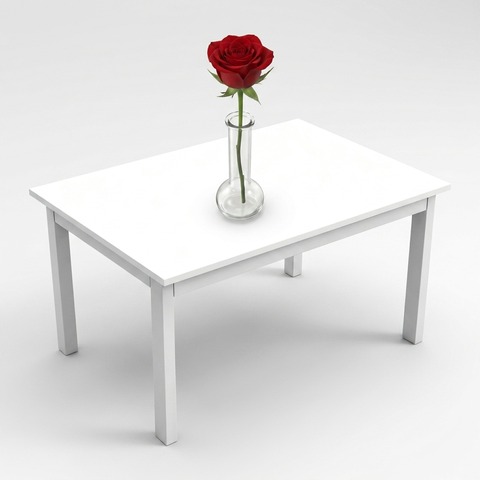} &
\includegraphics[width=0.18\linewidth]{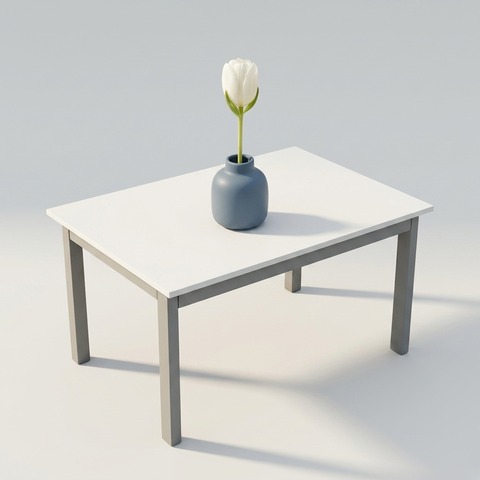} &
\includegraphics[width=0.18\linewidth]{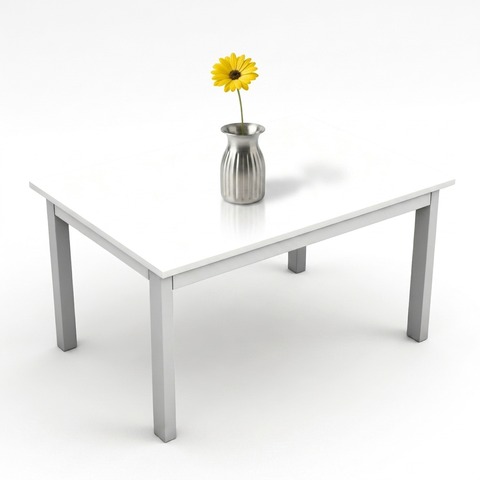} &
\includegraphics[width=0.18\linewidth]{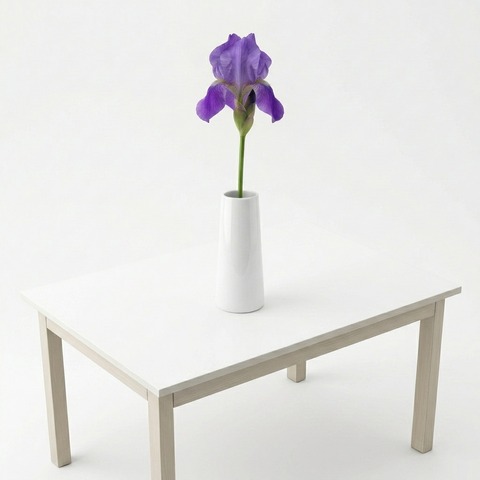} &
\includegraphics[width=0.18\linewidth]{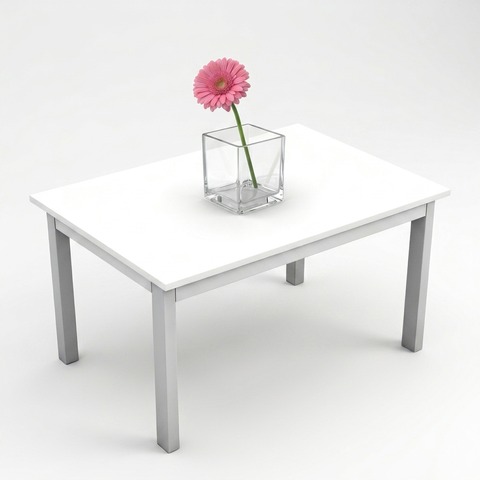} \\
\includegraphics[width=0.18\linewidth]{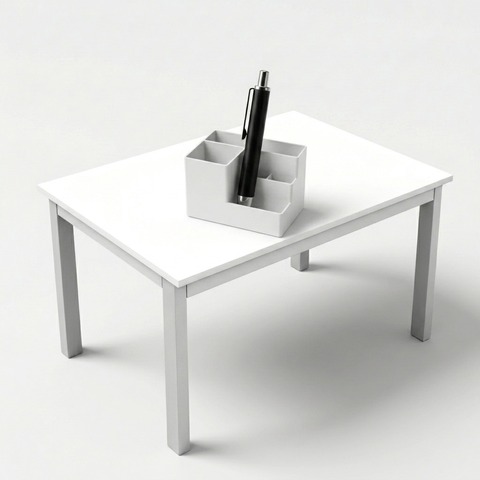} &
\includegraphics[width=0.18\linewidth]{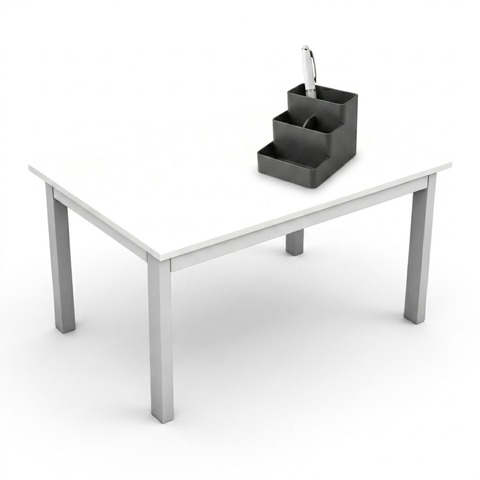} &
\includegraphics[width=0.18\linewidth]{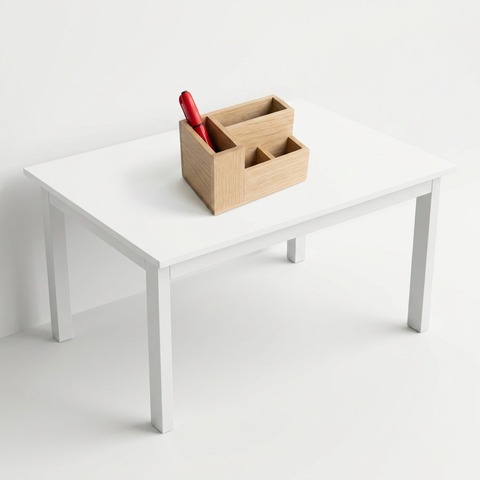} &
\includegraphics[width=0.18\linewidth]{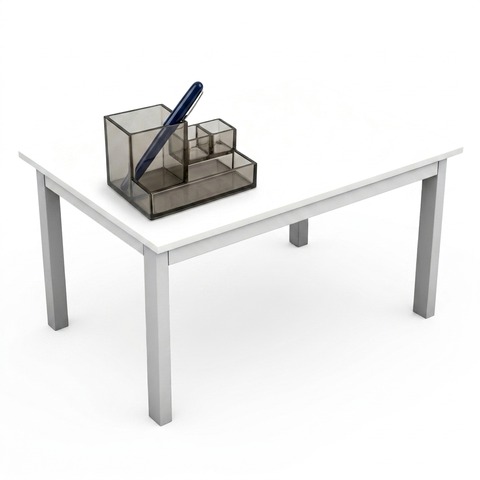} &
\includegraphics[width=0.18\linewidth]{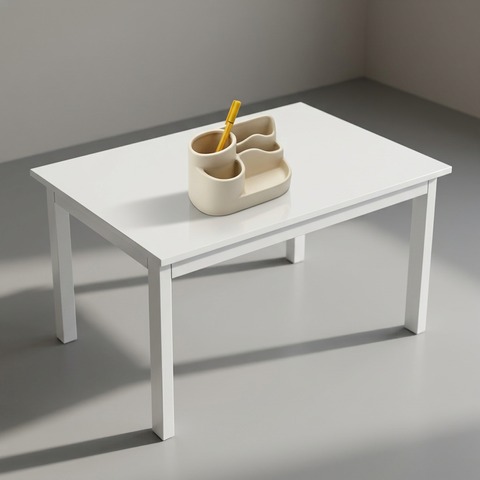} \\
\includegraphics[width=0.18\linewidth]{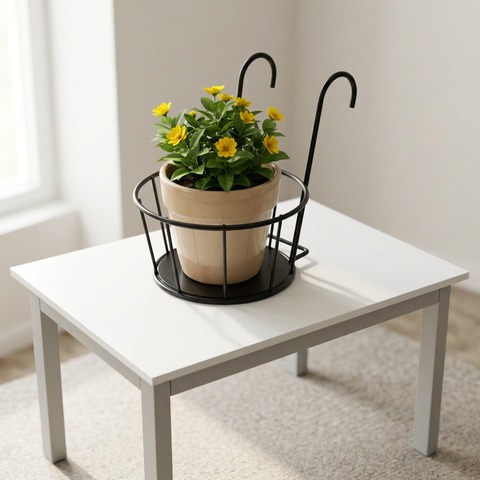} &
\includegraphics[width=0.18\linewidth]{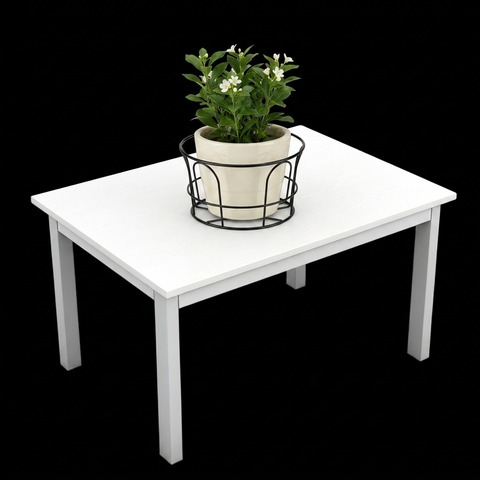} &
\includegraphics[width=0.18\linewidth]{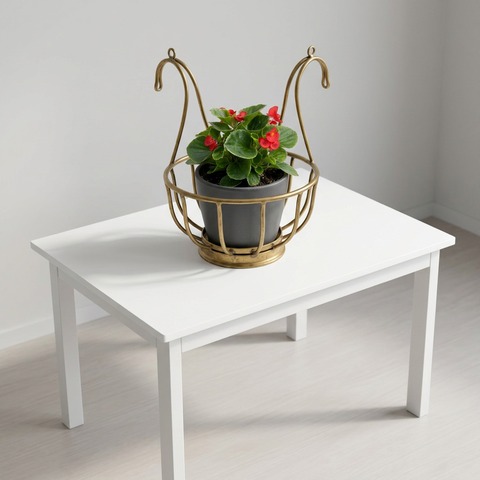} &
\includegraphics[width=0.18\linewidth]{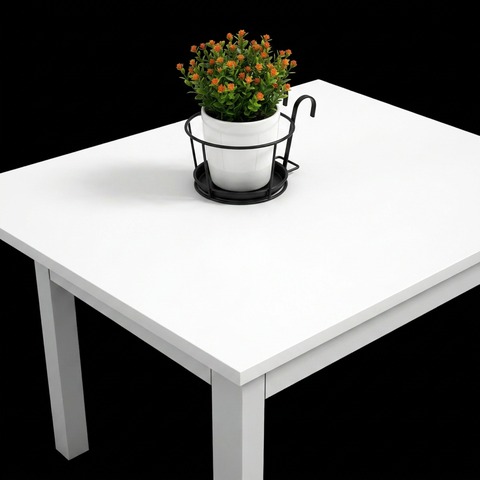} &
\includegraphics[width=0.18\linewidth]{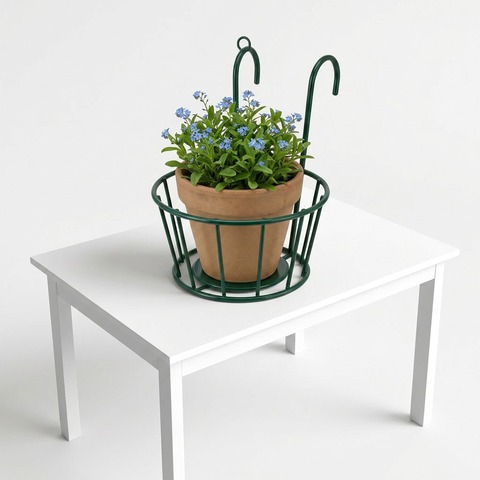} \\
\end{tabular}
\caption{Reference scenes for \textbf{C.1 Containment}. Top to bottom rows show flower-vase, pen-organizer, and pot-hanging-stand variants.}
\label{fig:supp:gallery:c1}
\end{figure}

\subsection{C.2 Capping}
\label{sec:supp:tasks:c2}

\noindent\textbf{Task 1: Fit a pot lid onto a pot.}
\begin{itemize}\setlength{\itemsep}{1pt}
    \item The lid is centered on the pot rim and covers the open circular top of the pot. The lid is a round transparent glass lid with a shiny metal rim and a black knob handle centered on top. The pot is a medium stainless steel pot with two side handles, straight walls, and an open circular rim.
    \item The lid sits flush on the upper rim of the pot and fully covers its opening. The lid is a heavy black cast iron lid with a raised circular knob and a slightly domed profile. The pot is a black cast iron Dutch oven with short side handles and a wide open mouth.
    \item The lid is aligned over the pot opening and rests on the rim as a cover. The lid is a glossy red enamel lid with a silver knob and a shallow dome. The pot is a matching glossy red enamel pot with a cream interior, loop side handles, and an open rim.
    \item The lid covers the saucepan opening with its edge seated on the pot rim. The lid is a flat brushed steel lid with a low black grip and a thin circular edge. The pot is a small stainless steel saucepan with one long handle and a clean open circular top.
    \item The lid is resting on the top rim of the stockpot and covers the opening. The lid is a round silver metal lid with a small steam vent hole and a dark round knob. The pot is a tall deep stockpot made of brushed stainless steel with two loop handles and an open top.
\end{itemize}

\noindent\textbf{Task 2: Fit a teacup lid onto its teacup.}
\begin{itemize}\setlength{\itemsep}{1pt}
    \item The teacup lid is seated on the circular rim of the matte black teacup. The teacup lid is a complete matching teacup lid with a small central knob, smooth rim, and clean unbranded surface, matte black accent details. The teacup is one complete matte black ceramic teacup with a visible handle, round open rim, stable base, and no tea bag or extra props.
    \item The teacup lid is seated on the circular rim of the natural wood teacup. The teacup lid is a complete matching teacup lid with a small central knob, smooth rim, and clean unbranded surface, natural wood accent details. The teacup is one complete natural wood ceramic teacup with a visible handle, round open rim, stable base, and no tea bag or extra props.
    \item The teacup lid is seated on the circular rim of the brushed metal teacup. The teacup lid is a complete matching teacup lid with a small central knob, smooth rim, and clean unbranded surface, brushed metal accent details. The teacup is one complete brushed metal ceramic teacup with a visible handle, round open rim, stable base, and no tea bag or extra props.
    \item The teacup lid is seated on the circular rim of the deep green teacup. The teacup lid is a complete matching teacup lid with a small central knob, smooth rim, and clean unbranded surface, deep green accent details. The teacup is one complete deep green ceramic teacup with a visible handle, round open rim, stable base, and no tea bag or extra props.
    \item The teacup lid is seated on the circular rim of the cream colored teacup. The teacup lid is a complete matching teacup lid with a small central knob, smooth rim, and clean unbranded surface, cream colored accent details. The teacup is one complete cream colored ceramic teacup with a visible handle, round open rim, stable base, and no tea bag or extra props.
\end{itemize}

\noindent\textbf{Task 3: Insert a wine cork into a red wine bottle.}
\begin{itemize}\setlength{\itemsep}{1pt}
    \item The wine cork is inserted into the mouth of the red wine bottle. The wine cork is one natural cylindrical wine cork with visible cork texture, flat top, and no label or text. The red wine bottle is one complete dark glass red wine bottle with a clean neck, round bottle mouth, simple unbranded body, and no label text.
    \item The natural wine cork is inserted into the mouth of the red wine bottle. The wine cork is one natural cylindrical wine cork with visible cork fibers and pores, a flat cut top and slightly irregular sides, no branding or text, clean and unbroken. The red wine bottle is one complete realistic dark glass red wine bottle with a round mouth opening and a neck sized for the cork; includes a practical applied paper label with generic winery-style graphics and subtle surface wear; visible glass reflections and fine texture on the bottle body.
    \item The wine cork is inserted into the mouth of the red wine bottle. The wine cork is one natural cylindrical wine cork with visible cork fibers and small irregular pores, flat cut top, intact rounded bottom, no text or markings, matte texture. The red wine bottle is one complete realistic dark glass red wine bottle with a clean round mouth (open neck), visible embossed glass imperfections, and a realistic paper wine label with decorative colors and subtle texture but no readable text.
    \item The wine cork is inserted into the mouth of the red wine bottle. The wine cork is one natural cylindrical wine cork with visible cork pores and slight unevenness, a flat top surface, and no markings, held as a single complete object. The red wine bottle is one complete red wine bottle made of dark glass with a clean realistic bottle neck and a round bottle mouth; the bottle body has a visible textured paper label with subtle wrinkles and printed-style artwork but no readable text, and the label is fully attached without any extra objects.
    \item The wine cork is inserted into the mouth of the red wine bottle. The wine cork is one natural cylindrical wine cork with visible cork texture, flat top, and no label or text. The red wine bottle is one complete dark glass red wine bottle with a clean neck, round bottle mouth, simple unbranded body, and no label text.
\end{itemize}

\begin{figure}[h]
\centering
\setlength{\tabcolsep}{1pt}
\renewcommand{\arraystretch}{0}
\begin{tabular}{ccccc}
\includegraphics[width=0.18\linewidth]{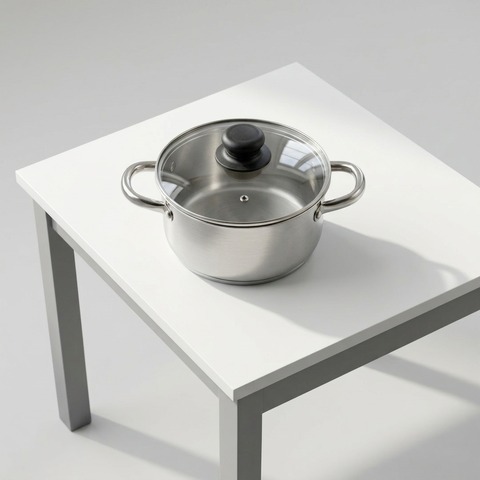} &
\includegraphics[width=0.18\linewidth]{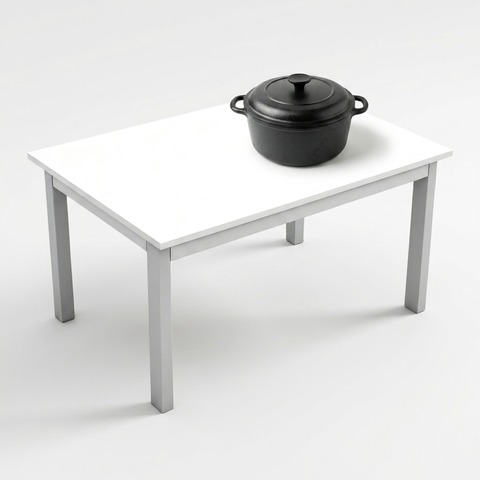} &
\includegraphics[width=0.18\linewidth]{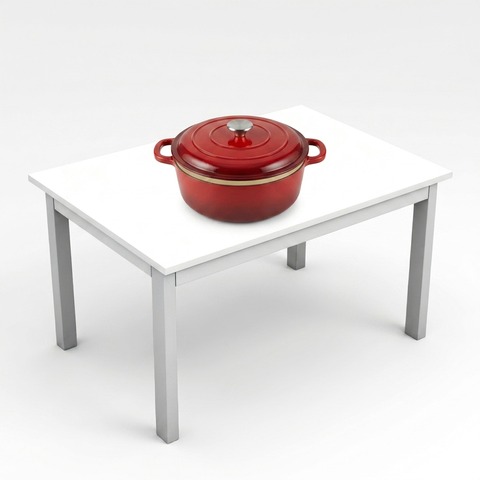} &
\includegraphics[width=0.18\linewidth]{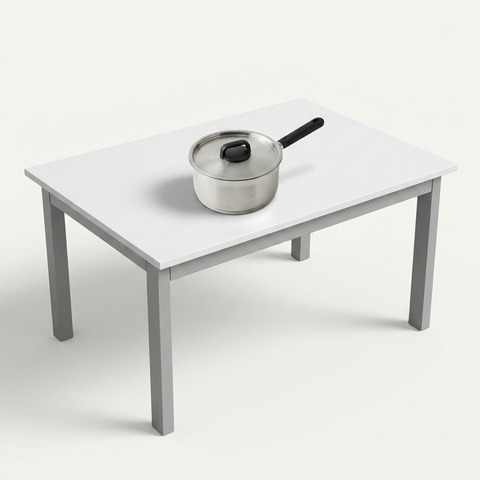} &
\includegraphics[width=0.18\linewidth]{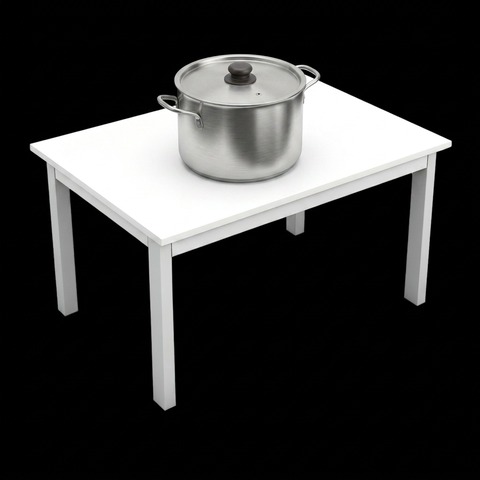} \\
\includegraphics[width=0.18\linewidth]{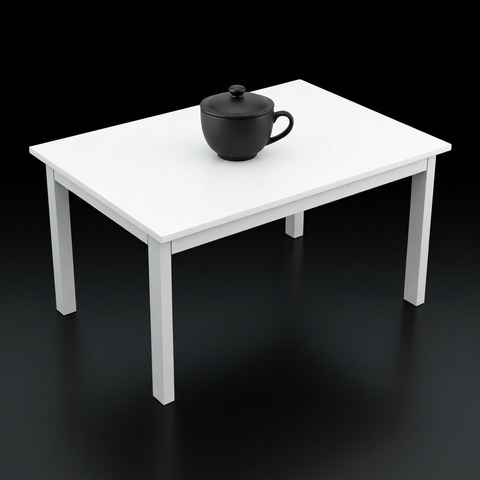} &
\includegraphics[width=0.18\linewidth]{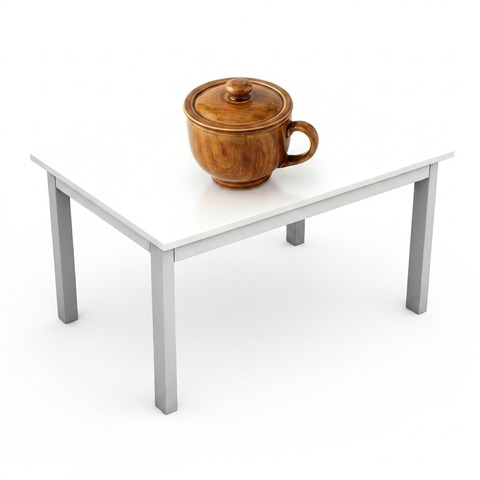} &
\includegraphics[width=0.18\linewidth]{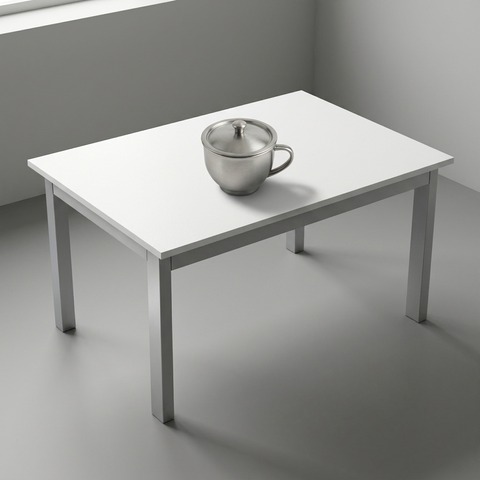} &
\includegraphics[width=0.18\linewidth]{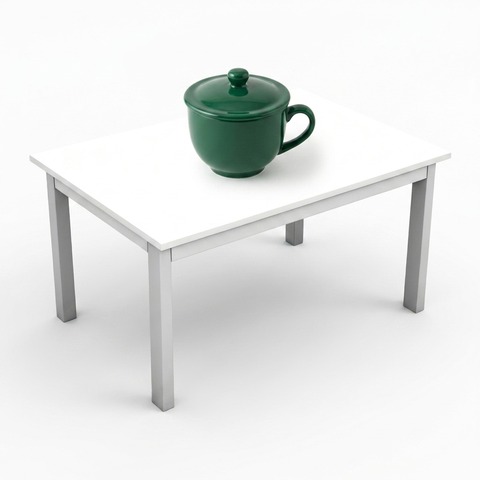} &
\includegraphics[width=0.18\linewidth]{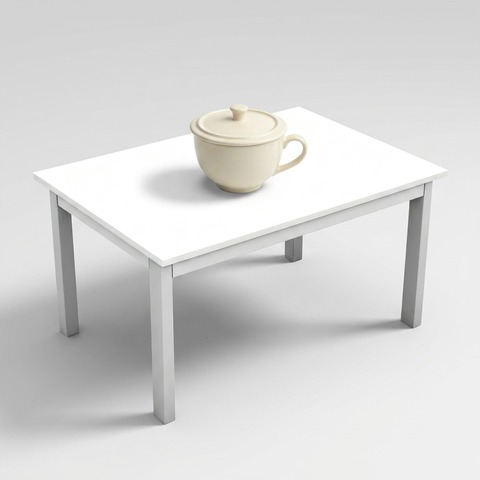} \\
\includegraphics[width=0.18\linewidth]{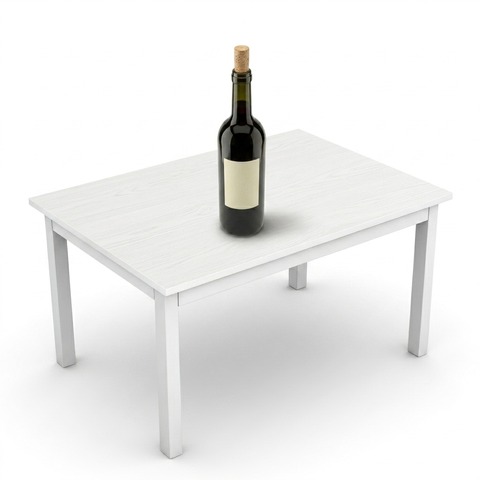} &
\includegraphics[width=0.18\linewidth]{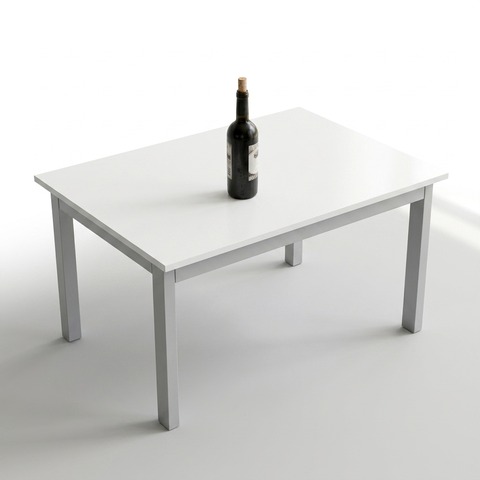} &
\includegraphics[width=0.18\linewidth]{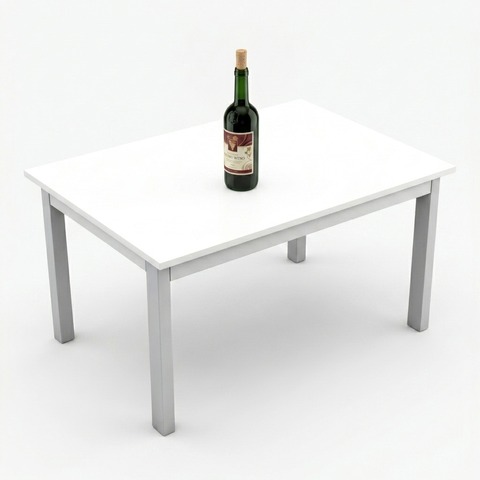} &
\includegraphics[width=0.18\linewidth]{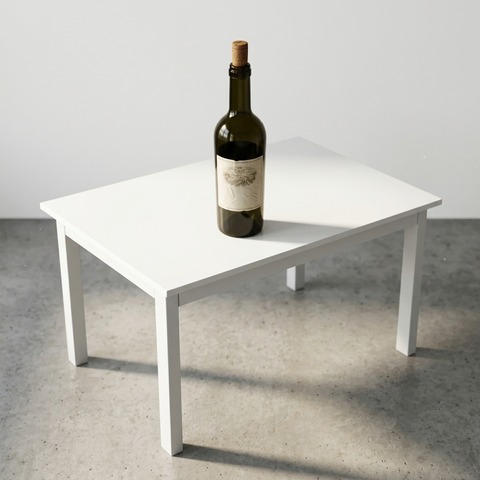} &
\includegraphics[width=0.18\linewidth]{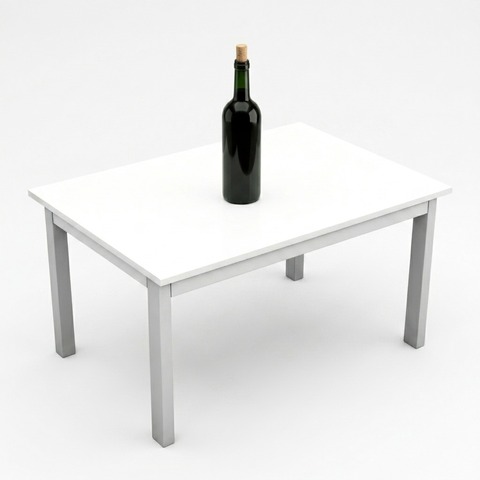} \\
\end{tabular}
\caption{Reference scenes for \textbf{C.2 Capping}. Top to bottom rows show pot-lid, teacup-lid, and wine-cork variants.}
\label{fig:supp:gallery:c2}
\end{figure}

\subsection{C.3 Resting}
\label{sec:supp:tasks:c3}

\noindent\textbf{Task 1: Rest a hat on a mannequin head.}
\begin{itemize}\setlength{\itemsep}{1pt}
    \item The hat is fitted over the crown of the mannequin head with the brim encircling the forehead. The hat is a tan woven straw hat with a medium circular brim and a shallow rounded crown. The mannequin head is a smooth white mannequin head with simplified facial features and a visible neck base.
    \item The hat sits on top of the mannequin head with the bill projecting forward over the face. The hat is a black baseball cap with a curved front bill and a soft rounded crown. The mannequin head is a matte gray mannequin head with minimal facial contours and an upright neck.
    \item The hat covers the upper part of the mannequin head and wraps around the forehead. The hat is a red ribbed knit beanie with a folded cuff and a rounded closed top. The mannequin head is a cream foam mannequin head with a simple nose, chin, and stable flat base.
    \item The hat is seated over the top of the mannequin head with the brim drooping around the sides. The hat is a navy fabric bucket hat with a soft downward brim and stitched crown panels. The mannequin head is a beige mannequin head with smooth facial features and a short cylindrical neck.
    \item The hat rests over the crown of the mannequin head and covers the top surface. The hat is a gray felt fedora with a pinched crown, narrow black band, and structured brim. The mannequin head is a black mannequin head with subtle facial contours, ears, and a flat neck base.
\end{itemize}

\noindent\textbf{Task 2: Rest a cup of coffee on a coffee machine.}
\begin{itemize}\setlength{\itemsep}{1pt}
    \item The cup of coffee sits on the drip tray directly under the spout of the warm white coffee machine. The cup of coffee is one complete small cup filled with dark coffee, visible coffee surface, simple handle, and no spoon or saucer. The coffee machine is one complete warm white compact coffee machine with a visible central spout, drip tray, simple body, and no brand text or extra cup.
    \item The cup of coffee sits on the drip tray directly under the spout of the natural wood coffee machine. The cup of coffee is one complete small cup filled with dark coffee, visible coffee surface, simple handle, and no spoon or saucer. The coffee machine is one complete natural wood compact coffee machine with a visible central spout, drip tray, simple body, and no brand text or extra cup.
    \item The cup of coffee sits on the drip tray directly under the spout of the brushed metal coffee machine. The cup of coffee is one complete small cup filled with dark coffee, visible coffee surface, simple handle, and no spoon or saucer. The coffee machine is one complete brushed metal compact coffee machine with a visible central spout, drip tray, simple body, and no brand text or extra cup.
    \item The cup of coffee sits on the drip tray directly under the spout of the translucent smoky acrylic coffee machine. The cup of coffee is one complete small cup filled with dark coffee, visible coffee surface, simple handle, and no spoon or saucer. The coffee machine is one complete translucent smoky acrylic compact coffee machine with a visible central spout, drip tray, simple body, and no brand text or extra cup.
    \item The cup of coffee sits on the drip tray directly under the spout of the deep green coffee machine. The cup of coffee is one complete small cup filled with dark coffee, visible coffee surface, simple handle, and no spoon or saucer. The coffee machine is one complete deep green compact coffee machine with a visible central spout, drip tray, simple body, and no brand text or extra cup.
\end{itemize}

\noindent\textbf{Task 3: Rest an egg in an egg carton.}
\begin{itemize}\setlength{\itemsep}{1pt}
    \item The single egg is sitting inside one molded cup of the matte black egg carton. The egg is one complete smooth chicken egg, oval shape, pale shell. The egg carton is one empty matte black egg carton section with multiple molded cups, raised dividers, and no other eggs.
    \item The single egg is sitting inside one molded cup of the warm white egg carton. The egg is one complete smooth chicken egg, oval shape, pale shell. The egg carton is one empty warm white egg carton section with multiple molded cups, raised dividers, and no other eggs.
    \item The single egg is sitting inside one molded cup of the brushed metal egg carton. The egg is one complete smooth chicken egg, oval shape, pale shell. The egg carton is one empty brushed metal egg carton section with multiple molded cups, raised dividers, and no other eggs.
    \item The single egg is sitting inside one molded cup of the translucent smoky acrylic egg carton. The egg is one complete smooth chicken egg, oval shape, pale shell. The egg carton is one empty translucent smoky acrylic egg carton section with multiple molded cups, raised dividers, and no other eggs.
    \item The single egg is sitting inside one molded cup of the deep green egg carton. The egg is one complete smooth chicken egg, oval shape, pale shell. The egg carton is one empty deep green egg carton section with multiple molded cups, raised dividers, and no other eggs.
\end{itemize}

\begin{figure}[h]
\centering
\setlength{\tabcolsep}{1pt}
\renewcommand{\arraystretch}{0}
\begin{tabular}{ccccc}
\includegraphics[width=0.18\linewidth]{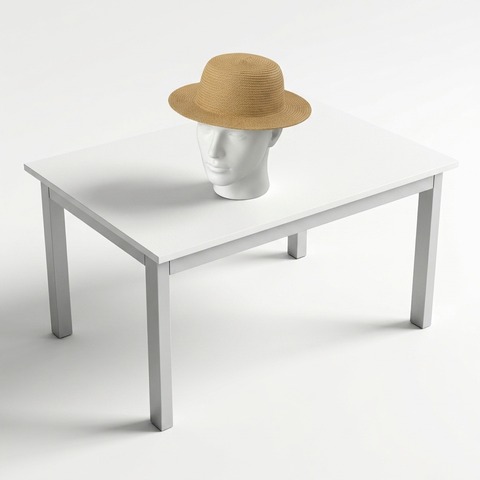} &
\includegraphics[width=0.18\linewidth]{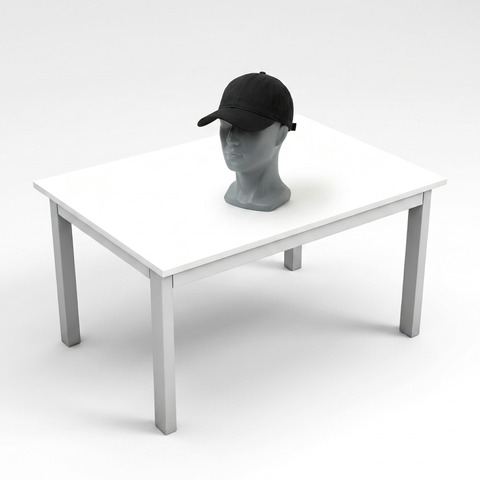} &
\includegraphics[width=0.18\linewidth]{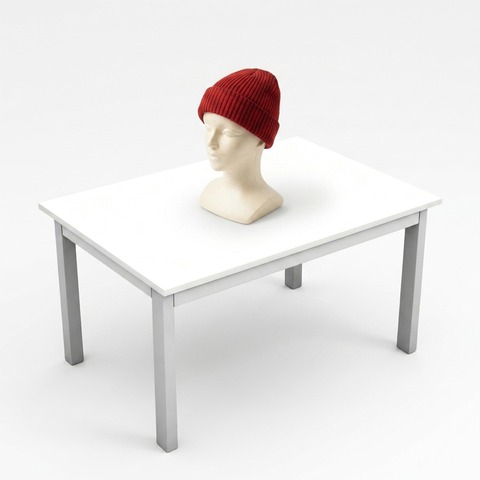} &
\includegraphics[width=0.18\linewidth]{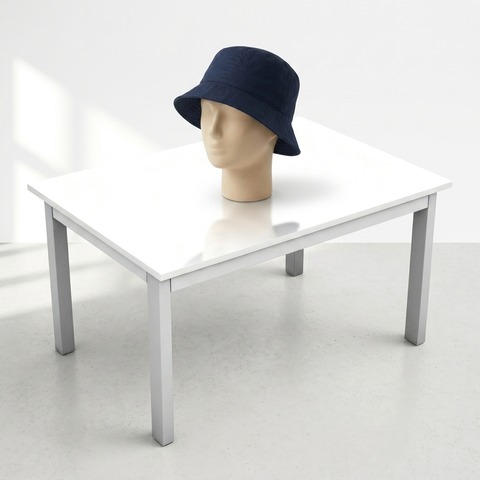} &
\includegraphics[width=0.18\linewidth]{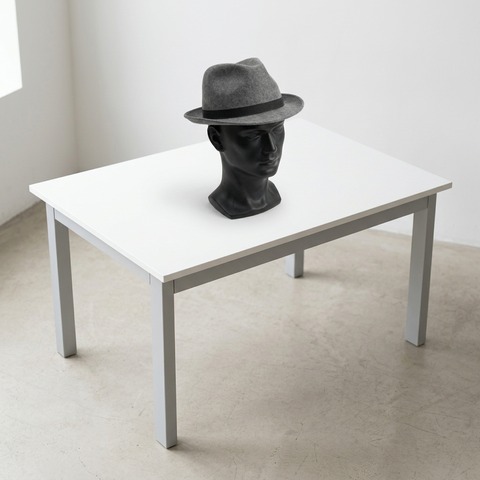} \\
\includegraphics[width=0.18\linewidth]{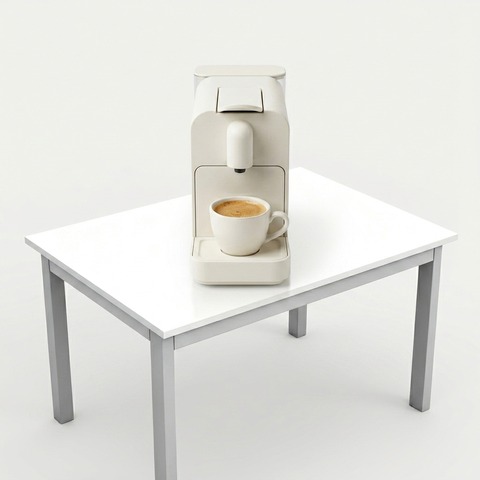} &
\includegraphics[width=0.18\linewidth]{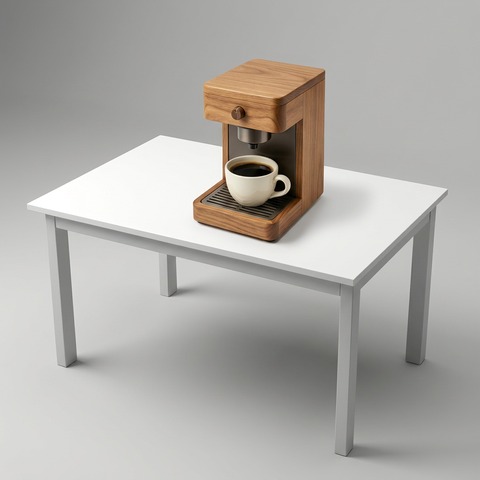} &
\includegraphics[width=0.18\linewidth]{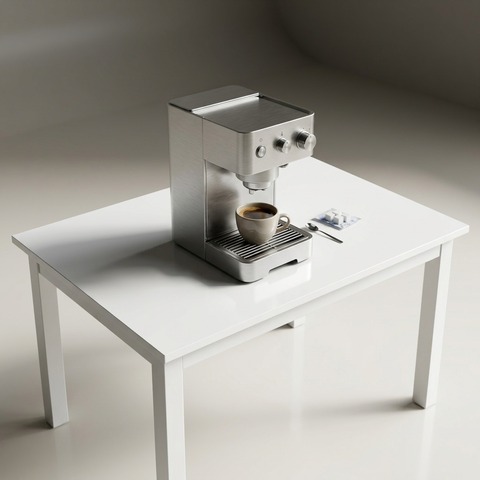} &
\includegraphics[width=0.18\linewidth]{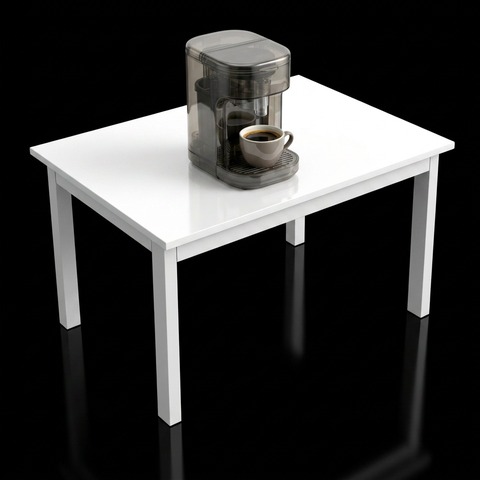} &
\includegraphics[width=0.18\linewidth]{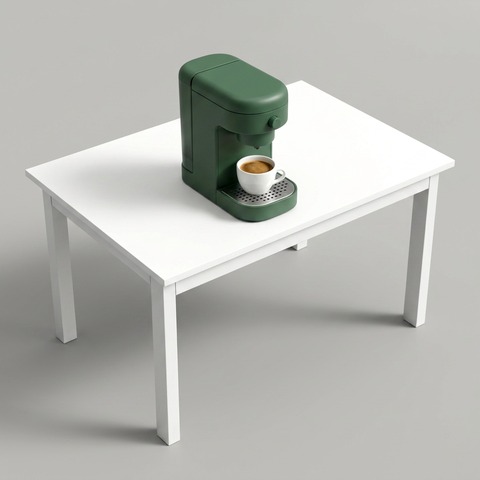} \\
\includegraphics[width=0.18\linewidth]{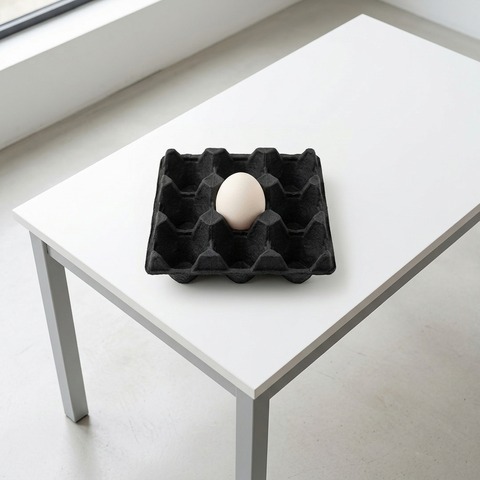} &
\includegraphics[width=0.18\linewidth]{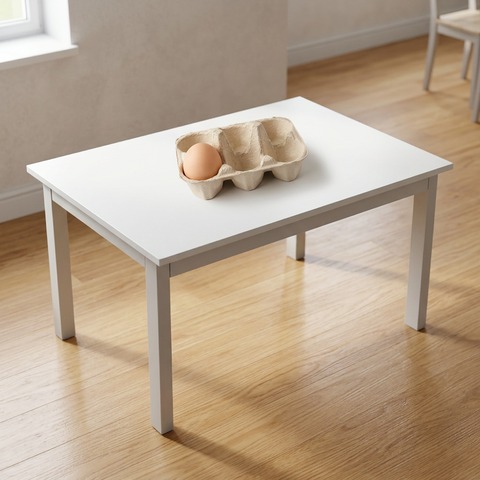} &
\includegraphics[width=0.18\linewidth]{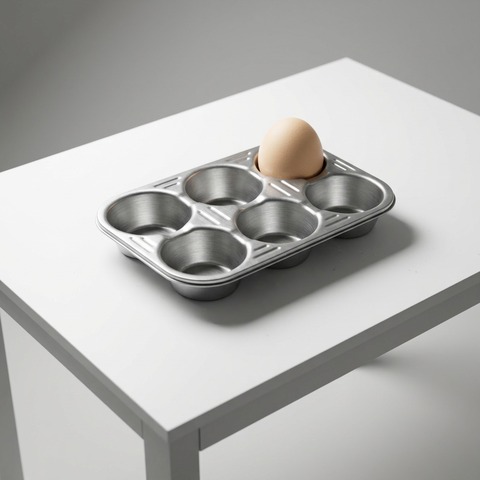} &
\includegraphics[width=0.18\linewidth]{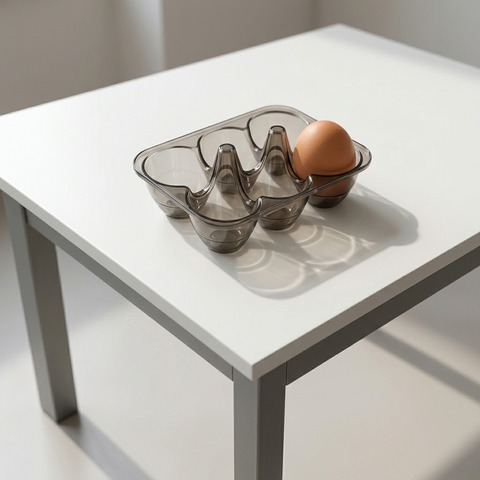} &
\includegraphics[width=0.18\linewidth]{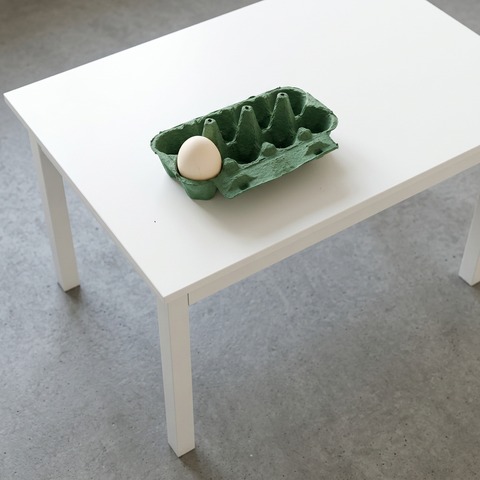} \\
\end{tabular}
\caption{Reference scenes for \textbf{C.3 Resting}. Top to bottom rows show hat-mannequin, coffee-machine, and egg-carton variants.}
\label{fig:supp:gallery:c3}
\end{figure}

\subsection{C.4 Leaning}
\label{sec:supp:tasks:c4}

\noindent\textbf{Task 1: Lean a phone against a phone stand.}
\begin{itemize}\setlength{\itemsep}{1pt}
    \item The phone is upright against the tilted back plate of the phone stand with its bottom edge resting on the support lips. The phone is a slim black smartphone with a blank glossy screen and rounded corners. The phone stand is a silver aluminum desktop phone stand with a tilted back plate, two small front support lips, and a rectangular base.
    \item The phone stands upright on the phone stand with its bottom edge seated on the horizontal ledge behind the front retaining lips and its back leaning against the angled support panel. The phone is a white smartphone with a dark inactive glass screen, a thin white bezel, rounded rectangular corners. The phone stand is a realistic compact black folding plastic desktop phone stand with two triangular side supports, a hinged angled back panel, a horizontal lower ledge, small front retaining lips, and a stable base.
    \item The phone is seated in the slot of the phone stand and leans back against the wooden support. The phone is a smartphone in a muted green protective case with a flat dark glass screen. The phone stand is a light wood phone stand with a slotted base and a vertical angled backing board.
    \item The phone is supported by the front prongs of the phone stand and rests against the angled support. The phone is a large dark gray smartphone with a blank reflective screen and rounded edges. The phone stand is a white adjustable phone stand with a hinged vertical support, two front prongs, and a broad base.
    \item The phone rests on the curved lower shelf of the transparent phone stand, held by the molded front lip, while the back of the phone leans against the clear slanted panel. The phone is a blue smartphone with a black inactive glass screen, slim side buttons, rounded corners. The phone stand is a realistic transparent acrylic desktop phone stand made from a single bent clear sheet, with a slanted back panel, broad stable base, curved lower shelf, and molded front lip that prevents slipping.
\end{itemize}

\noindent\textbf{Task 2: Lean a shoe against a sloped shoe display stand.}
\begin{itemize}\setlength{\itemsep}{1pt}
    \item The cream sneaker rests along the dark charcoal sloped ramp, with its sole flat against the incline and its front outsole blocked by the vertical front stop wall. The shoe is one complete cream athletic sneaker with beige suede panels, white laces, a padded collar, and a gum-colored outsole. The sloped shoe display stand is one dark charcoal wedge-shaped shoe display stand with a broad inclined top ramp and a vertical rectangular stop wall at the lower front edge, the stop wall touching the front outsole to prevent sliding.
    \item The black sneaker leans back on the white curved sloped surface while the vertical front stop wall catches the lower sole and prevents the shoe from sliding down. The shoe is one complete black low-top sneaker with black laces, smooth leather-like upper, rounded toe, and white midsole. The sloped shoe display stand is one glossy white sloped shoe display stand with a gently curved concave ramp surface and a short vertical front stop wall rising at the lower edge, stable rectangular base.
    \item The red running shoe rests on the brushed aluminum incline, and the vertical metal front stop plate supports the toe area so the shoe cannot slide forward. The shoe is one complete red running shoe with breathable mesh texture, black laces, sculpted white midsole, and black outsole. The sloped shoe display stand is one brushed aluminum sloped display stand with a triangular side profile, flat inclined ramp, and a vertical metal stop plate at the low front edge that is tall enough to block the shoe toe.
    \item The pastel pink sneaker rests on the clear acrylic sloped ramp, and the transparent vertical front stop panel blocks the lower sole from sliding forward. The shoe is one complete pastel pink sneaker with matching laces, soft fabric panels, rounded toe, and pale cream sole. The sloped shoe display stand is one clear translucent acrylic shoe display stand with a smooth sloped ramp, visible transparent side edges, and a vertical clear front stop panel at the low edge.
    \item The burgundy sneaker rests on the cream triangular sloped plane, and the raised vertical stop wall at the lower front edge prevents the shoe from sliding down. The shoe is one complete burgundy sneaker with dark red laces, soft suede-like upper, rounded toe, and white chunky sole. The sloped shoe display stand is one cream colored triangular wedge shoe display stand with a sloped top plane and a raised vertical stop wall at the lower front edge.
\end{itemize}

\noindent\textbf{Task 3: Lean a bamboo steamer lid against a bamboo steamer basket.}
\begin{itemize}\setlength{\itemsep}{1pt}
    \item The bamboo steamer lid is slightly open at front and leaning against the rim of the empty bamboo steamer basket. The bamboo steamer lid is one complete circular bamboo steamer lid with woven top pattern, raised rim, small central handle or cross rib, clean and empty. The bamboo steamer basket is one complete empty round pale bamboo bamboo steamer basket with cylindrical side wall, visible open interior, woven base.
    \item The lid is slanted and leaning against the rim of the empty basket, with the lid lower edge resting on the basket rim and side wall so it is physically supported rather than floating. The bamboo steamer lid is one complete circular bamboo steamer lid with a deep cinnabar red stained bamboo rim, pale woven bamboo center, thin gold-toned band around the raised rim, small central cross rib. The bamboo steamer basket is one complete empty round bamboo steamer basket with a deep cinnabar red stained cylindrical side wall, thin gold-toned bands near top and bottom, visible open interior with pale woven bamboo base.
    \item The lid leans off-center at a diagonal angle against the basket rim, with a clear contact point between the lid edge and the basket's upper rim while the basket interior remains visible. The bamboo steamer lid is one complete circular bamboo steamer lid with a muted jade green painted bamboo rim, a woven top panel using alternating pale straw and light green strips, raised outer rim, simple cross rib handle. The bamboo steamer basket is one complete empty round bamboo steamer basket with muted jade green painted cylindrical side wall, pale bamboo interior, visible woven base, slightly darker green binding bands.
    \item The bamboo steamer lid leans at about $45^{\circ}$ against the rim of the empty basket, with the lower rim of the lid touching the basket rim and the lid partially over but not covering the whole opening. The bamboo steamer lid is one complete circular bamboo steamer lid with a deep indigo-blue dyed bamboo rim, pale woven center panel, and small white geometric triangle motifs around the raised rim, small central cross rib. The bamboo steamer basket is one complete empty round bamboo steamer basket with deep indigo-blue dyed cylindrical side wall decorated with small white geometric triangle motifs, visible open interior and pale woven base.
    \item The lid is slanted over the open steamer and leaning against the upper rim, with the lid edge clearly supported by the basket rim and side wall and the empty interior still visible. The bamboo steamer lid is one complete circular bamboo steamer lid with a soft blue-gray dyed bamboo outer rim, pale woven center panel, raised rim, small central cross rib. The bamboo steamer basket is one complete empty round bamboo steamer basket with soft blue-gray dyed cylindrical side wall, pale bamboo interior and woven base, subtle darker blue-gray edge bands.
\end{itemize}

\begin{figure}[h]
\centering
\setlength{\tabcolsep}{1pt}
\renewcommand{\arraystretch}{0}
\begin{tabular}{ccccc}
\includegraphics[width=0.18\linewidth]{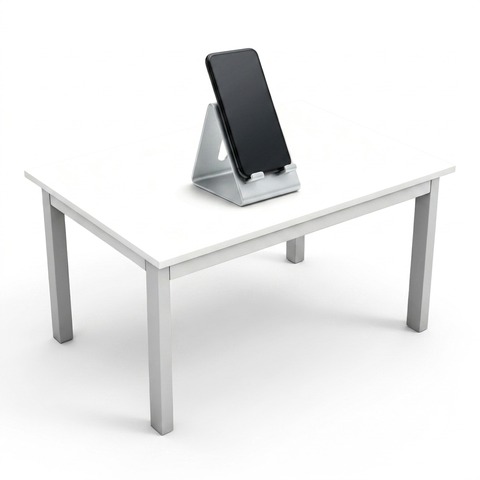} &
\includegraphics[width=0.18\linewidth]{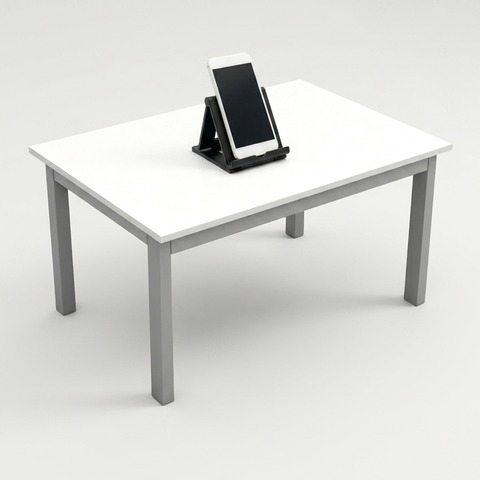} &
\includegraphics[width=0.18\linewidth]{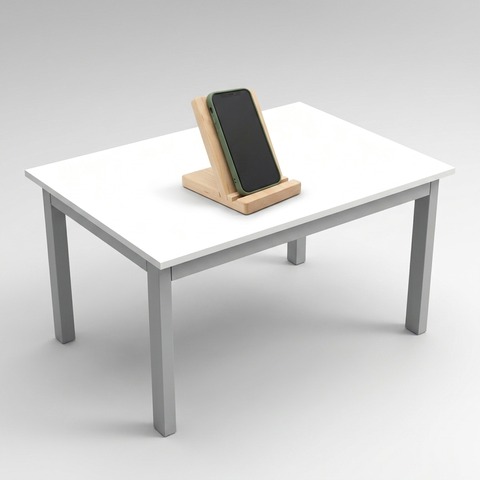} &
\includegraphics[width=0.18\linewidth]{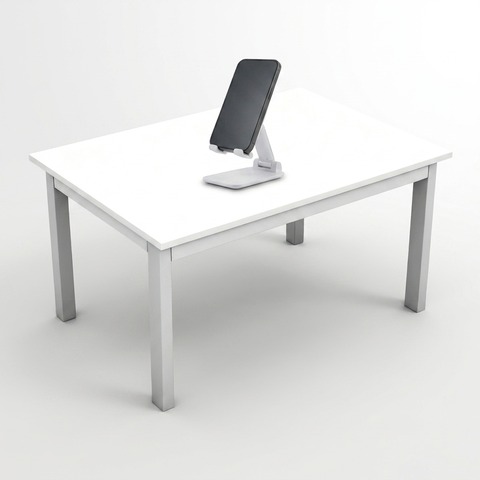} &
\includegraphics[width=0.18\linewidth]{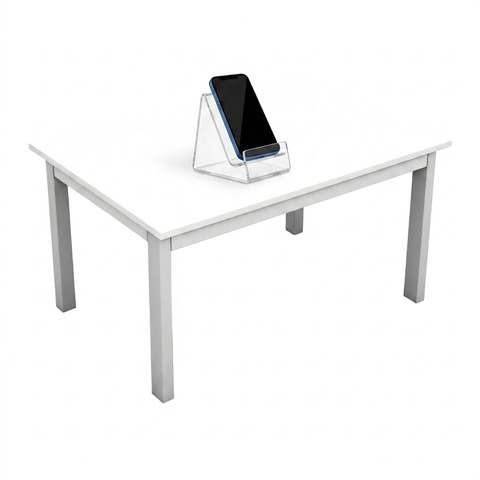} \\
\includegraphics[width=0.18\linewidth]{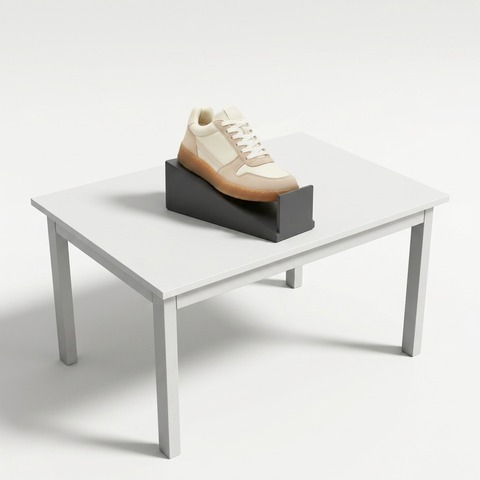} &
\includegraphics[width=0.18\linewidth]{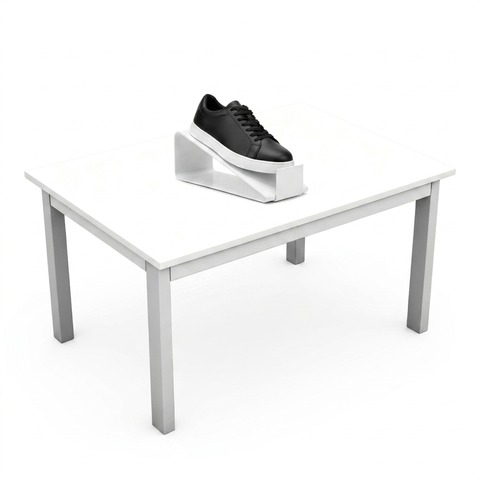} &
\includegraphics[width=0.18\linewidth]{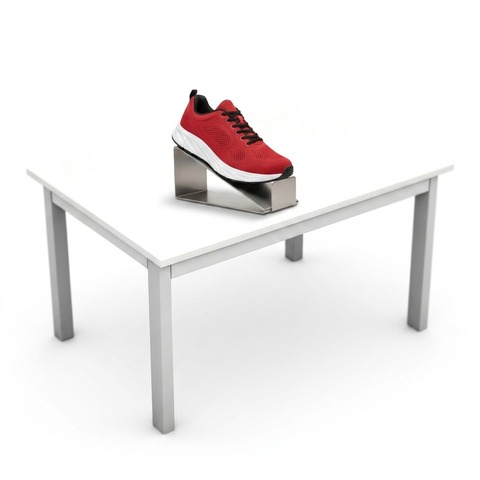} &
\includegraphics[width=0.18\linewidth]{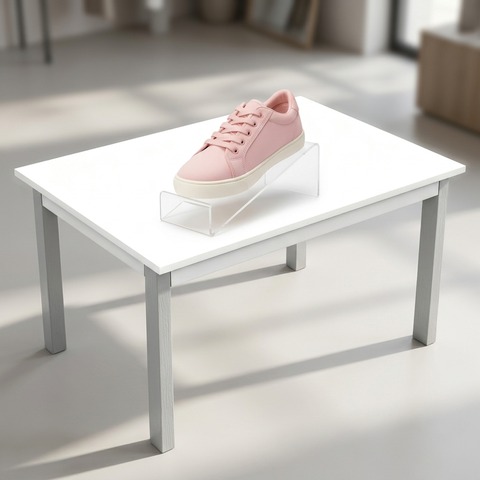} &
\includegraphics[width=0.18\linewidth]{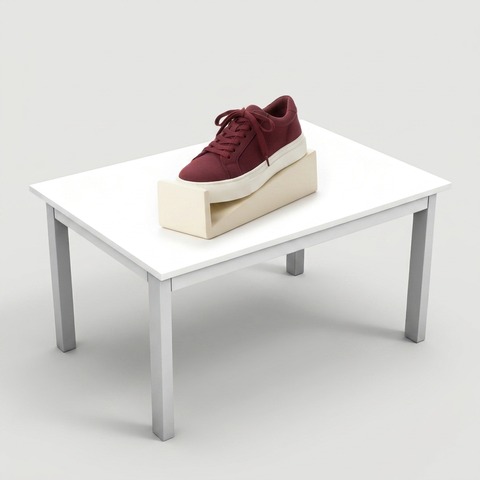} \\
\includegraphics[width=0.18\linewidth]{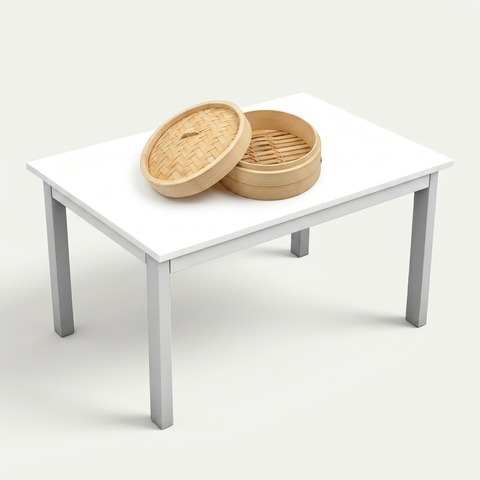} &
\includegraphics[width=0.18\linewidth]{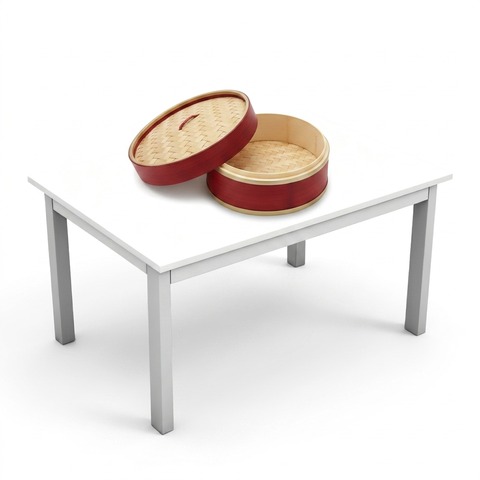} &
\includegraphics[width=0.18\linewidth]{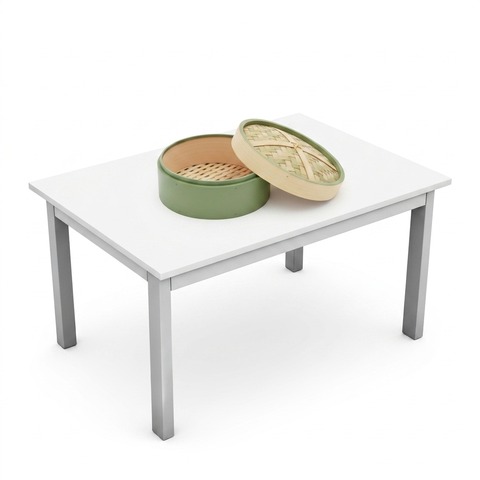} &
\includegraphics[width=0.18\linewidth]{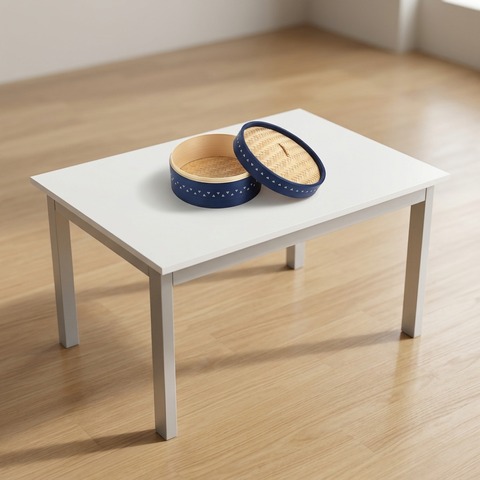} &
\includegraphics[width=0.18\linewidth]{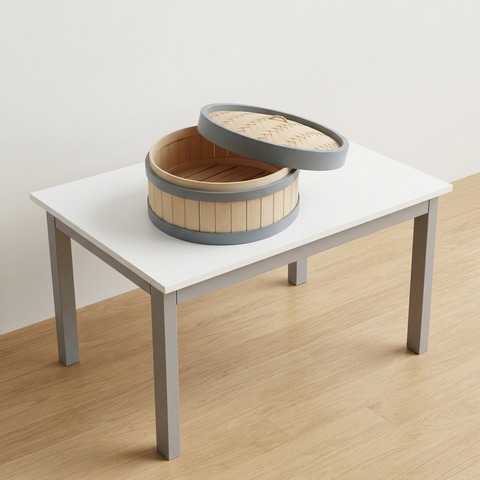} \\
\end{tabular}
\caption{Reference scenes for \textbf{C.4 Leaning}. Top to bottom rows show phone-stand, shoe-stand, and bamboo-steamer-lid variants.}
\label{fig:supp:gallery:c4}
\end{figure}

\subsection{C.5 Hooking}
\label{sec:supp:tasks:c5}

\noindent\textbf{Task 1: Hook a cup on a cup-rack tree.}
\begin{itemize}\setlength{\itemsep}{1pt}
    \item The cup handle is looped over one angled peg of the cup rack tree so the cup hangs freely beside the pole. The cup is a white ceramic mug with a round open top, smooth cylindrical body, and a C-shaped side handle. The cup rack tree is a light wooden cup rack tree with a central vertical pole, a round base, and several short upward angled pegs.
    \item The cup is suspended from one hook of the cup rack tree by its handle. The cup is a glossy blue ceramic cup with a thick side handle and a slightly flared rim. The cup rack tree is a black metal cup rack tree with a circular base, a straight center post, and four curved hook-like arms.
    \item The cup handle hangs over a horizontal peg of the cup rack tree while the cup body remains below the peg. The cup is a yellow ceramic mug with a solid handle, cylindrical body, and thick bottom. The cup rack tree is a shiny chrome cup rack tree with a weighted round base and multiple short horizontal pegs.
    \item The cup is hanging from a short peg of the cup rack tree through its small handle. The cup is a small cream espresso cup with a tiny loop handle, round body, and open top. The cup rack tree is a bamboo cup rack tree with a square base, one vertical stem, and short staggered pegs.
    \item The cup handle is hooked onto one peg of the cup rack tree and the cup hangs downward from it. The cup is a red ceramic mug with a smooth rounded body, open circular mouth, and large side handle. The cup rack tree is a white metal cup rack tree with a stable round base, central upright pole, and six rounded pegs.
\end{itemize}

\noindent\textbf{Task 2: Hook a hat on a coat rack.}
\begin{itemize}\setlength{\itemsep}{1pt}
    \item The hat is hanging from the outer part of exactly one long straight upper hook on the matte black coat rack; at least half of that hook shaft and its tip remain visible outside the hat, and all other hooks are empty and do not touch, support, pierce, or overlap the hat. The hat is one complete hat with a rounded crown and visible brim, soft fabric texture. The coat rack is one matte black coat rack with a stable vertical post, one prominent long straight upper hook or peg projecting from the post to hold the hat.
    \item The medium-small hat hangs from only the outer end of one curved hook on the warm white coat rack; more than half of that hook, including its root and rounded tip, is visible, and all other hooks are visible, empty, and not touching the hat. The hat is one complete medium-small hat with a rounded crown and visible brim, soft fabric texture. The coat rack is one warm white coat rack with a stable vertical post, several visible empty curved hooks near the top, and a simple base.
    \item The hat is hanging from the outer part of exactly one long straight upper hook on the natural wood coat rack. The hat is one complete hat with a rounded crown and visible brim, soft fabric texture. The coat rack is one natural wood coat rack with a stable vertical post, one prominent long straight upper hook projecting from the post to hold the hat.
    \item The medium-small hat hangs from the outer part of one translucent curved hook, leaving most of the hook visible and all other hooks empty, visible, and not in contact with the hat. The hat is one complete medium-small hat with a rounded crown and visible brim, soft fabric texture. The coat rack is one translucent smoky acrylic coat rack with a stable vertical post, several visible curved hooks, and a simple base.
    \item The hat is hanging from the outer part of exactly one long straight upper hook on the light gray coat rack. The hat is one complete hat with a rounded crown and visible brim, soft fabric texture. The coat rack is one light gray coat rack with a stable vertical post, one prominent long straight upper hook projecting from the post to hold the hat.
\end{itemize}

\noindent\textbf{Task 3: Hook a soup spoon on a hanging spoon hook.}
\begin{itemize}\setlength{\itemsep}{1pt}
    \item The soup spoon hangs from the spoon hook by its handle hole or hanging loop, with the hook supporting the spoon vertically. The soup spoon is one complete stainless steel soup ladle with a round bowl and long handle, hanging vertically by the handle end. The spoon hook is one small wall-mounted hook holding the soup ladle by its handle end.
    \item The soup spoon hangs from the spoon hook by its handle hole or hanging loop, with the hook supporting the spoon vertically. The soup spoon is one complete stainless steel soup ladle with a round bowl and long handle, hanging vertically from a rail hook. The spoon hook is one wall-mounted horizontal hook rail with several hooks, exactly one hook holding the soup ladle and the other hooks empty.
    \item The soup spoon hangs from the spoon hook by its handle hole or hanging loop. The soup spoon is one complete stainless steel soup ladle with a round bowl and long handle, hanging vertically by the handle end. The spoon hook is one small wall-mounted hook holding the soup ladle by its handle end.
    \item The soup spoon hangs from the spoon hook by its handle hole or hanging loop. The soup spoon is one complete black soup spoon with a deep round bowl and long handle, hanging vertically by the handle end. The spoon hook is one small wall-mounted hook holding the black soup spoon by its handle end.
    \item The soup spoon hangs from the spoon hook by its handle hole or hanging loop. The soup spoon is one complete matte black soup spoon with a deep bowl and long handle, hanging vertically by the handle end. The spoon hook is one square wall-mounted hook holding the black soup spoon by its handle end.
\end{itemize}

\begin{figure}[h]
\centering
\setlength{\tabcolsep}{1pt}
\renewcommand{\arraystretch}{0}
\begin{tabular}{ccccc}
\includegraphics[width=0.18\linewidth]{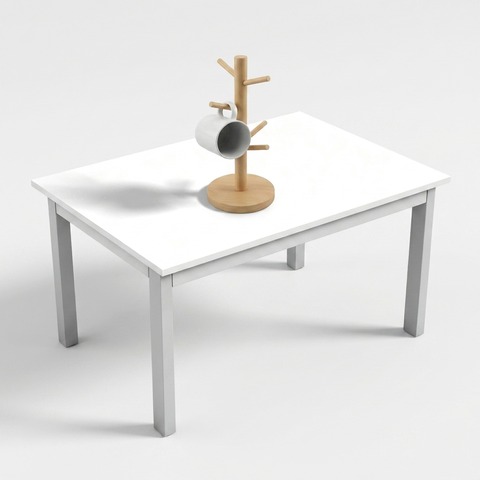} &
\includegraphics[width=0.18\linewidth]{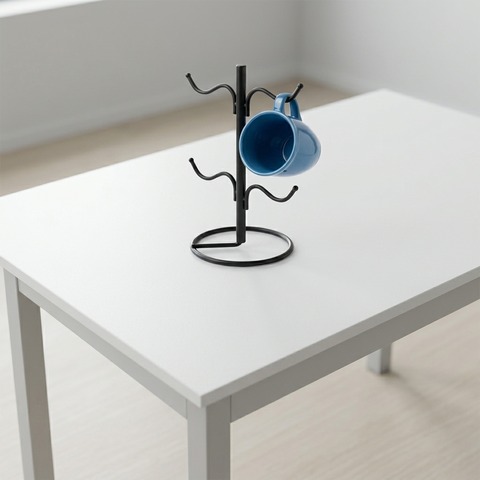} &
\includegraphics[width=0.18\linewidth]{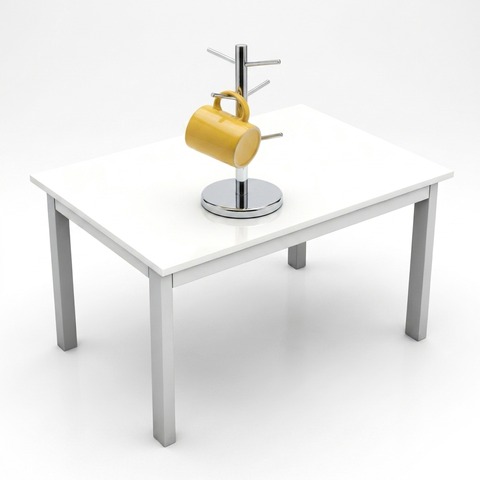} &
\includegraphics[width=0.18\linewidth]{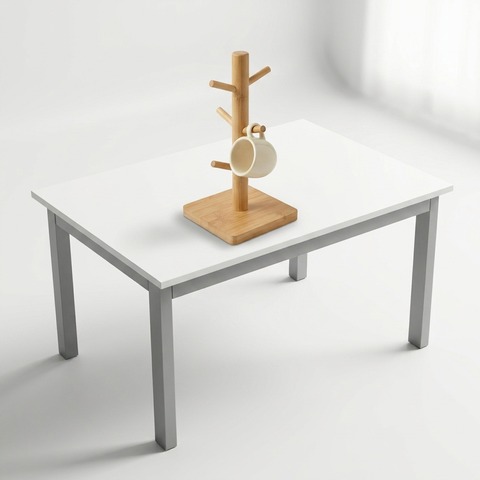} &
\includegraphics[width=0.18\linewidth]{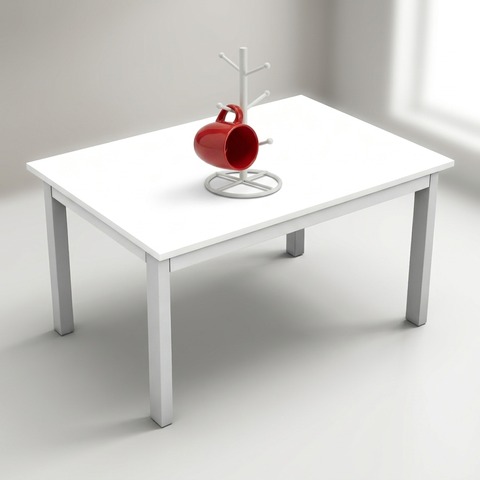} \\
\includegraphics[width=0.18\linewidth]{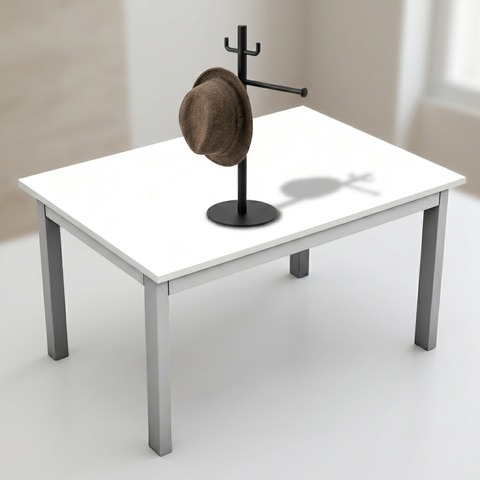} &
\includegraphics[width=0.18\linewidth]{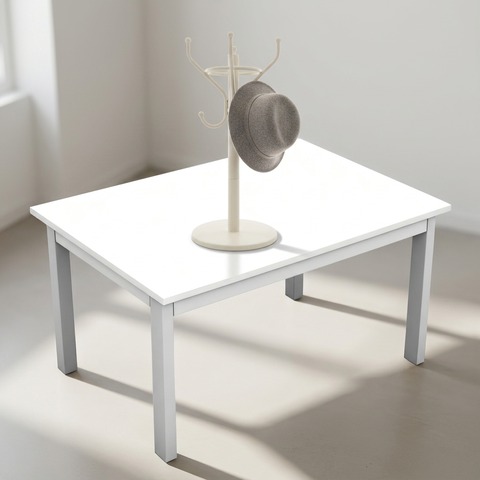} &
\includegraphics[width=0.18\linewidth]{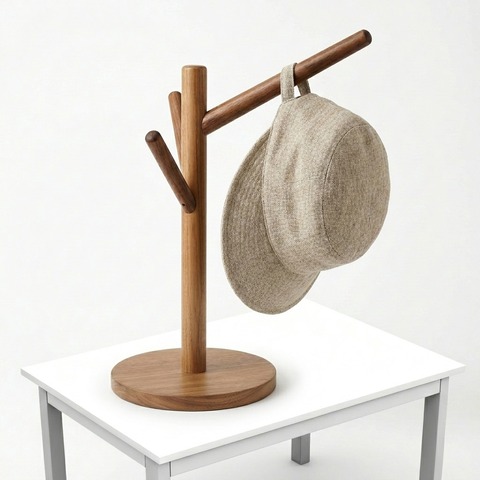} &
\includegraphics[width=0.18\linewidth]{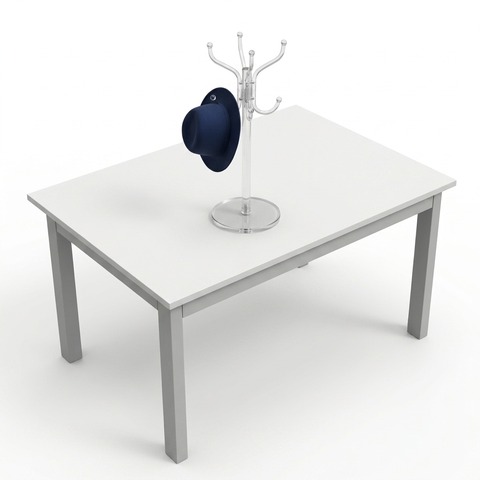} &
\includegraphics[width=0.18\linewidth]{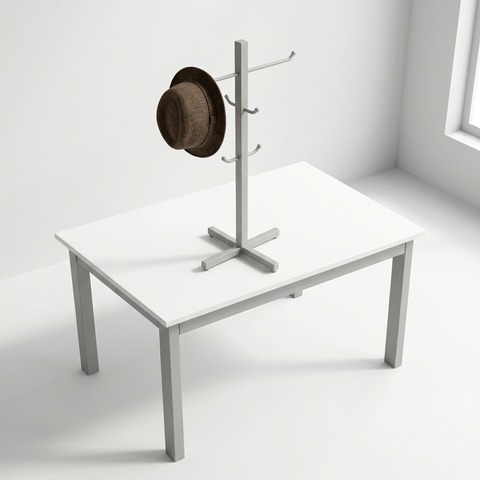} \\
\includegraphics[width=0.18\linewidth]{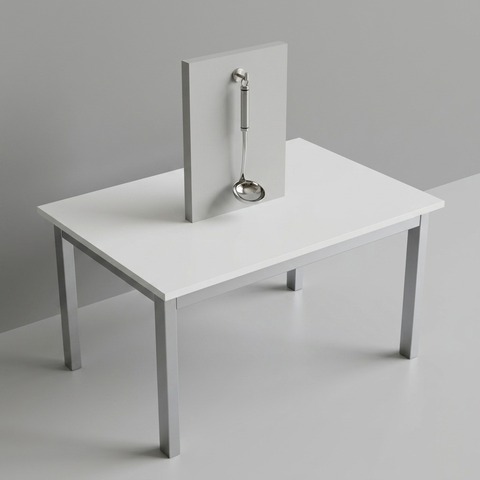} &
\includegraphics[width=0.18\linewidth]{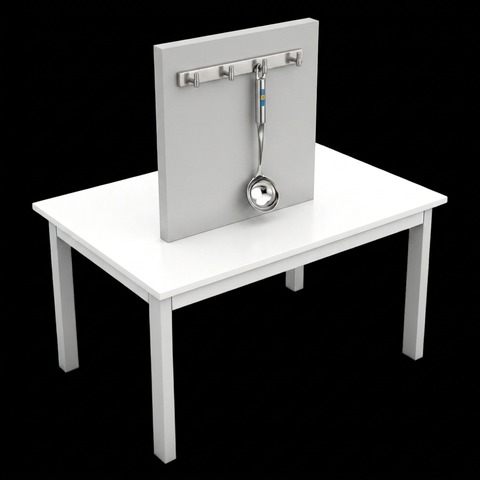} &
\includegraphics[width=0.18\linewidth]{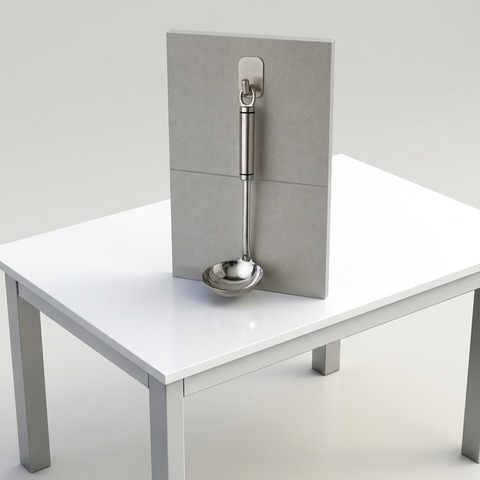} &
\includegraphics[width=0.18\linewidth]{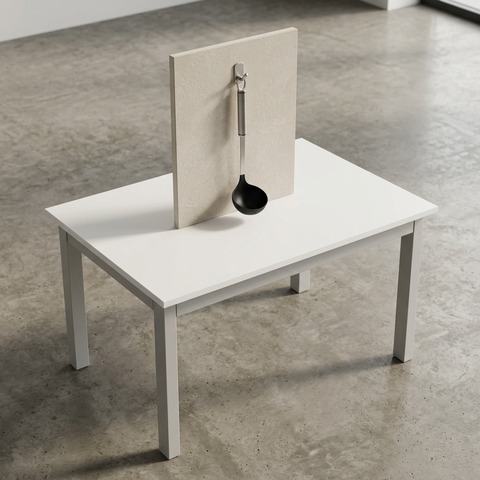} &
\includegraphics[width=0.18\linewidth]{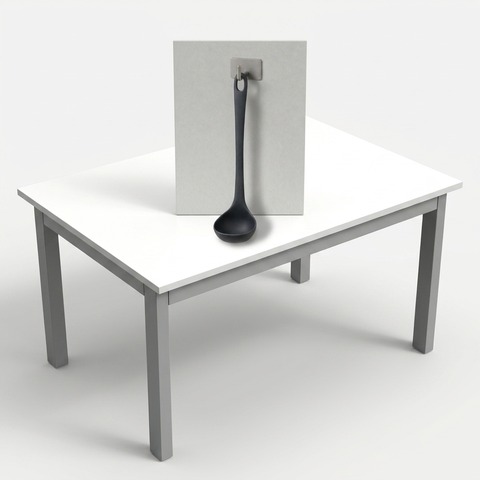} \\
\end{tabular}
\caption{Reference scenes for \textbf{C.5 Hooking}. Top to bottom rows show cup-rack-tree, hat-coat-rack, and soup-spoon-hook variants.}
\label{fig:supp:gallery:c5}
\end{figure}

\subsection{C.6 Slotting}
\label{sec:supp:tasks:c6}

\noindent\textbf{Task 1: Slot a plate into a dish rack.}
\begin{itemize}\setlength{\itemsep}{1pt}
    \item The plate stands nearly vertical in one slot between two upright bamboo pegs of the dish rack, with its lower curved edge resting between the two open parallel bamboo base rails. The plate is a round white ceramic dinner plate with a shallow raised rim, glossy surface, complete circular outline. The dish rack is a realistic narrow single-row bamboo tabletop dish rack like a simple plate stand, made from two straight parallel bamboo base rails connected by short cross rods, with one line of evenly spaced upright round bamboo pegs.
    \item The plate is inserted vertically into one slot of the dish rack and rests against the wooden dividers. The plate is a round matte blue ceramic plate with a raised rim and flat center. The dish rack is a bamboo dish rack with evenly spaced vertical wooden slots and a simple rectangular frame.
    \item The square plate is held upright in one slot between two round bamboo pegs of the dish rack, with its bottom edge resting between the low parallel wooden base rails and its face leaning lightly against an adjacent peg. The plate is a square white porcelain plate with softly rounded corners, a shallow raised edge, a glossy face. The dish rack is a realistic single-row bamboo tabletop dish rack with two long parallel wooden base rails, evenly spaced upright round bamboo pegs, a low open frame, small wooden feet.
    \item The plate is slotted upright between two round bamboo pegs of the dish rack, with its lower rim resting between the two open parallel wooden base rails. The plate is a pale green shallow ceramic plate with a slightly concave center, smooth rounded rim, glossy finish. The dish rack is a realistic single-row bamboo tabletop dish rack with two long parallel wooden base rails, evenly spaced upright round bamboo pegs, a simple low rectangular frame.
    \item The plate is inserted into the dish rack. The plate is a round ivory ceramic plate with a thin dark rim pattern, glossy face, shallow raised lip. The dish rack is a single-row bamboo tabletop dish rack with two long parallel wooden base rails and evenly spaced round vertical pegs, an open low frame.
\end{itemize}

\noindent\textbf{Task 2: Slot a red wine bottle into a wine rack.}
\begin{itemize}\setlength{\itemsep}{1pt}
    \item The red wine bottle is seated in the wine rack, supported by the rack structure so it cannot roll away. The red wine bottle is one complete dark glass red wine bottle with a cream label and burgundy neck capsule, seated through one diamond opening of the rack. The wine rack is one black metal diamond-shaped wire wine rack with crossed open cells, one cell supporting the bottle.
    \item The red wine bottle is seated in the wine rack, lying horizontally in one upper groove of the rack. The red wine bottle is one complete dark glass red wine bottle with a cream label and burgundy neck capsule. The wine rack is one light wooden rectangular wine rack with horizontal rails and U-shaped cradle grooves, one upper groove supporting the bottle.
    \item The red wine bottle is seated in the wine rack, supported by the rack structure so it cannot roll away. The red wine bottle is one complete dark glass red wine bottle with a cream label and burgundy neck capsule, seated diagonally in one crossed wooden cell of the rack. The wine rack is one light wooden crossed-board wine rack with diagonal boards forming open cells, one cell supporting the bottle.
    \item The red wine bottle is seated in the wine rack, supported in the curved wire holder. The red wine bottle is one complete dark glass red wine bottle with a cream label and burgundy neck capsule. The wine rack is one black decorative metal wire wine rack with rounded loop supports, one support holding the bottle.
    \item The red wine bottle is seated in the wine rack, lying horizontally in one cell of the wooden grid rack. The red wine bottle is one complete dark glass red wine bottle with a cream label and burgundy neck capsule. The wine rack is one pale wooden grid wine rack with rectangular cells and horizontal supports, one cell supporting the bottle.
\end{itemize}

\noindent\textbf{Task 3: Slot a book into a multi-compartment book stand.}
\begin{itemize}\setlength{\itemsep}{1pt}
    \item The book is placed upright inside exactly one rectangular compartment of the multi-compartment book stand, fully contained within that compartment and not touching the dividers of adjacent compartments. The book is one complete closed book with a plain cover and a visible spine. The multi-compartment book stand is a realistic natural wood multi-compartment book stand with a stable base and several rectangular grid compartments separated by vertical dividers; at least one compartment is clearly sized so a single upright book fits within it.
    \item The closed book is placed upright inside exactly one rectangular compartment of the dark ceramic multi-compartment book stand, fitting entirely within that compartment without occupying neighboring compartments. The book is one complete closed book with a plain dark cover and visible spine. The multi-compartment book stand is one dark ceramic multi-compartment book stand with a stable base and clearly separated rectangular compartments separated by vertical divider walls.
    \item The book stands upright inside one rectangular compartment of the translucent smoky acrylic multi-compartment book stand. The book is one complete closed book with plain cover and visible spine. The multi-compartment book stand is one translucent smoky acrylic multi-compartment book stand with several empty rectangular grid slots, vertical dividers, and stable base.
    \item The closed book is placed upright inside exactly one compartment of the brass multi-compartment book stand, contained fully within that single slot and not spanning across the divider into neighboring compartments. The book is one complete closed book with a plain cover and visible spine. The multi-compartment book stand is one brass colored multi-compartment book stand with several clearly separated rectangular compartments, vertical dividers, and a stable base.
    \item The book stands upright inside one rectangular compartment of the light gray multi-compartment book stand. The book is one complete closed book with plain cover and visible spine. The multi-compartment book stand is one light gray multi-compartment book stand with several empty rectangular grid slots, vertical dividers, and stable base.
\end{itemize}

\begin{figure}[h]
\centering
\setlength{\tabcolsep}{1pt}
\renewcommand{\arraystretch}{0}
\begin{tabular}{ccccc}
\includegraphics[width=0.18\linewidth]{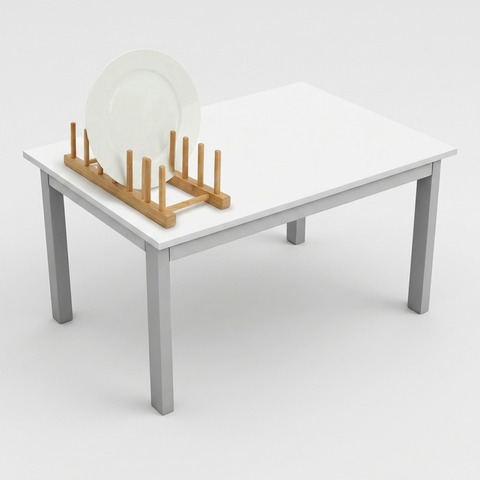} &
\includegraphics[width=0.18\linewidth]{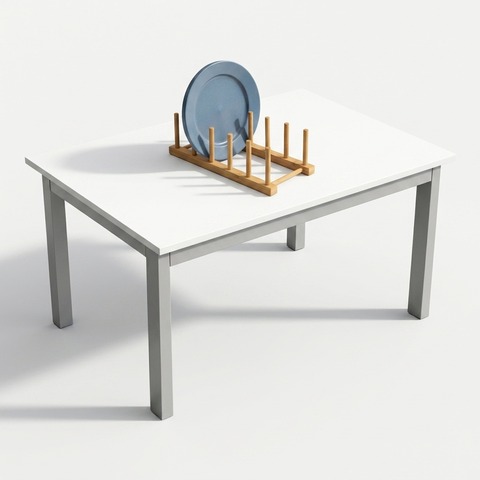} &
\includegraphics[width=0.18\linewidth]{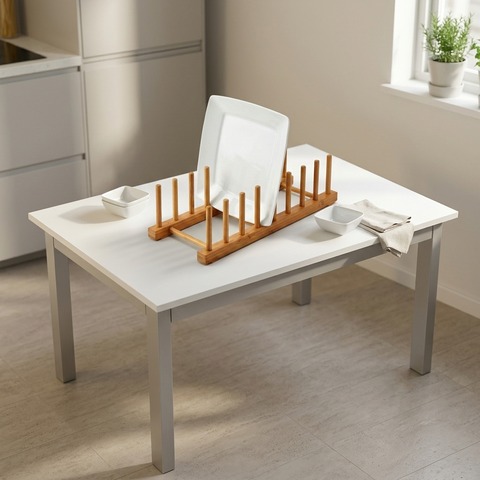} &
\includegraphics[width=0.18\linewidth]{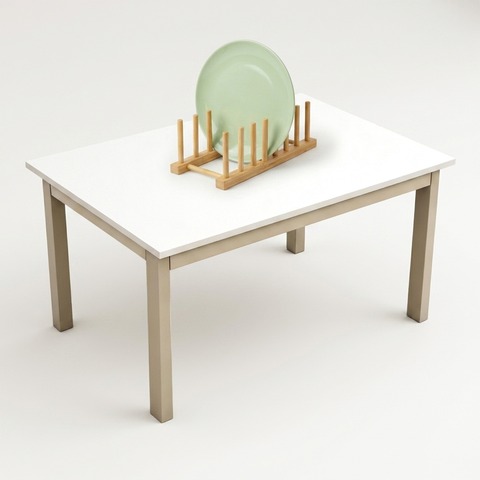} &
\includegraphics[width=0.18\linewidth]{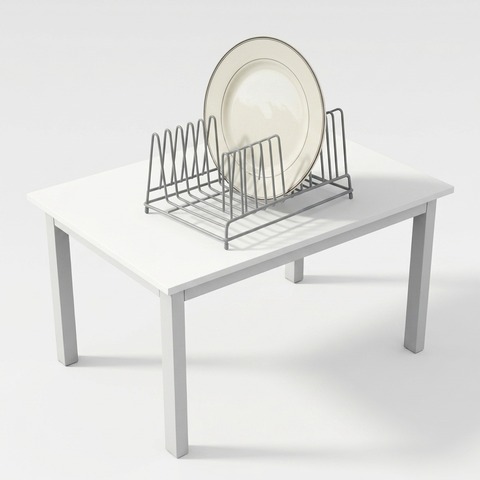} \\
\includegraphics[width=0.18\linewidth]{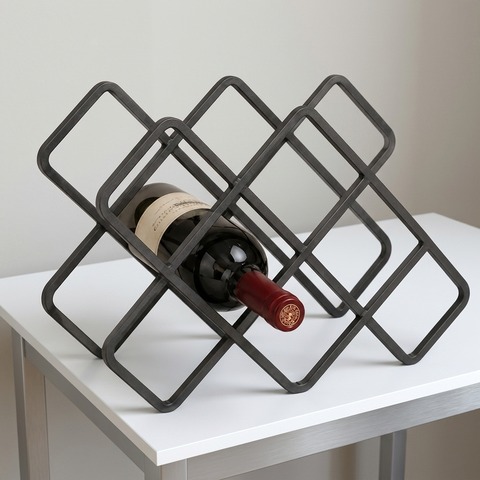} &
\includegraphics[width=0.18\linewidth]{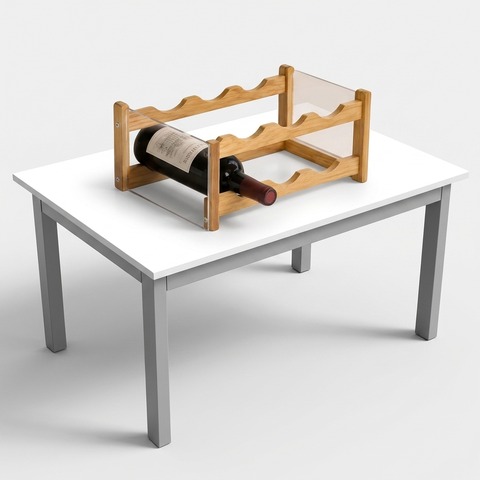} &
\includegraphics[width=0.18\linewidth]{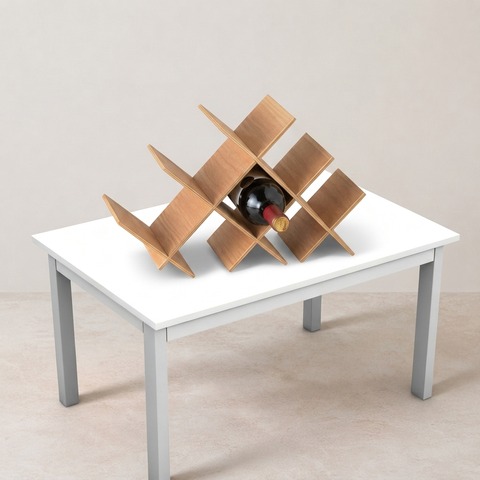} &
\includegraphics[width=0.18\linewidth]{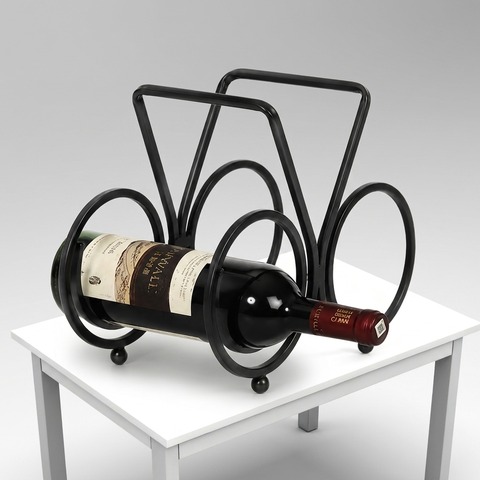} &
\includegraphics[width=0.18\linewidth]{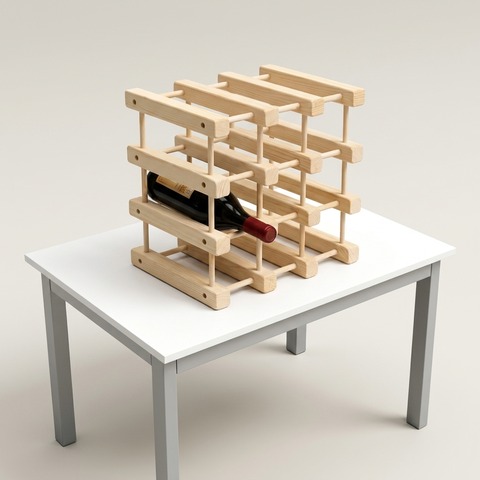} \\
\includegraphics[width=0.18\linewidth]{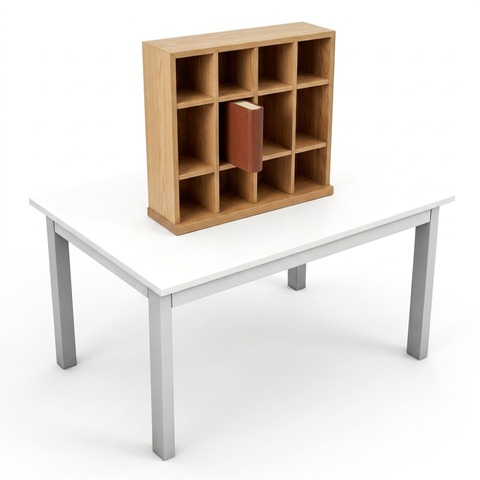} &
\includegraphics[width=0.18\linewidth]{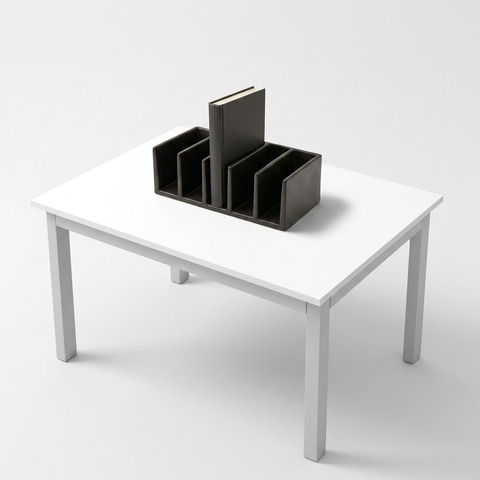} &
\includegraphics[width=0.18\linewidth]{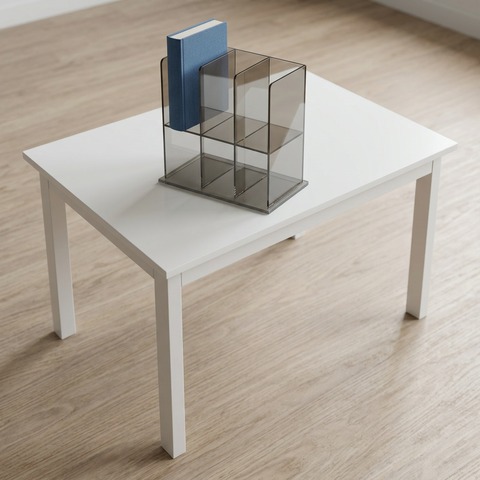} &
\includegraphics[width=0.18\linewidth]{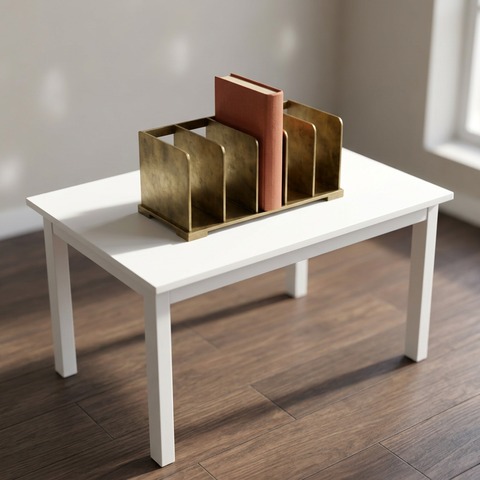} &
\includegraphics[width=0.18\linewidth]{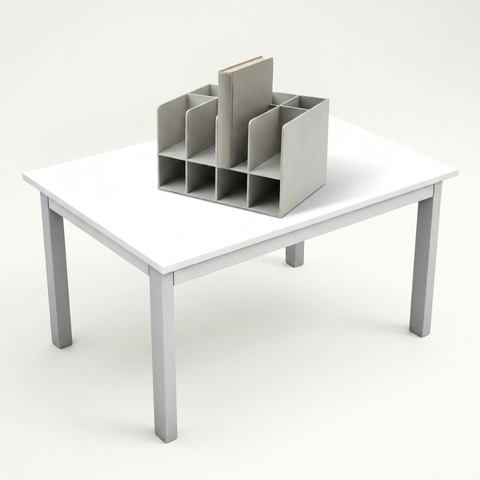} \\
\end{tabular}
\caption{Reference scenes for \textbf{C.6 Slotting}. Top to bottom rows show plate-dish-rack, wine-rack, and book-stand variants.}
\label{fig:supp:gallery:c6}
\end{figure}

\subsection{C.7 Spanning}
\label{sec:supp:tasks:c7}

\noindent\textbf{Task 1: Span a fork across a plate rim.}
\begin{itemize}\setlength{\itemsep}{1pt}
    \item The fork bridges the plate rim like a small ramp: the fork head and downward-facing tines are inside the plate, the fork neck crosses the rim, and the handle slopes downward outside the plate until the handle end rests on the tabletop, not on the plate. The fork is a polished stainless steel dinner fork with four tines and a long narrow handle, shown upside down with the back side facing upward. The plate is a round white ceramic plate with a shallow raised rim and glossy surface, fully visible on a clean white tabletop.
    \item The fork is tilted like a ramp over the plate rim: the fork head and downward-facing tines touch the inside of the plate, the fork neck is supported by the raised rim, and the handle angles downward outside the plate with the rounded handle tip visibly touching the tabletop below the rim. The fork is a matte black metal fork with four slim tines and a straight handle, shown upside down with the back side facing upward. The plate is a round matte gray stoneware plate with a raised rim, fully visible on a clean white tabletop.
    \item The fork lies diagonally over the rim of the plate, with the tines pointing downward toward the center of the plate and the handle extending outside the rim. The fork is a gold-colored fork with a reflective handle and four rounded tines. The plate is a glossy blue ceramic plate with a shallow bowl-like center and smooth circular rim.
    \item The fork is tilted like a ramp over one side edge of the plate: the fork head and downward-facing tines touch the inside of the plate, the fork neck is supported by the raised plate edge, and the handle angles downward outside the plate with the rounded handle tip visibly touching the tabletop below the edge. The fork is a small silver dessert fork with short tines and a tapered handle, shown upside down with the back side facing upward. The plate is a square white porcelain plate with rounded corners and a raised edge.
    \item The fork is tilted like a ramp over the plate rim with the handle angling downward outside the plate and the rounded handle tip visibly touching the tabletop below the rim. The fork has a brushed steel head, four tines, and a light wooden handle, shown upside down with the back side facing upward. The plate is a round cream ceramic plate with a slightly speckled glaze and a raised rim.
\end{itemize}

\noindent\textbf{Task 2: Span a calligraphy brush across a brush stand.}
\begin{itemize}\setlength{\itemsep}{1pt}
    \item The calligraphy brush rests on the brush stand with visible support under the brush shaft so it is stable and cannot roll away. The calligraphy brush is one complete traditional calligraphy brush with a pale wooden handle, black ferrule, tapered dark bristles, and a small hanging loop at the rear end. The brush stand is one small dark wooden brush stand with carved U-shaped cradle notches supporting the brush shaft from below.
    \item The calligraphy brush rests on the brush stand with visible support under the brush shaft. The calligraphy brush is one complete calligraphy brush with dark bristles, a dark ferrule, warm brown handle, and rear hanging loop. The brush stand is one dark carved wooden brush stand with multiple U-shaped notches, two notches visibly supporting the brush shaft from below.
    \item The calligraphy brush rests on the brush stand with visible support under the brush shaft. The calligraphy brush is one complete calligraphy brush with dark pointed bristles, black ferrule, orange-brown handle, and rear hanging loop. The brush stand is one small blue and white ceramic brush stand with a raised cradle groove supporting the brush shaft.
    \item The calligraphy brush rests on the brush stand with visible support under the brush shaft. The calligraphy brush is one complete calligraphy brush with dark pointed bristles, black ferrule, brown wooden handle, and rear hanging loop. The brush stand is one small cream ceramic brush stand with two rounded cradle humps supporting the brush shaft.
    \item The calligraphy brush rests on the brush stand with visible support under the brush shaft. The calligraphy brush is one complete calligraphy brush with dark pointed bristles, black ferrule, warm brown handle, and rear hanging loop. The brush stand is one small blue and white ceramic brush stand with a triangular upright cradle supporting the brush shaft.
\end{itemize}

\noindent\textbf{Task 3: Span a soup ladle across a vertical ladle stand.}
\begin{itemize}\setlength{\itemsep}{1pt}
    \item The soup ladle rests upright in the lower circular cradle of the vertical ladle stand, while the handle leans upward through and is guided by the stand's tall wire guide. The soup ladle is one complete soup ladle with a deep rounded bowl and a long slim handle, clean reflective metal with subtle brushed highlights, plus a richer mixed-tone finish, with a hanging hole at the handle tip. The vertical ladle stand is one complete brushed-metal vertical ladle stand with a tall wire frame shaped like a triangular tri-prong guide, a stable weighted base, and a lower circular saucer cradle specifically sized to hold the ladle bowl.
    \item The soup ladle bowl rests in the lower circular cradle while the long handle leans upright through the tall wire guide of the gold-toned wire ladle stand. The soup ladle is one complete soup ladle with a deep rounded bowl, long slim handle with hanging hole, clean reflective metal surface. The vertical ladle stand is one gold-toned wire vertical soup ladle stand with a decorative upright loop and small round bowl support, stable wire base.
    \item The soup ladle rests upright with its bowl sitting in the lower circular cradle of the vertical ladle stand while the long handle passes through and is guided by the tall wire channel. The soup ladle is one complete soup ladle with a deep rounded bowl and a long slim handle, clean surface with a richer matte color finish, subtle realistic metal sheen, and a hanging hole at the handle end. The vertical ladle stand is one realistic matte steel vertical ladle stand with a minimal rod-and-wire frame, a front lower circular cradle that supports the ladle bowl, and a tall rear wire channel that aligns the handle upright.
    \item The multicolored soup ladle rests upright on the lower circular cradle of the white enamel vertical ladle stand, with the long handle guided vertically through the stand's tall wire frame. The soup ladle is one complete soup ladle with a deep rounded bowl and a long slim handle featuring a hanging hole; the bowl has a rich multicolor enamel finish, and the handle is smooth metallic or enamel coated. The vertical ladle stand is one white enamel wire vertical ladle stand with a stable base and a lower circular cradle that supports the ladle bowl; the upper section forms a tall, open wire guide channel that the ladle handle passes through.
    \item The soup ladle bowl rests in the lower circular cradle while the long handle leans upright through the tall wire guide of the dark bronze wire ladle stand. The soup ladle is one complete soup ladle with a deep rounded bowl, long slim handle with hanging hole, clean reflective metal surface. The vertical ladle stand is one dark bronze wire vertical soup ladle stand, upright loop frame with stable circular foot.
\end{itemize}

\begin{figure}[h]
\centering
\setlength{\tabcolsep}{1pt}
\renewcommand{\arraystretch}{0}
\begin{tabular}{ccccc}
\includegraphics[width=0.18\linewidth]{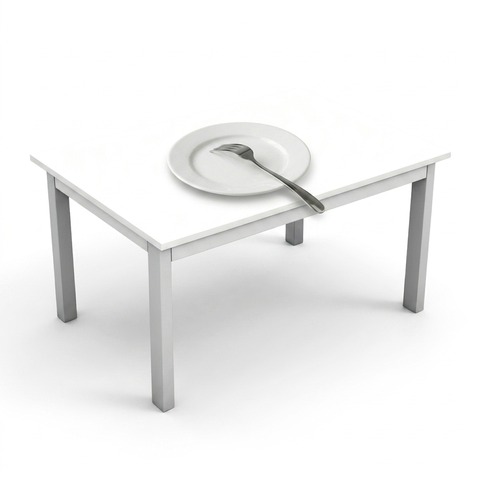} &
\includegraphics[width=0.18\linewidth]{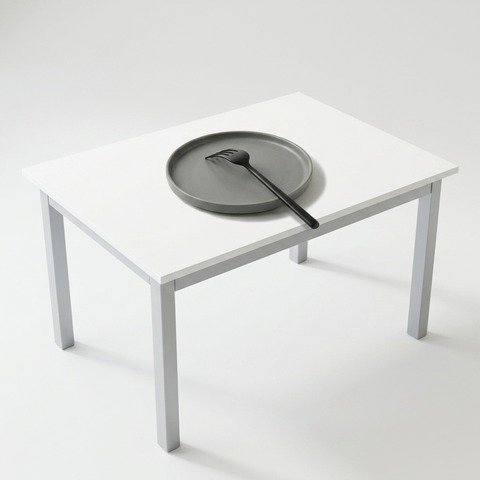} &
\includegraphics[width=0.18\linewidth]{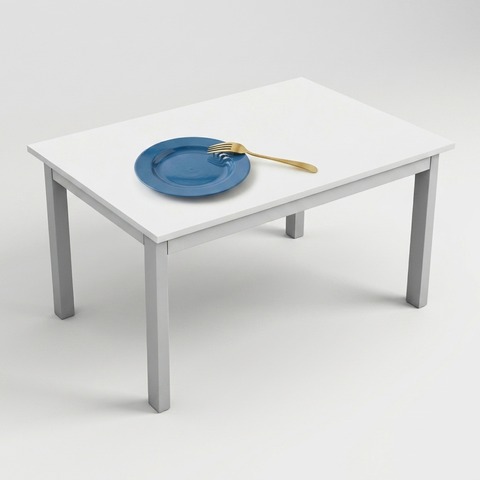} &
\includegraphics[width=0.18\linewidth]{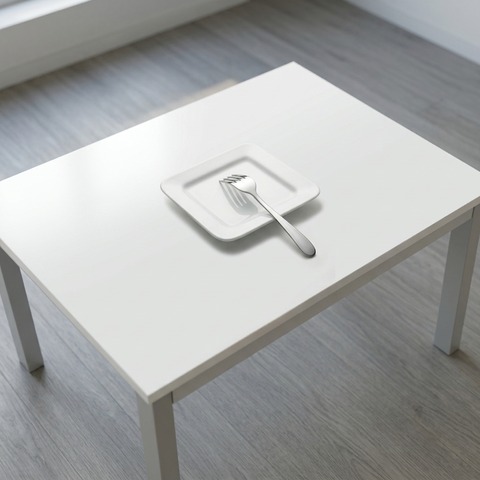} &
\includegraphics[width=0.18\linewidth]{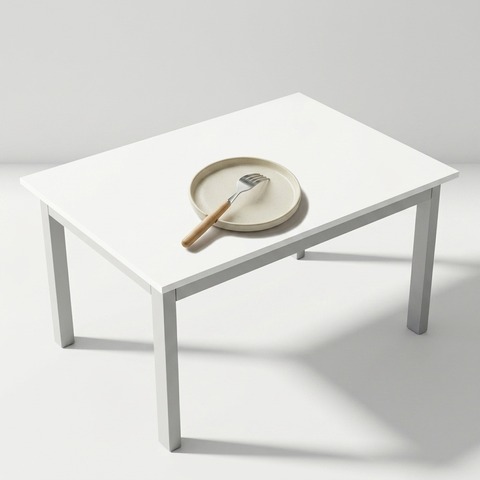} \\
\includegraphics[width=0.18\linewidth]{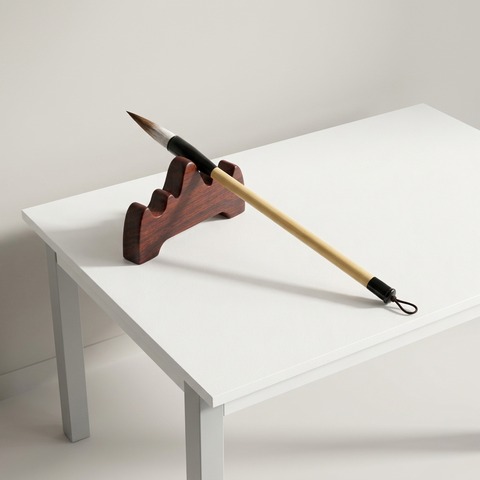} &
\includegraphics[width=0.18\linewidth]{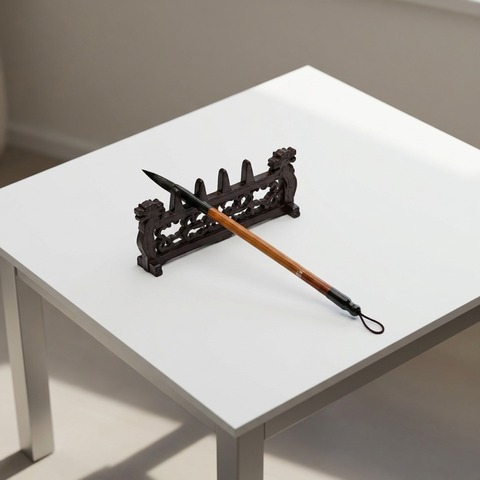} &
\includegraphics[width=0.18\linewidth]{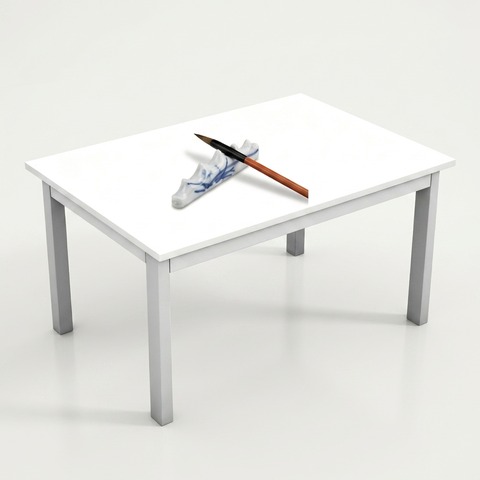} &
\includegraphics[width=0.18\linewidth]{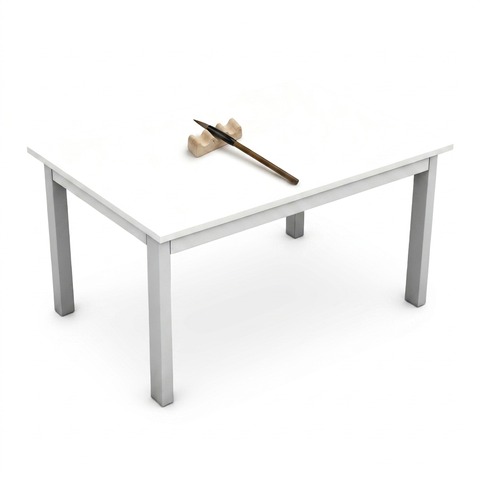} &
\includegraphics[width=0.18\linewidth]{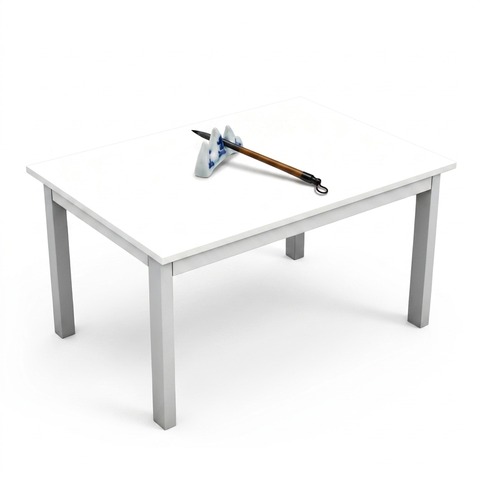} \\
\includegraphics[width=0.18\linewidth]{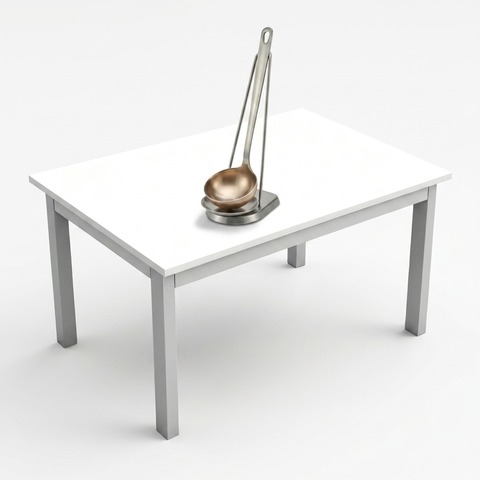} &
\includegraphics[width=0.18\linewidth]{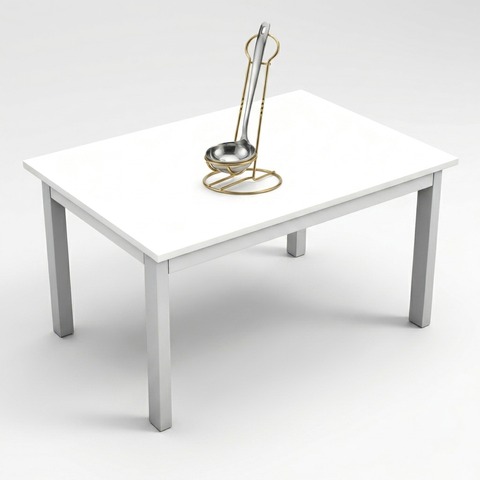} &
\includegraphics[width=0.18\linewidth]{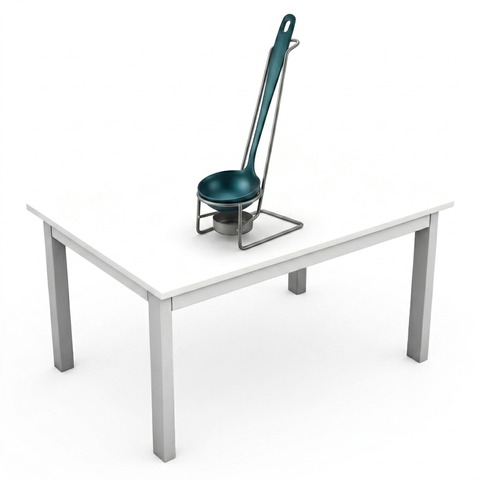} &
\includegraphics[width=0.18\linewidth]{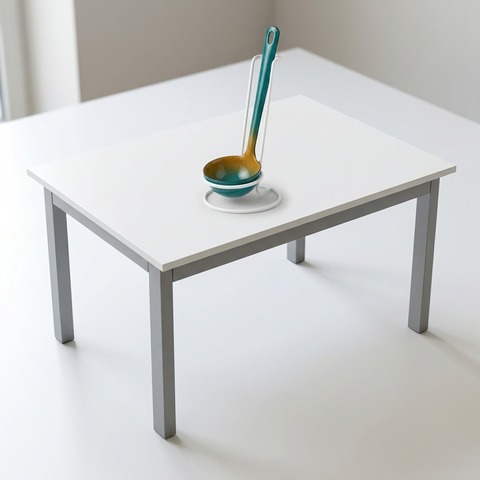} &
\includegraphics[width=0.18\linewidth]{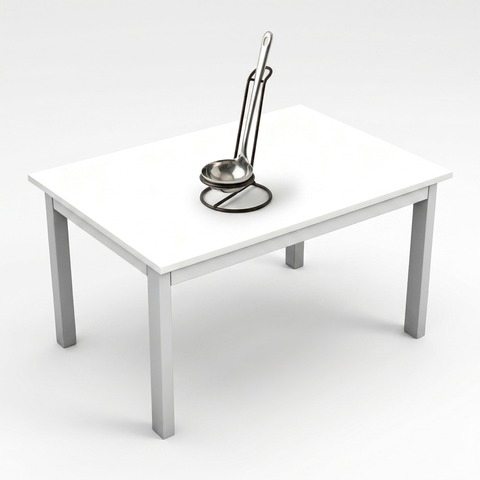} \\
\end{tabular}
\caption{Reference scenes for \textbf{C.7 Spanning}. Top to bottom rows show fork-plate-rim, calligraphy-brush-stand, and ladle-stand variants.}
\label{fig:supp:gallery:c7}
\end{figure}

\subsection{C.8 Pegging}
\label{sec:supp:tasks:c8}

\noindent\textbf{Task 1: Peg a toss ring onto a ring toss stand.}
\begin{itemize}\setlength{\itemsep}{1pt}
    \item The single red toss ring is looped around one upright peg of the ring toss stand and sits visibly on the peg. The toss ring is one complete red circular toss ring with smooth rounded tube shape. The ring toss stand is one toy-like ring toss stand with a stable base and several upright empty pegs, colorful simple plastic or painted wood construction.
    \item The single yellow toss ring is looped around one upright peg of the ring toss stand and sits visibly on the peg. The toss ring is one complete yellow circular toss ring with smooth rounded tube shape. The ring toss stand is one toy-like ring toss stand with a stable base and several upright empty pegs.
    \item The single green ring encircles the vertical tapered cone post near the lower third of the stand; the post passes through the ring's center hole and the ring rests against the cone above the solid base. The toss ring is one complete green translucent circular toss ring with smooth rounded tube shape. The ring toss stand is one toy-like ring toss stand with a single solid round disk base and one vertical tapered cone post, stacked colored bands on the post.
    \item The orange ring is looped around the tapered wooden cone so the cone passes through the ring center; the ring sits tilted slightly on the lower cone and touches the base, physically captured by the post. The toss ring is one complete matte orange rubber toss ring with a thick rounded tube. The ring toss stand is one single-post ring toss stand with a circular wooden base and a smooth vertical tapered wooden cone, natural light wood grain with a darker walnut top cap.
    \item The pink ring is looped around the vertical cone post and rests around the lower part of the cone above the solid pedestal base. The toss ring is one complete pink silicone circular toss ring with thick soft rounded tube shape. The ring toss stand is one single upright ring toss cone stand with a solid oval pastel lavender pedestal base and one vertical tapered cone post, satin plastic material.
\end{itemize}

\noindent\textbf{Task 2: Peg a jewelry ring onto a ring holder.}
\begin{itemize}\setlength{\itemsep}{1pt}
    \item The jewelry ring is fitted around the tapered support of the matte black ring holder. The jewelry ring is one complete small jewelry ring with a smooth metallic band, simple gemstone or plain top. The ring holder is one matte black ring holder with a tapered cone or upright finger-shaped support, stable base, and no other jewelry.
    \item The jewelry ring is fitted around the tapered support of the warm white ring holder. The jewelry ring is one complete small jewelry ring with a smooth metallic band, simple gemstone or plain top. The ring holder is one warm white ring holder with a tapered cone or upright finger-shaped support, stable base.
    \item The jewelry ring is fitted around the tapered support of the natural wood ring holder. The jewelry ring is one complete small jewelry ring with a smooth metallic band, simple gemstone or plain top. The ring holder is one natural wood ring holder with a tapered cone or upright finger-shaped support, stable base.
    \item The jewelry ring is fitted around the tapered support of the brass colored ring holder. The jewelry ring is one complete small jewelry ring with a smooth metallic band, simple gemstone or plain top. The ring holder is one brass colored ring holder with a tapered cone or upright finger-shaped support, stable base.
    \item The jewelry ring is fitted around the tapered support of the light gray ring holder. The jewelry ring is one complete small jewelry ring with a smooth metallic band, simple gemstone or plain top. The ring holder is one light gray ring holder with a tapered cone or upright finger-shaped support, stable base.
\end{itemize}

\noindent\textbf{Task 3: Peg a mug onto a cup rack.}
\begin{itemize}\setlength{\itemsep}{1pt}
    \item The mug hangs upside down on one vertical peg of the wooden cup rack, with the peg entering the open mouth of the mug and supporting it securely. The mug is one complete dark navy blue mug with a handle, open mouth facing downward. The cup rack is one complete light wooden cup rack with a rectangular base, multiple vertical cylindrical pegs, and simple cross rails.
    \item The mug hangs upside down on one vertical peg of the cup rack, with the peg entering the open mouth of the mug. The mug is one complete clear glass cup with open mouth facing downward. The cup rack is one white tabletop cup rack with a round base and multiple vertical curved pegs, one peg supporting the cup.
    \item The mug hangs upside down on one vertical peg of the cup rack, with the peg entering the open mouth of the mug. The mug is one complete sage green mug with handle and open mouth facing downward. The cup rack is one gray rectangular cup rack with several vertical cylindrical pegs and a flat base.
    \item The mug hangs upside down on one vertical peg of the cup rack, with the peg entering the open mouth of the mug. The mug is one complete clear glass cup with blue rim and open mouth facing downward. The cup rack is one chrome rotating cup rack with a circular base and multiple vertical arms.
    \item The mug hangs upside down on one vertical peg of the cup rack, with the peg entering the open mouth of the mug. The mug is one complete amber glass cup with open mouth facing downward. The cup rack is one silver metal cup rack with forked vertical supports and a rectangular base.
\end{itemize}

\begin{figure}[h]
\centering
\setlength{\tabcolsep}{1pt}
\renewcommand{\arraystretch}{0}
\begin{tabular}{ccccc}
\includegraphics[width=0.18\linewidth]{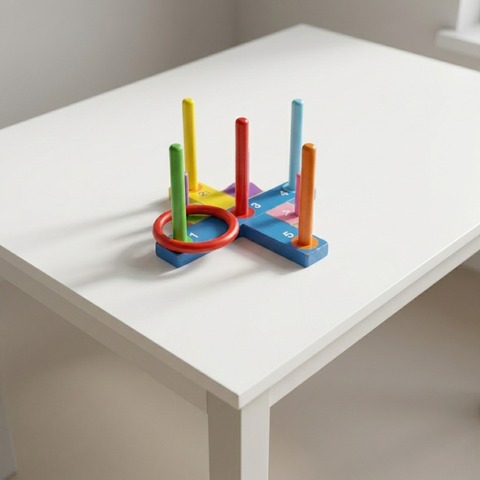} &
\includegraphics[width=0.18\linewidth]{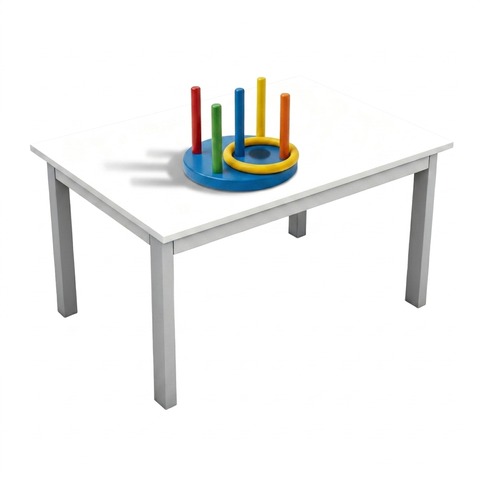} &
\includegraphics[width=0.18\linewidth]{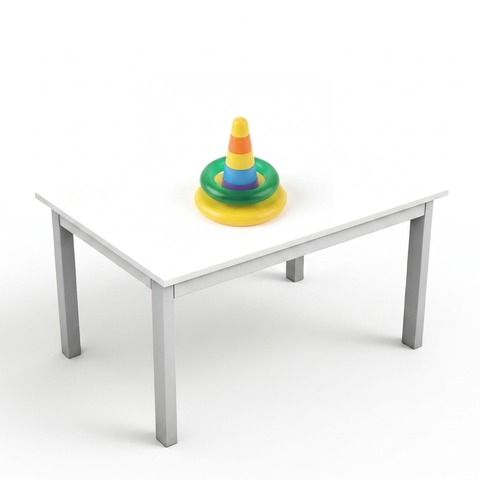} &
\includegraphics[width=0.18\linewidth]{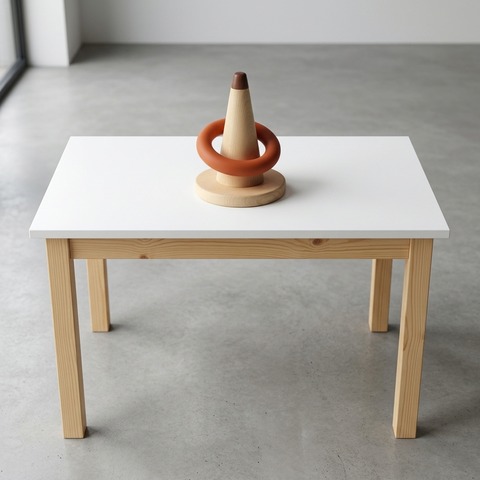} &
\includegraphics[width=0.18\linewidth]{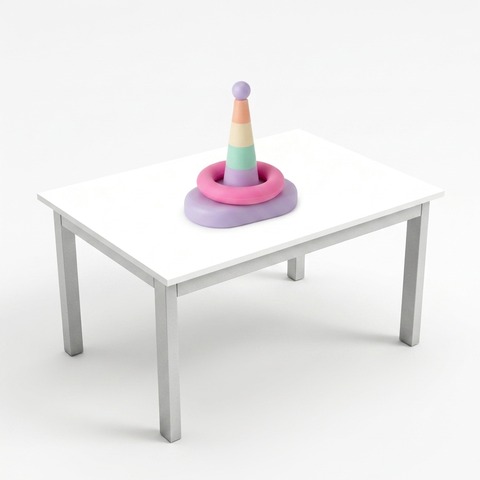} \\
\includegraphics[width=0.18\linewidth]{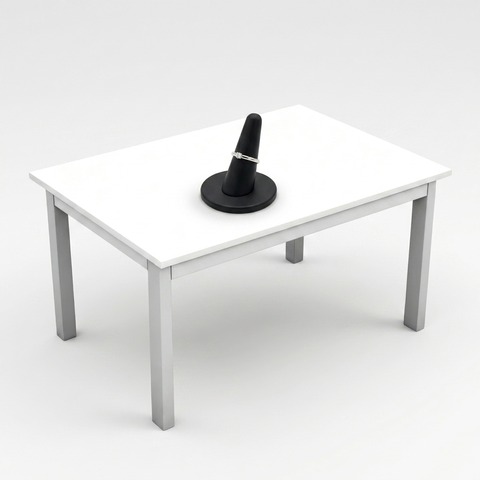} &
\includegraphics[width=0.18\linewidth]{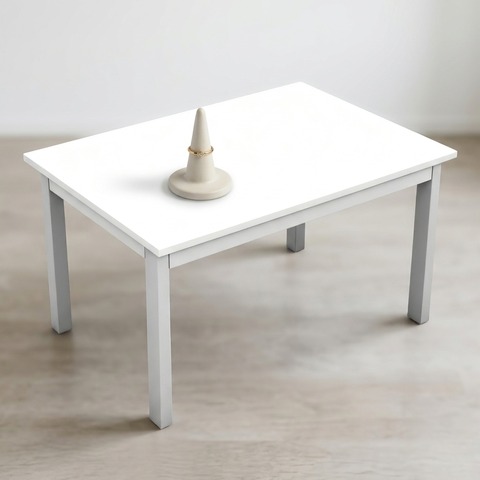} &
\includegraphics[width=0.18\linewidth]{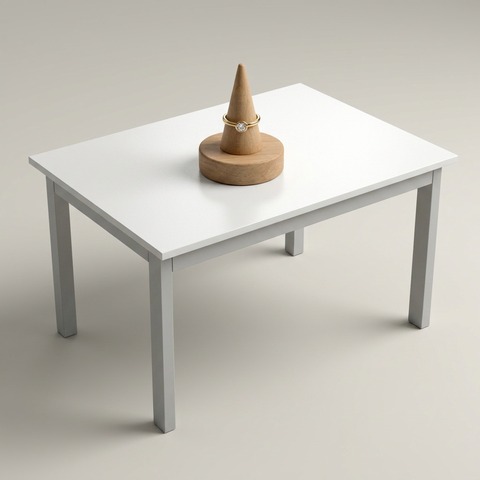} &
\includegraphics[width=0.18\linewidth]{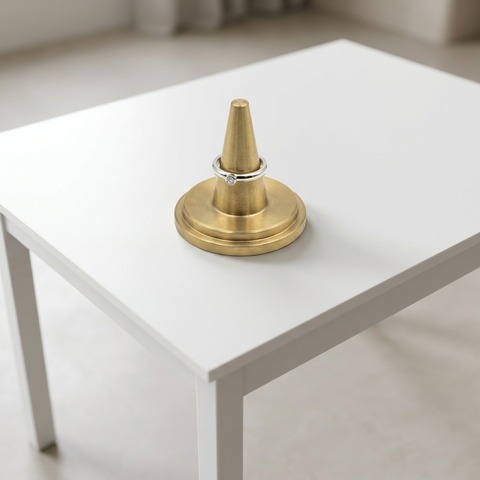} &
\includegraphics[width=0.18\linewidth]{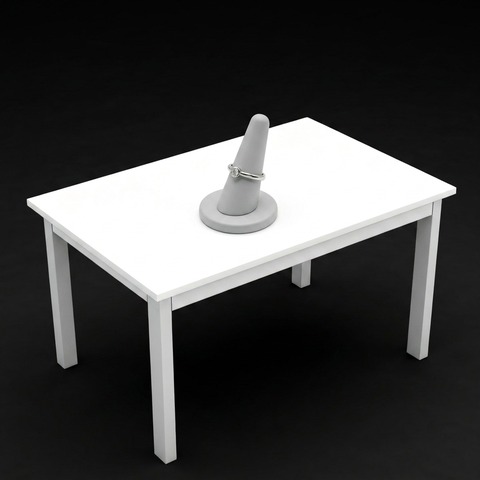} \\
\includegraphics[width=0.18\linewidth]{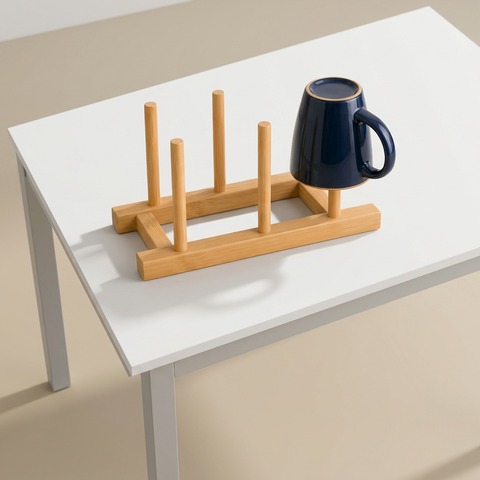} &
\includegraphics[width=0.18\linewidth]{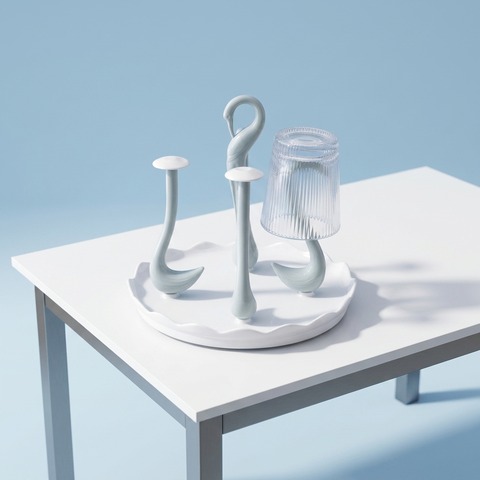} &
\includegraphics[width=0.18\linewidth]{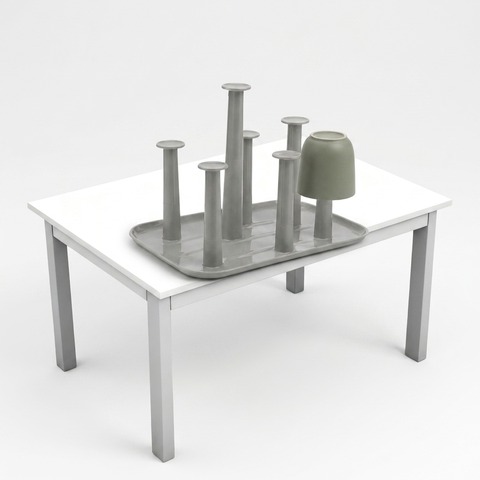} &
\includegraphics[width=0.18\linewidth]{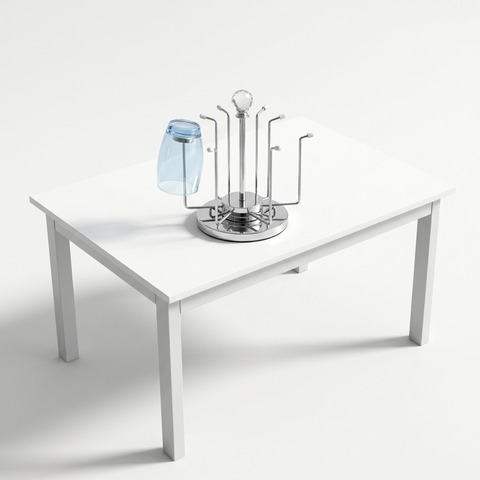} &
\includegraphics[width=0.18\linewidth]{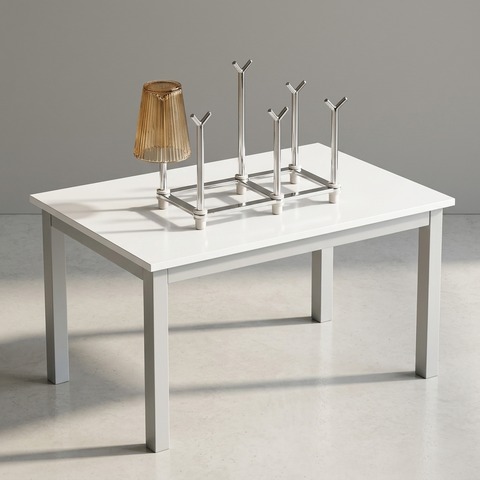} \\
\end{tabular}
\caption{Reference scenes for \textbf{C.8 Pegging}. Top to bottom rows show ring-toss-stand, jewelry-ring-holder, and mug-cup-rack variants.}
\label{fig:supp:gallery:c8}
\end{figure}

\section{Prompt Templates}
\label{sec:supp:prompts}

This section lists the prompt templates that instantiate the agentic decisions described in the main paper and earlier sections of the supplement. Each box preserves the operative constraints and machine-readable output schema of the corresponding template, while omitting incidental boilerplate.

\subsection{Input scene agent}
\label{sec:supp:prompt:input_scene_agent}
The input scene agent of Sec.~\ref{sec:supp:method:input} consumes an arbitrary combination of language, scene image, and asset images, and decides whether to draw a fresh scene or edit the user-supplied one. Its system prompt encodes the scene rules, the closed vocabulary of relations, the image-quality criteria, and the JSON output schema.

\begin{promptbox}[title={Input scene agent (system)}]
\promptsection{Goal}
Produce a clean reference image suitable for 3D reconstruction together with a RootData JSON that records the primary target, the contact object, their relation, and a background color.

\promptsection{Scene rules}
Exactly two objects, primary target plus contact object, with no distractors. Solid background such as white, light gray, beige, or a hex value. Camera at an elevated angle so both objects are clearly visible. Target and contact must be physically separable as independent rigid bodies and must not be connected by thin continuous structures such as stems, vines, chains, or ropes.

\promptsection{Relation vocabulary}
on, in, inside, above, under, stack\_on, cover, hang\_on, lean\_against, rest\_on, attach\_to, insert\_into, fit\_in.

\promptsection{Inputs and policy}
You may receive a task description, a scene image that may be messy, one or two asset images each labeled \texttt{TARGET} or \texttt{CONTACT}, and an optional task hint. Analyze all provided modalities to identify the target, contact, and relation. Then act: prefer \texttt{edit\_image} when a usable scene image is given and only minor cleanup is needed, otherwise call \texttt{draw\_image} with the available asset and scene paths in \texttt{reference\_image\_paths} so the generator preserves their appearance. Fall back to text-only generation only when no images are provided. After every action, verify the result against the quality criteria below and retry up to \{max\_iterations\} times.

\promptsection{Quality criteria}
1.~Background is solid or near-solid with no textures, walls, or floor patterns.
2.~Object count is exactly one target and one contact; structurally connected parts of the same object are still one object.
3.~Both objects are substantially visible; functional occlusion such as contents inside a container is acceptable.
4.~The depicted contact is the way a human would actually arrange these specific objects for the given relation.
5.~Camera is elevated and not directly overhead or directly side-on.
6.~No floating, interpenetrating, or geometrically broken parts.
7.~Target and contact are independent rigid bodies, with no thin continuous links.

\promptsection{Output JSON}
\{ task\_description, primary\_target\{name, description, role\}, contact\_object\{name, description, role\}, target\_contact\_relation, reference\_image \}
\end{promptbox}

\subsection{Synonym retry for mask preparation}
\label{sec:supp:prompt:synonyms}
When the mask-preparation step returns an empty or low-confidence mask, the synonym module proposes alternative phrasings that are more likely to ground to the intended object.

\begin{promptbox}[title={Mask-preparation synonym generator}]
\promptsection{Role}
You generate alternative text prompts for a mask-preparation step that fails on the original object name. The mask-preparation step prefers short noun phrases, struggles with overly specific or niche terms, and benefits from visual descriptors of color, material, and shape.

\promptsection{Strategies (apply in order)}
1.~Generalization: move up the category hierarchy, e.g.\ ``sundial'' $\to$ ``statue''.\\
2.~Visual adjectives: add color, material, or shape, e.g.\ ``ball'' $\to$ ``red ball''.\\
3.~Category names: use the superclass, e.g.\ ``mug'' $\to$ ``cup''.\\
4.~Creative indirect terms: use related visual concepts, e.g.\ ``nose'' $\to$ ``black marking''.\\
5.~Simplification: strip non-visual modifiers, e.g.\ ``ceramic coffee mug with handle'' $\to$ ``mug''.

\promptsection{Output JSON}
\{ alternatives [exactly 5 entries, ranked best-first, all distinct from the original name], reasoning \}
\end{promptbox}

\subsection{Asset erase agent}
\label{sec:supp:prompt:erase_agent}
The erase agent of Sec.~\ref{sec:supp:method:cogen} alternates between editing and verifying with a strict pose-first rubric. Its system prompt is reproduced below.

\begin{promptbox}[title={Asset erase agent (system)}]
\promptsection{Goal}
Erase the named target object from the input image while preserving the partner object exactly in its current position, orientation, scale, and posture. Minor appearance changes such as color shift or texture smoothing are acceptable; pose changes are not.

\promptsection{Workflow (repeat up to \{max\_iterations\} times)}
1.~Call \texttt{edit\_image} with a precise prompt that names the target object and explicitly forbids any motion of the preserved object.
2.~Call \texttt{compare\_images} to obtain a labeled side-by-side comparison whose left panel is the original and whose right panel is the current edit.
3.~Inspect both panels and call \texttt{score\_image} with an integer score in [0,100] and a short feedback string. Apply the scoring procedure below.

\promptsection{Scoring procedure}
Step A: locate the preserved object in both panels and check whether its pixel position, size, and orientation match.
Step B: check whether the target object has been fully removed from the right panel.
Step C: check the background fill for naturalness and absence of artefacts.
Step D: apply the rubric of Tab.~\ref{tab:supp:erase_rubric}: $-50$ for any pose change, $-25$ for residual visibility, $-15$ for appearance change of the preserved object with intact pose, $-10$ for background artefacts.

\promptsection{Tips for edit prompts}
Name the target precisely with color, shape, and screen location. Say explicitly that the preserved object must not move, rotate, or resize. Specify that the background fill should match the surrounding area seamlessly. If the previous attempt shifted the preserved object, repeat in the next prompt that the preserved object must stay at exactly the same pixel location and orientation.

\promptsection{Termination}
If the score is at least 90, output the result path and stop. Otherwise, if iterations remain, call \texttt{edit\_image} again. After all iterations, output the path of the best-scored result.
\end{promptbox}

\subsection{Asset symmetry analyzer}
\label{sec:supp:prompt:symmetry}
Symmetry tagging guides the pose-equivalence sampling of Sec.~\ref{sec:method:data}. Rather than asking for purely geometric symmetry, the analyzer is constrained to task-driven functional equivalence: rotations are admitted only when the rotated object still fulfills the contact relation.

\begin{promptbox}[title={Asset symmetry analyzer}]
\promptsection{Role}
You are an expert in 3D kinematics, geometry, and robotic grasping. Symmetry types are the cyclic groups C1, C2, C3, C4, Cinf.

\promptsection{Coordinate system}
Do NOT use default axis assumptions. Read the labeled arrows in the input image: red+``X'' is X, green+``Y'' is Y, blue+``Z'' is Z.

\promptsection{Task-driven functional equivalence}
Symmetry must be judged with respect to the contact relation. Identify the alignment requirement, idealize away minor bends and textures, and apply the lift-rotate-place test: lift the object, rotate it by a candidate angle, place it back; if the post-placement state still fulfills the task, that rotation is admissible. Path collisions during the rotation are irrelevant.

\promptsection{Examples}
Mug hang\_on hook: C2 about vertical axis, since a $180^{\circ}$ flip still lets the hook pass through the handle.\\
Mug rest\_on coaster: Cinf about vertical axis, since the bottom contacts identically at any yaw.\\
Sausage in rectangular box: Cinf about long axis after idealization, C2 about vertical axis.

\promptsection{Output JSON}
\{ object\_name, step1\_shape\_idealization\{...\}, step2\_task\_requirements\{...\}, x\_axis\{motion, 180/90 checks, symmetry\_type, reasoning\}, y\_axis\{...\}, z\_axis\{...\}, overall\_confidence \}
\end{promptbox}

\subsection{Multi-view scene verifier}
\label{sec:supp:prompt:verify_multi}
After the pose pipeline produces a candidate scene, the verifier inspects MuJoCo renderings, optionally arranged as multi-view or time-series collages, and decides whether the contact relation is preserved under gravity.

\begin{promptbox}[title={Multi-view scene verifier}]
\promptsection{Context}
Images are MuJoCo renders, not AI-generated. Physics, multi-view consistency, and temporal consistency are guaranteed by the simulator. A time-series, when present, is a passive gravity-only stability rollout, not a robot executing the task. The final frame is therefore the gravity-settled state, not the expected manipulation outcome.

\promptsection{Task}
Verify three properties: (i) the target-contact relation matches the expected relation; (ii) for time series, the relation is preserved at the final settled frame; (iii) the resulting scene is feasible for grasping. Do not reward candidates whose ``final'' state corresponds to a removed or separated target, even if it visually resembles a completed manipulation.

\promptsection{Critical defects}
Target or contact object missing, wrong spatial relation, or final settled state breaking the expected relation. Texture artefacts and minor lighting differences are acceptable.

\promptsection{Output JSON}
\{ passes, relation\_ok, reason, issues, relation\_issues, confidence \}
\end{promptbox}

\subsection{Candidate voter}
\label{sec:supp:prompt:voter}
When multiple pose candidates survive verification, the voter compares them on a single horizontal montage where each candidate occupies one labeled panel separated by red dividers, and selects the best one.

\begin{promptbox}[title={Candidate voter}]
\promptsection{Image structure}
A single horizontal montage. Candidates are arranged left to right, separated by thick red vertical bars. Each panel carries a CANDIDATE \#N header. Inside the panel a per-candidate timeline (perspective view on the left, top view on the right) progresses left to right; this timeline is a passive gravity-only MuJoCo rollout, not a robot execution.

\promptsection{Critical interpretation rule}
A candidate is an initial scene that should already satisfy the contact relation; the gravity-settled final frame must continue to satisfy that relation. Do not prefer a candidate where the target is removed, separated, or fallen away merely because it resembles task completion.

\promptsection{Evaluation priorities}
1.~Relation preservation under gravity at the final frame.\\
2.~Asset quality: target and contact must be separate clean meshes, not fused or doubled; reject reconstruction artefacts even if alignment looks tight.\\
3.~Physical plausibility: no severe interpenetration, flying, or instability.\\
4.~Overall reasonableness for robotic manipulation. A candidate with perfect alignment but corrupted geometry must score lower than one with mild misplacement and clean meshes.

\promptsection{Output JSON}
\{ best\_index, reason, confidence \}
\end{promptbox}

\subsection{Multi-object scene expander}
\label{sec:supp:prompt:expand}
The expander plans up to \{max\_new\_objects\} insertions for an already verified scene.
It reads a pullback rendering together with the current object list, where each entry contains the object name, role, and description.
If a user request is provided, the planner must include the requested objects subject to the insertion budget; otherwise, it autonomously selects semantically related additions.
In both cases, it outputs reconstruction-oriented visual descriptions, spatial relations to existing objects, an image-editing prompt, and a preservation description for the editor.

\begin{promptbox}[title={Scene expander}]
\promptsection{System}
You are a scene expansion planner for robotic grasping scenarios.
Given an existing scene with objects, plan additional objects to add.
When the user specifies objects, follow their instructions.
When no user request is given, suggest objects that make the scene more realistic and natural.
Output valid JSON only, with no markdown.

\promptsection{Inputs}
Analyze the rendered scene image and plan up to \{max\_new\_objects\} new objects to add.
Existing objects are provided as \texttt{- name (role): description}.
An optional user request may specify the objects to add.

\promptsection{Planning policy}
If a user request is provided, include the requested objects in the plan.
Parse the request to determine object names, visual descriptions, reasonable positions relative to existing objects, and appropriate spatial relations.
Descriptions may be adjusted to improve 3D reconstruction quality, but the core requested objects must not be changed.
If the insertion budget is restrictive, add fewer requested objects while prioritizing the user's choices.
If no user request is provided, suggest objects that are semantically related to the existing scene, e.g.\ pen-holder $\to$ pens, desk $\to$ books or ruler, coat rack $\to$ coat or hat.

\promptsection{Requirements}
1.~Each new object has a clear spatial relation to an existing object.\\
2.~Sizes are reasonable relative to existing objects.\\
3.~Generate an image-editing prompt that preserves the existing scene, viewpoint, lighting, and style.

\promptsection{Output JSON}
\{ objects\_to\_add[\{name, description, suggested\_position\}], spatial\_relations[\{subject, relation $\in$ \{in, on, left\_of, right\_of, behind, stack\_on, lean\_against, in\_front\_of, cover, hang\_on\}, object\}], edit\_prompt, preserve\_description, reasoning \}
\end{promptbox}

\subsection{VLM scene scorer}
\label{sec:supp:prompt:score}
The VLM scorer rates a single rendered scene on Object Matching and Relationship Matching, supplying the per-scene scores aggregated in Tab.~\ref{tab:main_results} of the main paper.

\begin{promptbox}[title={VLM scene scorer}]
\promptsection{Context}
A 3D scene was reconstructed from images, placed in MuJoCo, and allowed to settle under gravity. Some upstream objects may be held static, so the settled image can still expose floating, balanced, or overhung configurations. Judge plausibility under gravity directly, do not assume settling guarantees plausibility.

\promptsection{Language-conditioned input}
You receive (1) the natural-language task description and (2) the gravity-settled MuJoCo render. Judge whether both generated objects match the identities and attributes named in the instruction; do not penalize valid appearance choices that the text leaves unspecified.

\promptsection{Image-conditioned input}
You receive (1) the task description, (2) the input reference image, and (3) the gravity-settled MuJoCo render. Judge object appearance and geometry against the reference image.

\promptsection{Score 1, Object Matching (0-10)}
0: at least one object missing. 1-3: both present but identities or specified attributes differ substantially. 4-6: rough semantic match with noticeable discrepancies. 7-9: close match to the instruction (and, for image-conditioned evaluation, the reference image) with minor differences only. 10: perfect match under the applicable modality.

\promptsection{Score 2, Relationship Matching (0-10)}
Apply the decision procedure: (a) does the geometry satisfy the named relation; (b) is the arrangement common-sense correct, that is, how a human would actually place the objects for this relation; (c) is there visible interpenetration; (d) is the pose physically plausible under gravity. Any failure caps the score: 0 for absent or reversed; 1-3 for a clear failure on geometry, plausibility, or common sense; 4-6 for partial satisfaction with a clear defect; 7-9 only when geometry, plausibility, and common-sense placement are all unambiguously satisfied; 10 for perfect.

\promptsection{Output JSON}
\{ target\_present, contact\_present, object\_match\_score, object\_match\_reasoning, relation\_score, relation\_reasoning \}
\end{promptbox}

\subsection{Ego4D-driven taxonomy clustering}
\label{sec:supp:prompt:taxonomy}
The benchmark of Sec.~\ref{sec:supp:exp:benchmark} is built by an LLM agent that ingests filtered Ego4D narrations in shards. Each shard is summarized into candidate relation phrases, the phrases are merged across shards, and the merged set is clustered by contact geometry. The three prompts that drive these stages are listed below.

\begin{promptbox}[title={Stage 1: per-shard relation summarization}]
\promptsection{Role}
You read short egocentric narrations of human manipulation, each describing one camera-wearer manipulating one named target with respect to one named contact partner.

\promptsection{Task}
For every narration, output a single relation phrase that names the contact geometry between the target and the partner, ignoring linguistic surface form. Use short noun-style phrases such as ``ring around peg'' or ``lid covering opening'', and do not invent objects that are not in the narration.

\promptsection{Output JSON}
\{ narrations[\{video\_uid, narration\_text, target, contact, relation\_phrase\}] \}
\end{promptbox}

\begin{promptbox}[title={Stage 2: cross-shard phrase merging}]
\promptsection{Role}
You merge relation phrases that describe the same contact geometry but were produced by different shards. Phrases are equivalent only when the rotation, translation, and contact pattern of the target with respect to the partner are interchangeable.

\promptsection{Task}
Group equivalent phrases under a single canonical phrase, choose the most precise wording for each canonical, and discard phrases that no longer describe a single contact geometry after merging.

\promptsection{Output JSON}
\{ canonical\_phrases[\{canonical, members[...] \}] \}
\end{promptbox}

\begin{promptbox}[title={Stage 3: contact-geometry clustering}]
\promptsection{Role}
You partition canonical relation phrases into a small number of contact-geometry classes. The partition must be mutually exclusive, exhaustive, and have between six and ten classes.

\promptsection{Constraints}
Each class must correspond to a distinct contact pattern that admits a closed-form geometric template, such as a containment volume, a planar contact, an axial peg, or a hooking link. Avoid splits that depend only on object identity. Provide a one-line definition for each class together with two to three representative member phrases.

\promptsection{Output JSON}
\{ classes[\{class\_id, name, definition, members[...] \}] \}
\end{promptbox}

\begin{promptbox}[title={Per-class task selection}]
\promptsection{Role}
You select three representative manipulation tasks per contact-geometry class for benchmark construction.

\promptsection{Constraints}
For each class produce three tasks under the following rules: (i) each task instantiates the class with two distinct everyday objects; (ii) the two objects must be physically separable as independent rigid bodies; (iii) the union of the three tasks should cover both household and workplace scenarios. Provide a one-sentence natural-language instruction per task and a short justification per choice.

\promptsection{Output JSON}
\{ class\_id, tasks[\{task\_name, target, contact, instruction, justification\}] \}
\end{promptbox}

\end{document}